\documentclass[10pt,journal,letterpaper,compsoc]{IEEEtran}
\usepackage{times}
\usepackage{epsfig}
\usepackage{graphicx}
\usepackage{amsmath}
\usepackage{amssymb}
\usepackage{tikz}
\usepackage{algpseudocode}
\usepackage{algorithm}
\usepackage{mathrsfs}
\usepackage[top=0.5in, left=.5in, bottom=.5in, right=.5in]{geometry}

\renewenvironment{quote}{%
  \list{}{%
    \leftmargin3.7mm   
    \rightmargin\leftmargin
  }
  \item\relax\it
}
{\endlist}

\newtheorem{theorem}{\bf{Theorem}}
\newtheorem{lemma}{\bf{Lemma}}
\newtheorem{proposition}{\bf{Proposition}}

\newtheorem{definition}{\bf{Definition}}
\newtheorem{remark}{\bf{Remark}}
\newtheorem{example}{\bf{Example}}
\def\QED{~\rule[-1pt]{5pt}{5pt}\par\medskip}
\newenvironment{proof}{\emph{Proof.}}{\hfill\QED}

\newcommand{\argmin}{\operatorname{argmin}}
\newcommand{\diag}{\operatorname{diag}}
\newcommand{\tr}{\operatorname{tr}}
\newcommand{\st}{\operatorname{s.t.}}

\newcommand{\rank}{\operatorname{rank}}
\renewcommand{\dim}{\operatorname{dim}}

\renewcommand{\P}{\mathcal{P}}
\renewcommand{\S}{\mathcal{S}}
\newcommand{\A}{\boldsymbol{A}}
\newcommand{\B}{\boldsymbol{B}}
\newcommand{\C}{\boldsymbol{C}}

\newcommand{\edge}{\mathcal{E}}
\newcommand{\E}{\boldsymbol{E}}
\newcommand{\J}{\boldsymbol{J}}
\newcommand{\G}{\mathcal{G}}
\newcommand{\U}{\boldsymbol{U}}
\newcommand{\V}{\mathcal{V}}
\newcommand{\W}{\boldsymbol{W}}

\newcommand{\Y}{\boldsymbol{Y}}
\newcommand{\Z}{\boldsymbol{Z}}
\newcommand{\I}{\boldsymbol{I}}
\renewcommand{\a}{\boldsymbol{a}}
\renewcommand{\b}{\boldsymbol{b}}
\renewcommand{\c}{\boldsymbol{c}}

\newcommand{\e}{\boldsymbol{e}}

\newcommand{\x}{\boldsymbol{x}}
\newcommand{\y}{\boldsymbol{y}}
\newcommand{\z}{\boldsymbol{z}}

\renewcommand{\u}{\boldsymbol{u}}
\renewcommand{\v}{\boldsymbol{v}}
\newcommand{\1}{\boldsymbol{1}} 
\newcommand{\0}{\boldsymbol{0}} 
\renewcommand{\Re}{\mathbb{R}}
\newcommand{\ie}{\text{i.e.}}
\newcommand{\eg}{\text{e.g.}}
\newcommand{\deltab}{\boldsymbol{\delta}}
\newcommand{\Deltab}{\boldsymbol{\Delta}}
\newcommand{\myparagraph}[1]{\medskip\noindent\textbf{#1.} }

\def\L{\boldsymbol{\cal L}}

\ifCLASSOPTIONcompsoc
\else
\fi
\ifCLASSINFOpdf
\else
\fi
\begin{document}
%
\title{Sparse Subspace Clustering:\\Algorithm, Theory, and Applications}
%
%
%
%

\author{Ehsan~Elhamifar,~\IEEEmembership{Student Member,~IEEE,}
        and~Ren\'e~Vidal,~\IEEEmembership{Senior Member,~IEEE}
\IEEEcompsocitemizethanks{\IEEEcompsocthanksitem E. Elhamifar is with the Department
of Electrical Engineering and Computer Science, University of California, Berkeley, USA. E-mail: ehsan@eecs.berkeley.edu.
\IEEEcompsocthanksitem R. Vidal is with the Center for Imaging Science and the Department of Biomedical
Engineering, The Johns Hopkins University, USA. E-mail: rvidal@cis.jhu.edu.}
\thanks{}}
\IEEEcompsoctitleabstractindextext{%
\begin{abstract}
Many real-world problems deal with collections of high-dimensional data, such as images, videos, text and web documents, DNA microarray data, and more. Often, such high-dimensional data lie close to low-dimensional structures corresponding to several classes or categories to which the data belong. In this paper, we propose and study an algorithm, called Sparse Subspace Clustering (SSC), to cluster data points that lie in a union of low-dimensional subspaces. The key idea is that, among the infinitely many possible representations of a data point in terms of other points, a sparse representation corresponds to selecting a few points from the same subspace. This motivates solving a sparse optimization program whose solution is used in a spectral clustering framework to infer the clustering of the data into subspaces. Since solving the sparse optimization program is in general NP-hard, we consider a convex relaxation and show that, under appropriate conditions on the arrangement of the subspaces and the distribution of the data, the proposed minimization program succeeds in recovering the desired sparse representations. 
The proposed algorithm is efficient and can handle data points near the intersections of subspaces. Another key advantage of the proposed algorithm with respect to the state of the art is that it can deal directly with data nuisances, such as noise, sparse outlying entries, and missing entries, by incorporating the model of the data into the sparse optimization program. We demonstrate the effectiveness of the proposed algorithm through experiments on synthetic data as well as the two real-world problems of motion segmentation and face clustering.

\end{abstract}

}

\maketitle

\IEEEdisplaynotcompsoctitleabstractindextext

%
\IEEEpeerreviewmaketitle

\section{Introduction}
%
%

%
%
%
%

%
\begin{figure*}[t]
\centering
\includegraphics[width=0.132\linewidth, trim = 65 40 65 20 , clip]{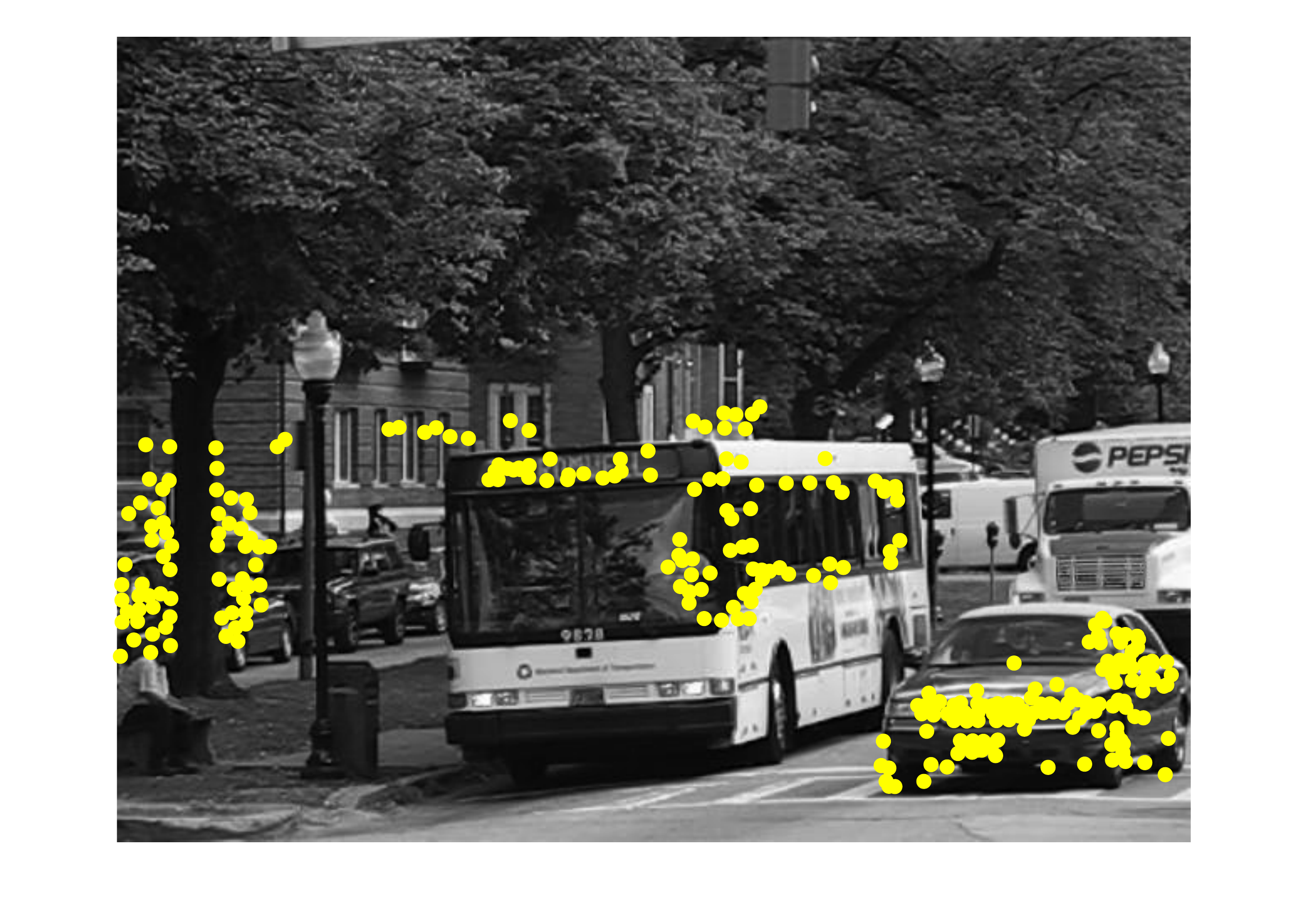}\hspace{0.5mm}
\includegraphics[width=0.132\linewidth, trim = 65 40 65 20 , clip]{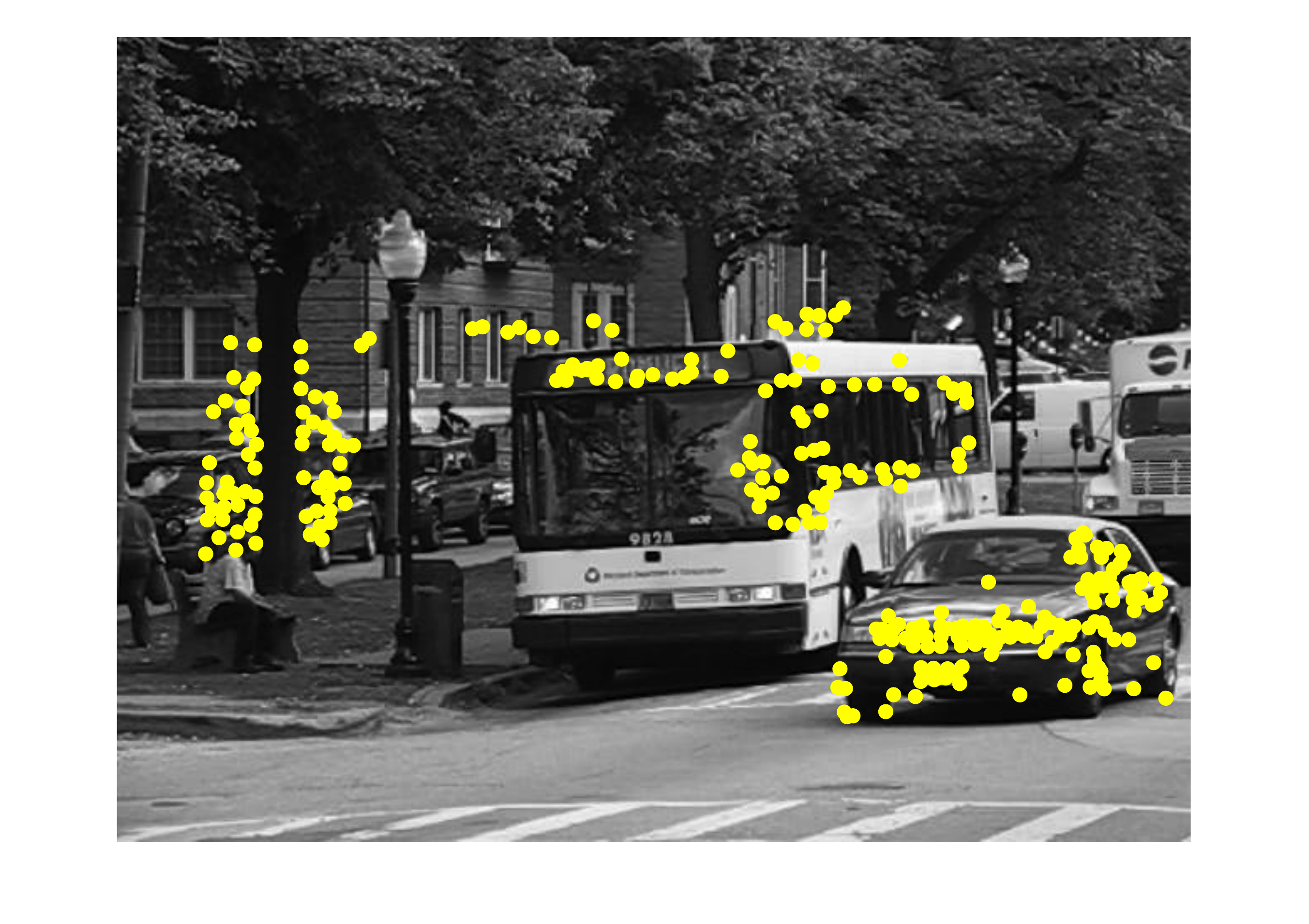}\hspace{0.5mm}
\includegraphics[width=0.132\linewidth, trim = 65 40 65 20 , clip]{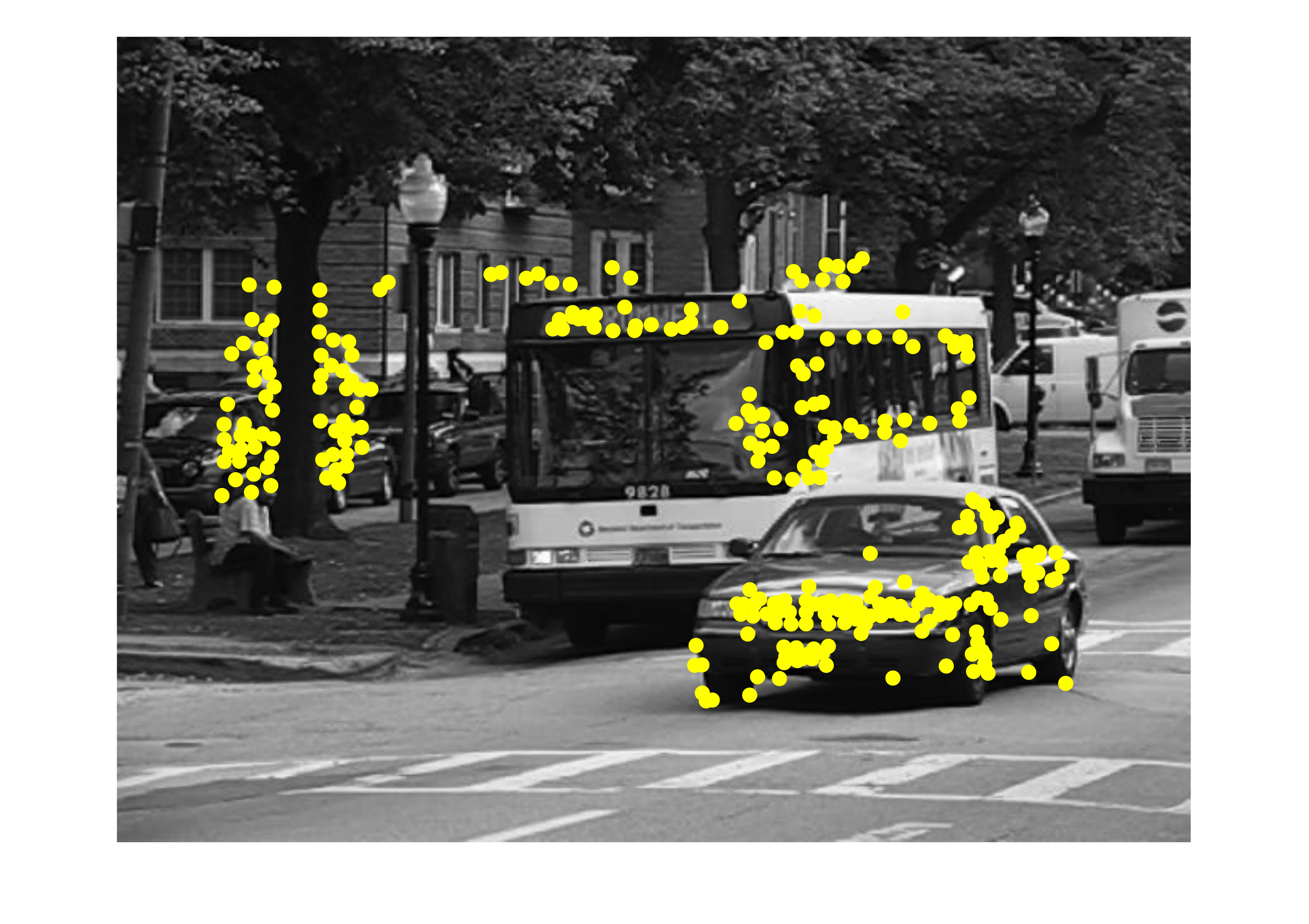}\hspace{0.5mm}
\includegraphics[width=0.132\linewidth, trim = 65 40 65 20 , clip]{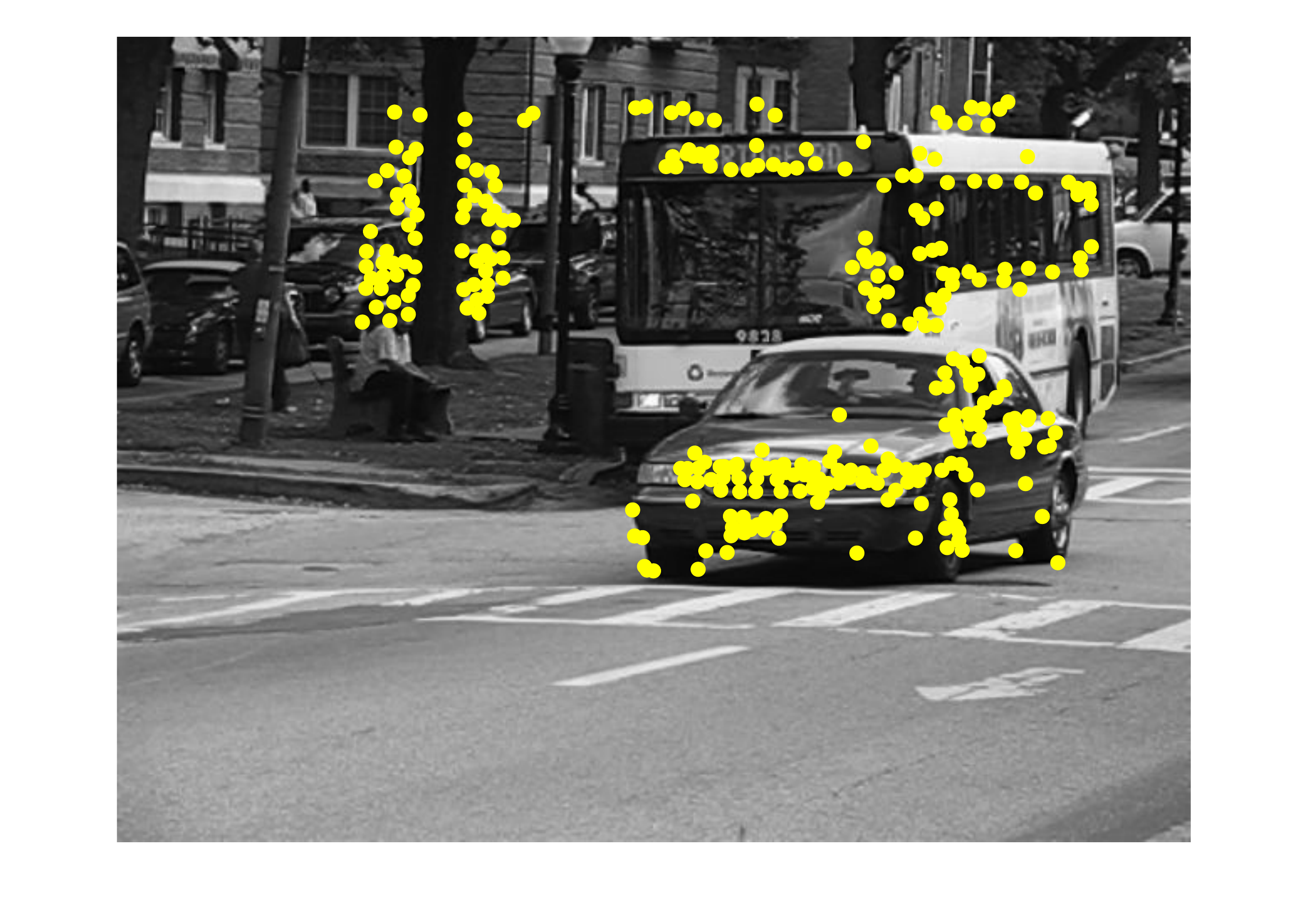}\hspace{0.5mm}
\includegraphics[width=0.132\linewidth, trim = 65 40 65 20 , clip]{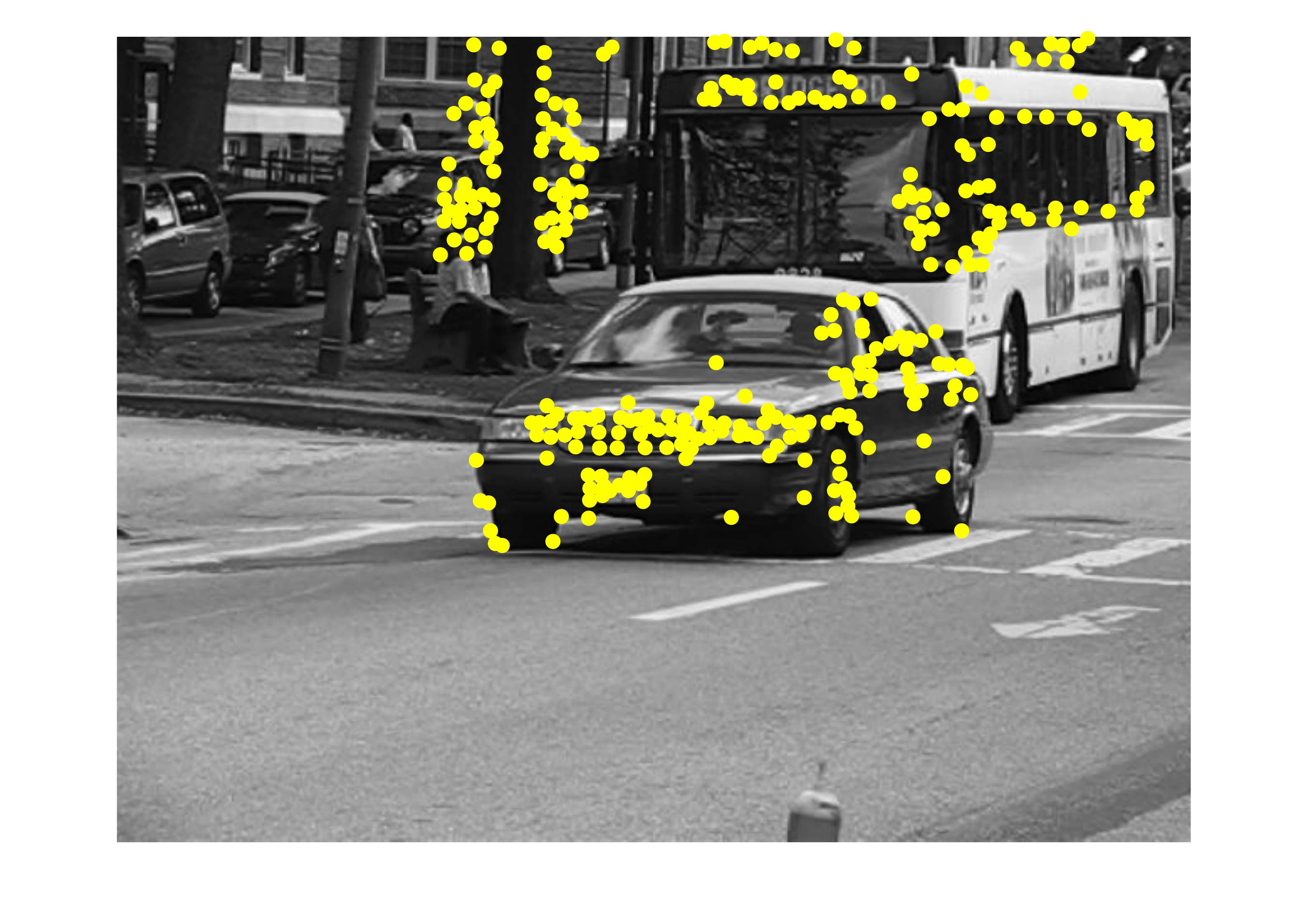}\hspace{0.5mm}
\includegraphics[width=0.132\linewidth, trim = 65 40 65 20 , clip]{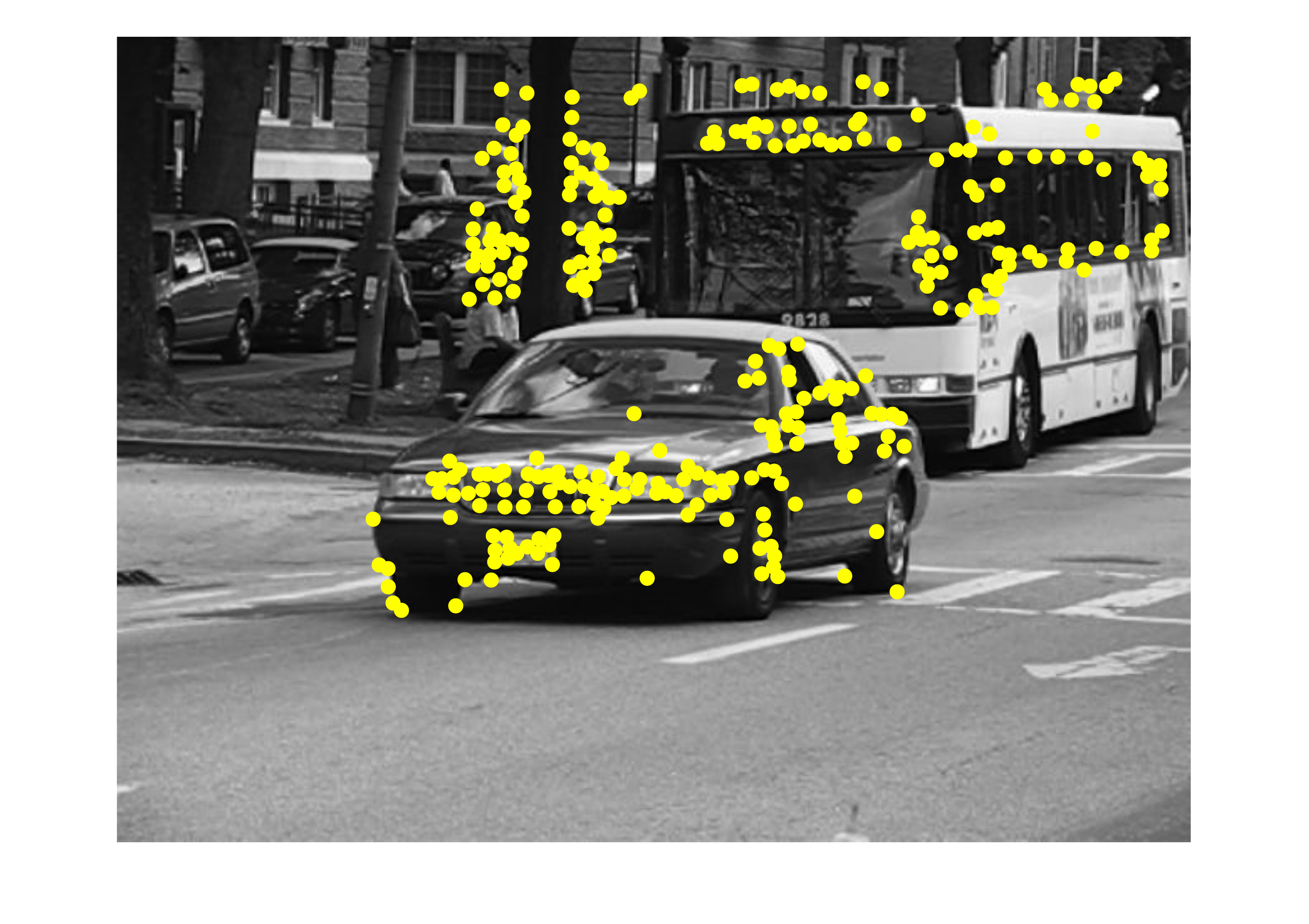}\hspace{0.5mm}
\includegraphics[width=0.132\linewidth, trim = 65 40 65 20 , clip]{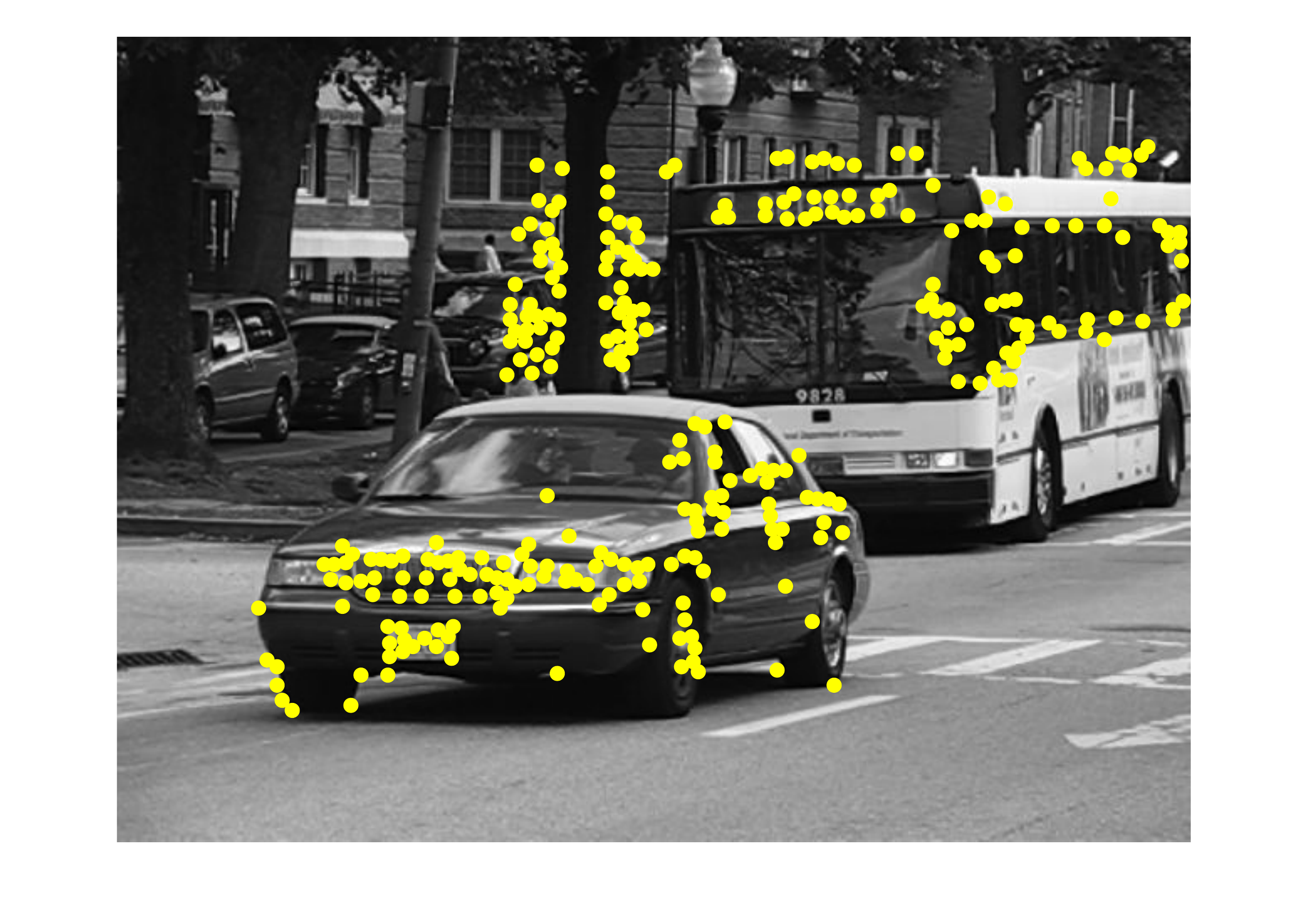}\hspace{0.5mm}
\\ \vspace{0.5mm} \
\includegraphics[width=0.132\linewidth, trim = 65 40 65 20 , clip]{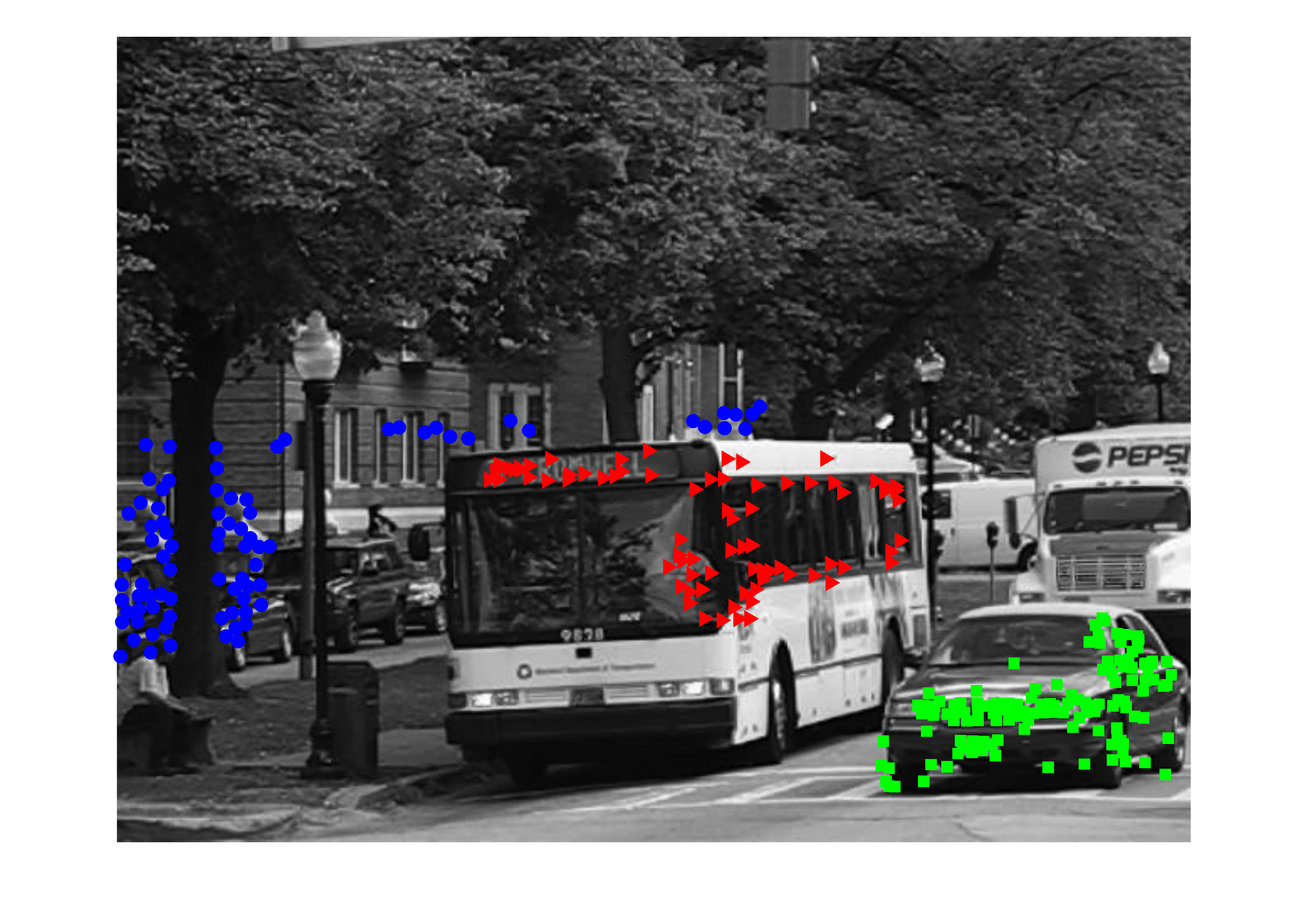}\hspace{0.5mm}
\includegraphics[width=0.132\linewidth, trim = 65 40 65 20 , clip]{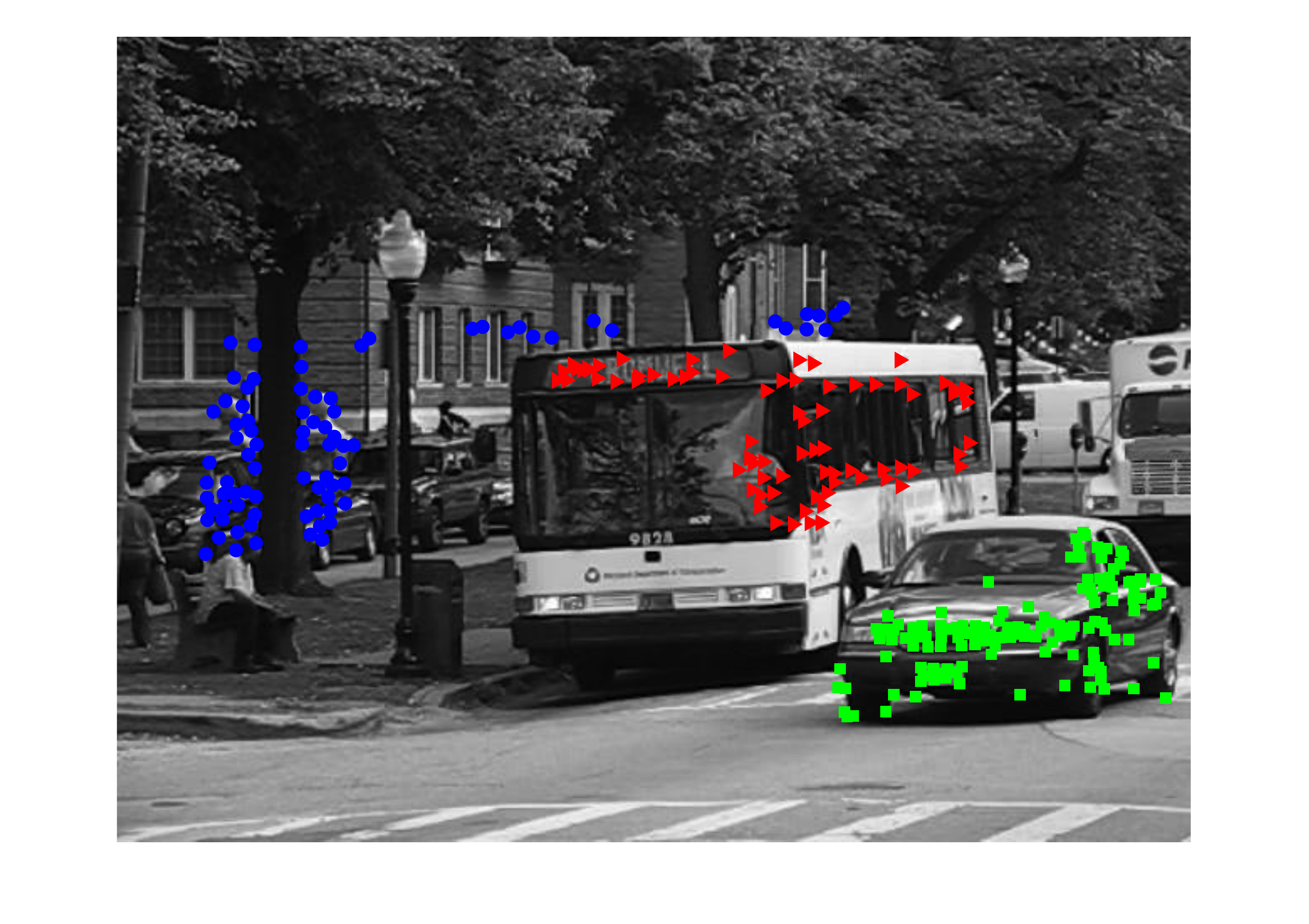}\hspace{0.5mm}
\includegraphics[width=0.132\linewidth, trim = 65 40 65 20 , clip]{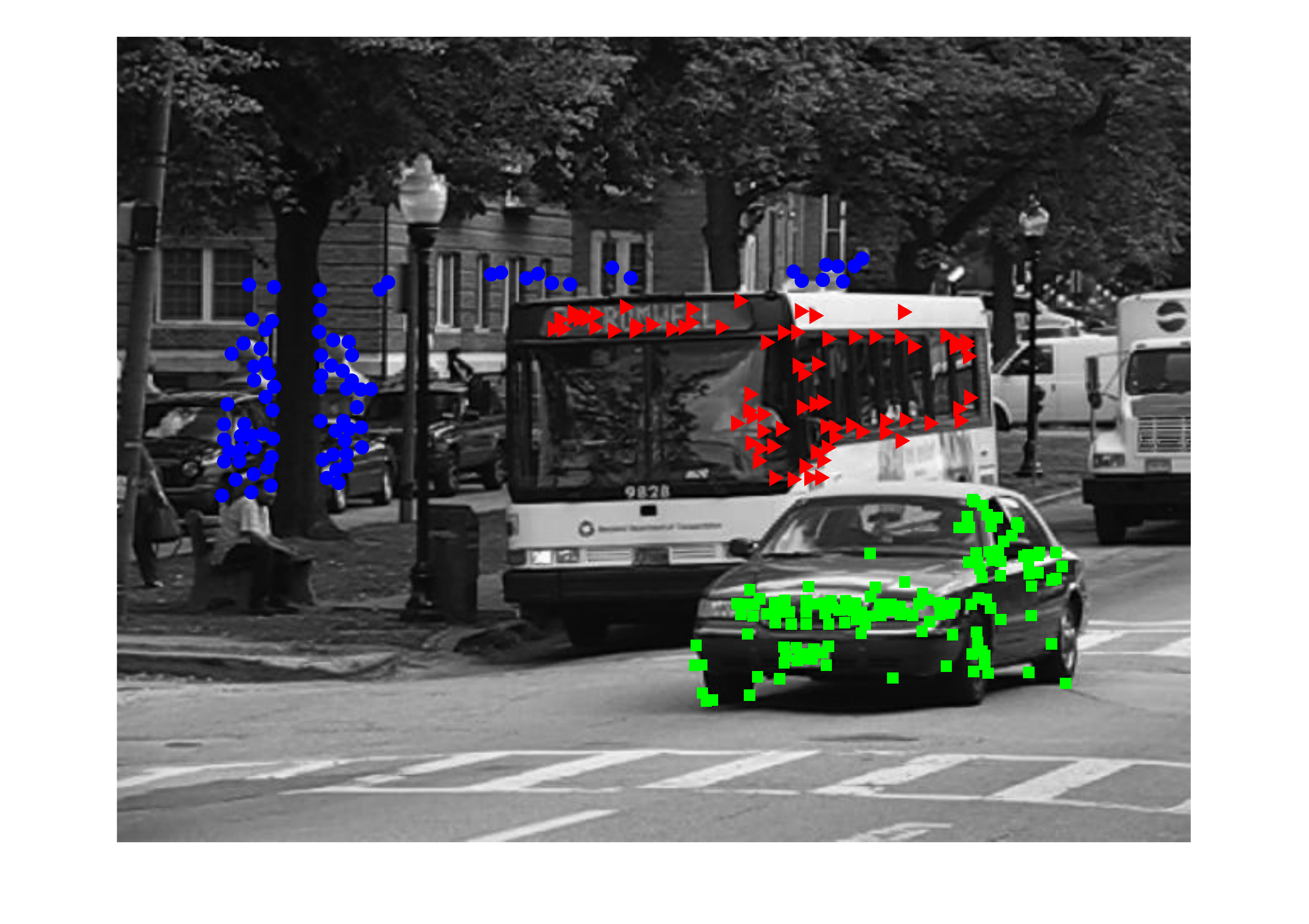}\hspace{0.5mm}
\includegraphics[width=0.132\linewidth, trim = 65 40 65 20 , clip]{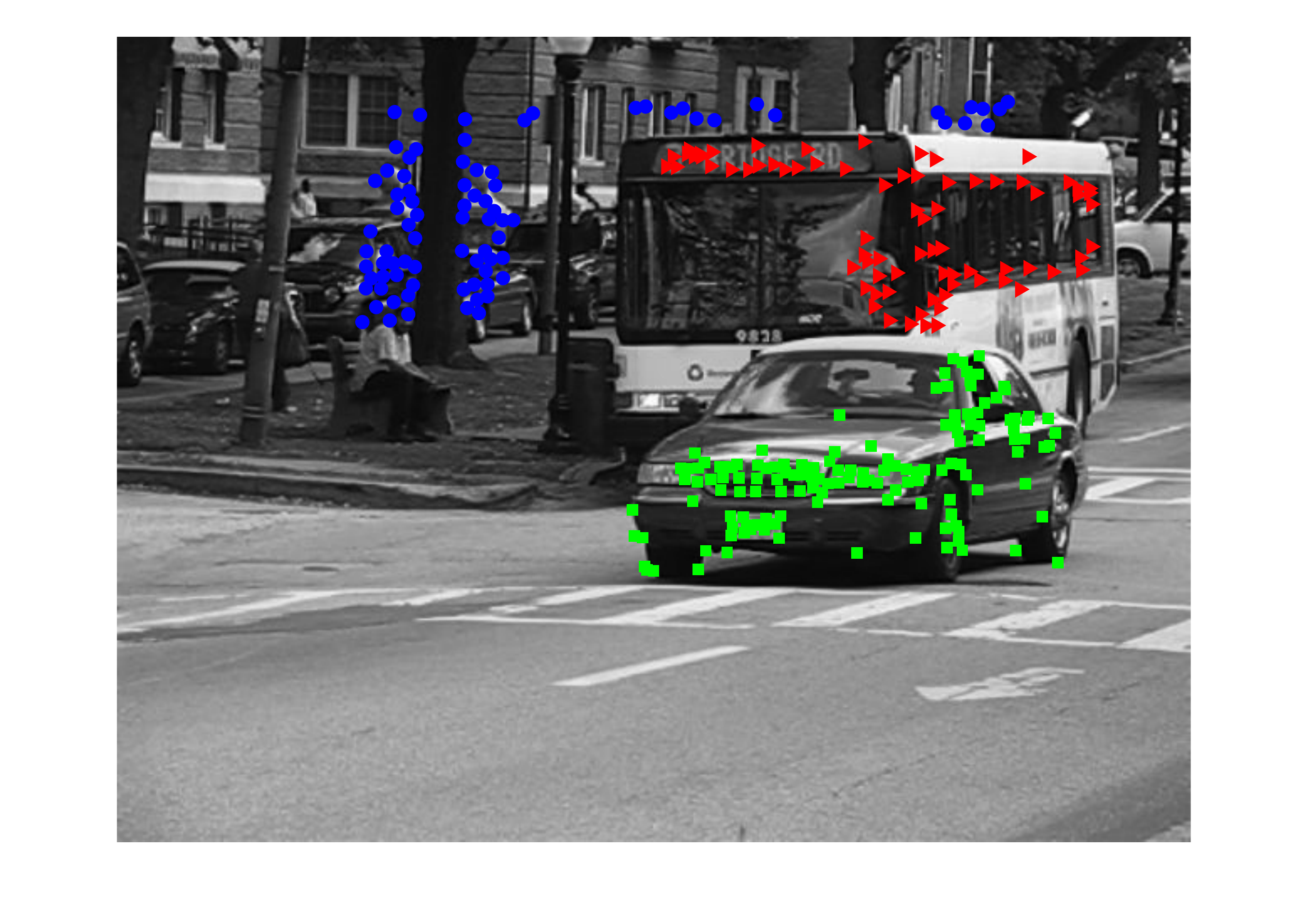}\hspace{0.5mm}
\includegraphics[width=0.132\linewidth, trim = 65 40 65 20 , clip]{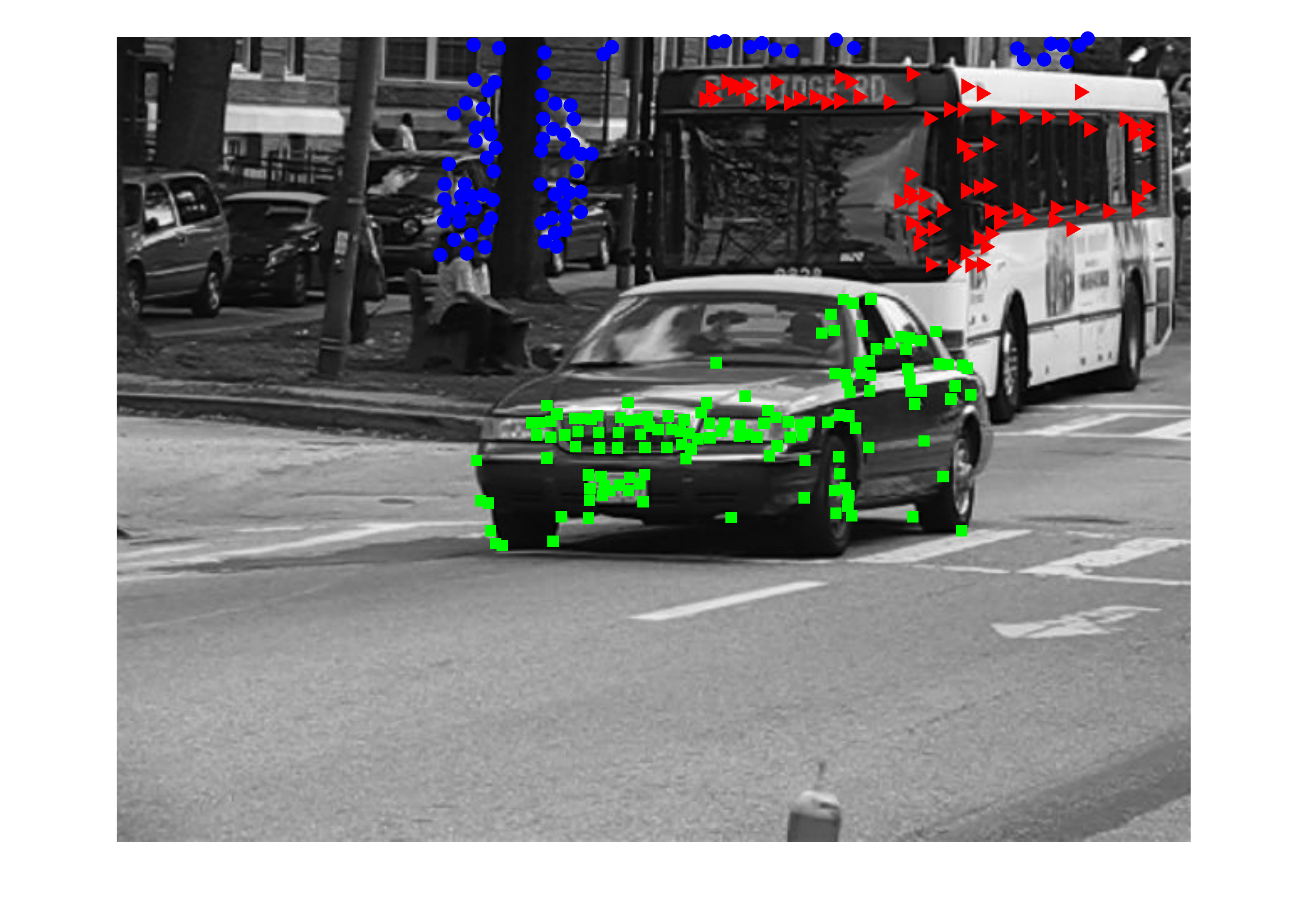}\hspace{0.5mm}
\includegraphics[width=0.132\linewidth, trim = 65 40 65 20 , clip]{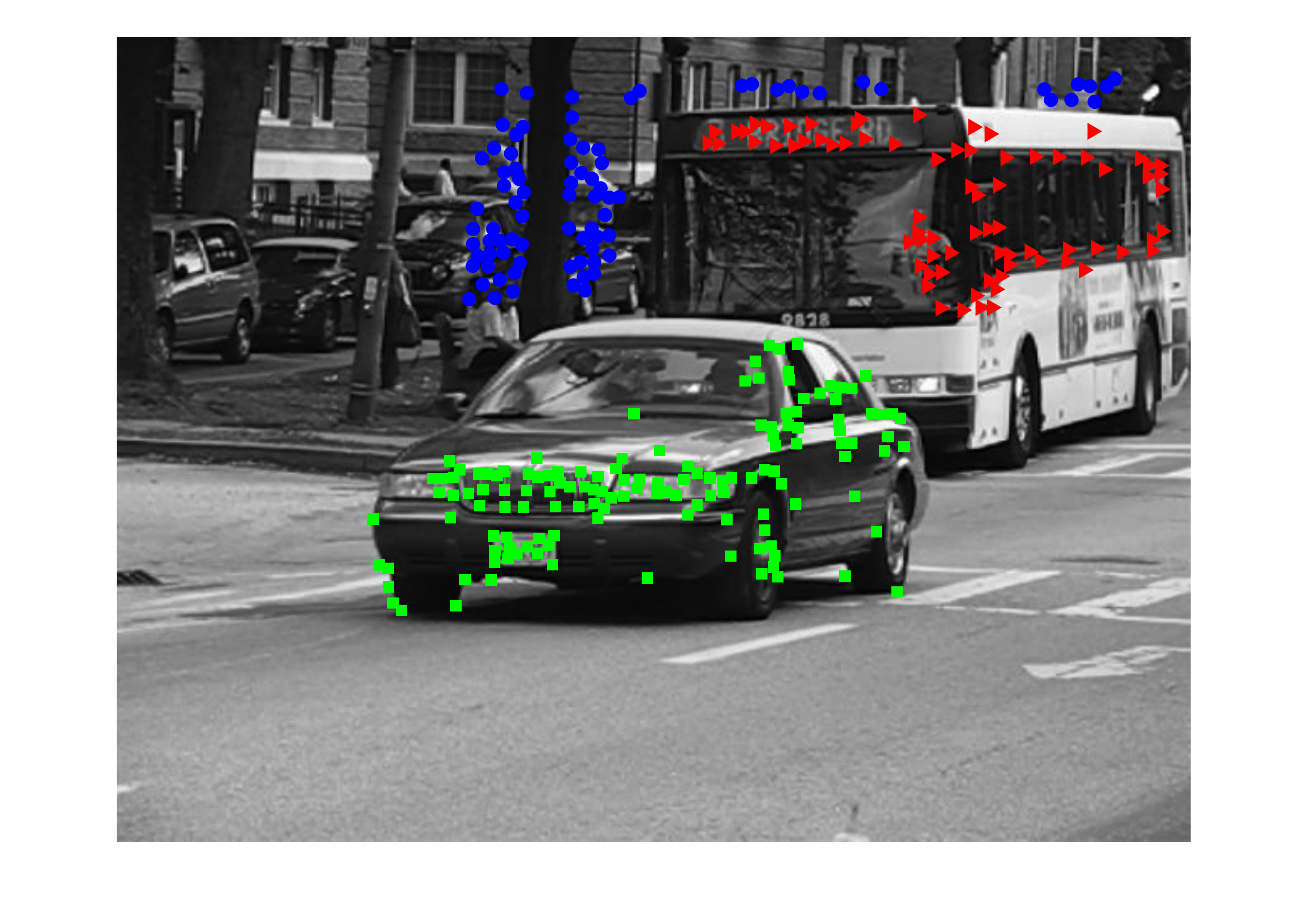}\hspace{0.5mm}
\includegraphics[width=0.132\linewidth, trim = 65 40 65 20 , clip]{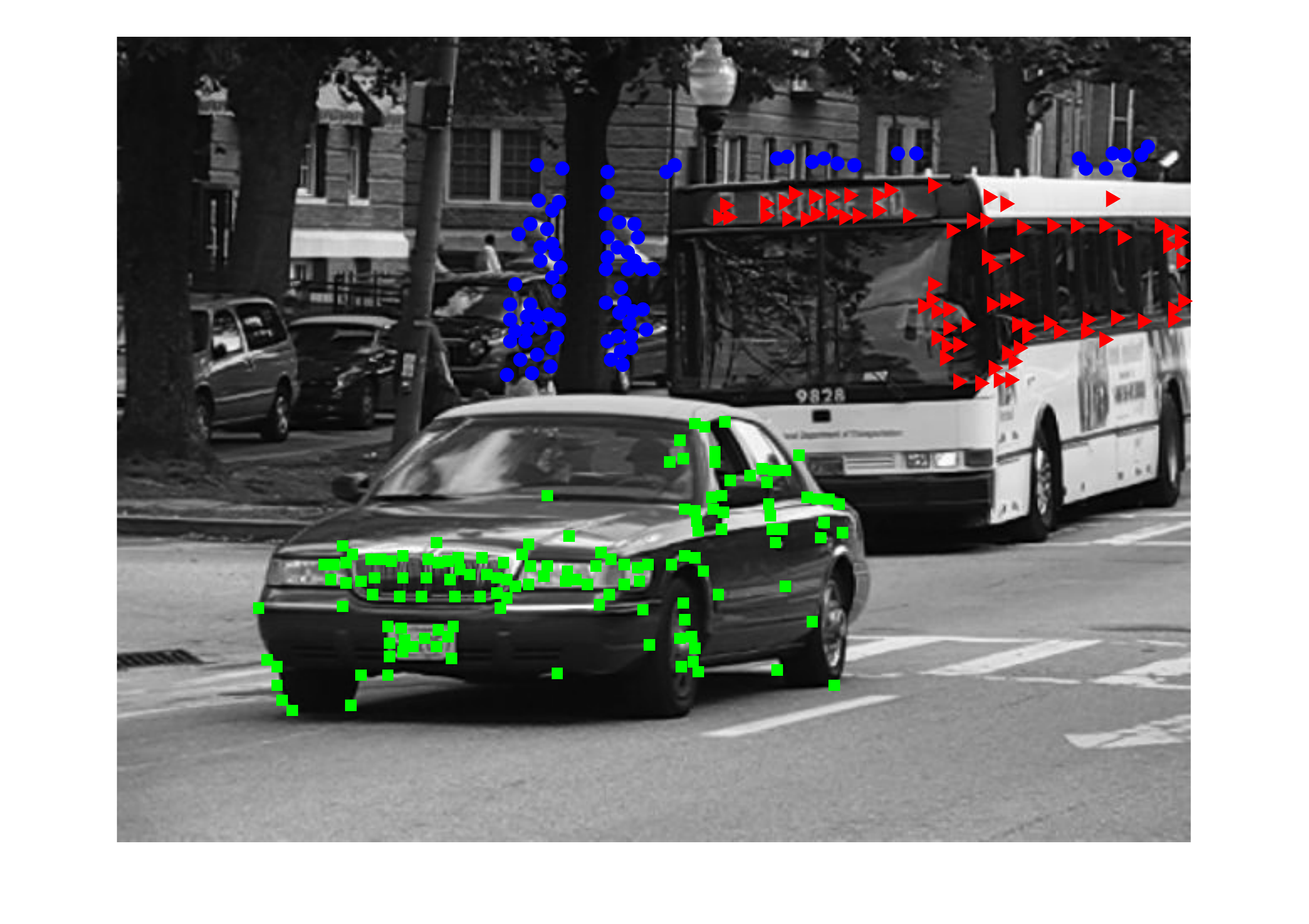}\hspace{0.5mm}
\vspace{-1.5mm}
\caption{\small{Motion segmentation: given feature points on multiple rigidly moving objects tracked in multiple frames of a video (top), the goal is to separate the feature trajectories according to the moving objects (bottom).}}
\label{fig:example-MS}
\end{figure*}
\begin{figure*}[t]
\centering
\includegraphics[width=0.054\linewidth, trim = 70 48 70 22 , clip]{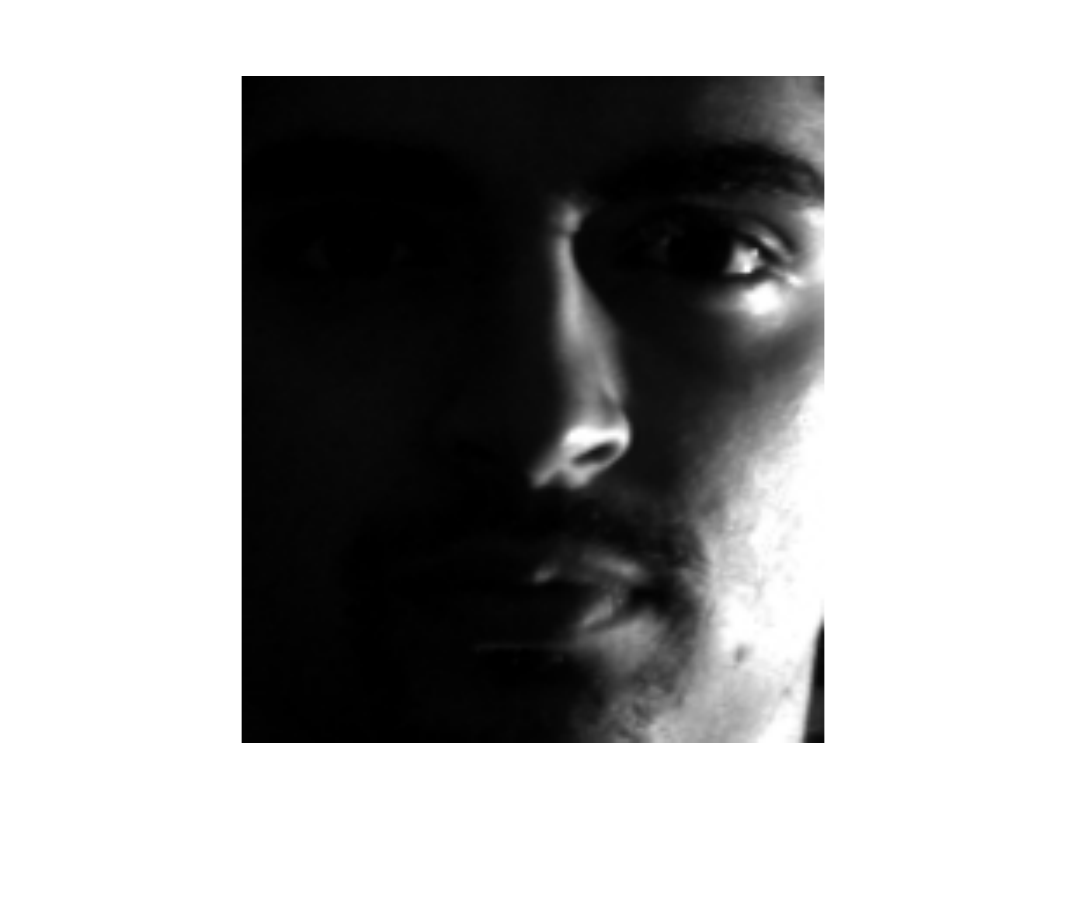}\hspace{1.1mm}
\includegraphics[width=0.054\linewidth, trim = 70 48 70 22 , clip]{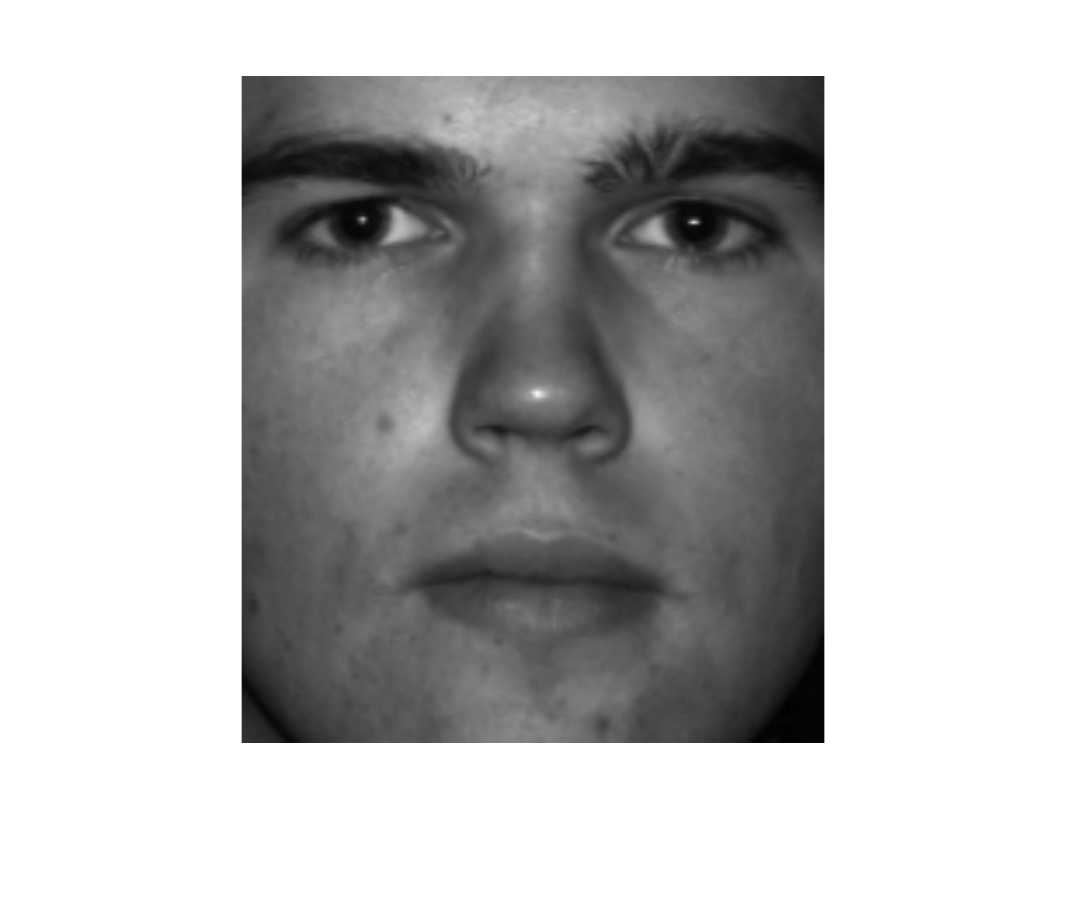}\hspace{1.1mm}
\includegraphics[width=0.054\linewidth, trim = 70 48 70 22 , clip]{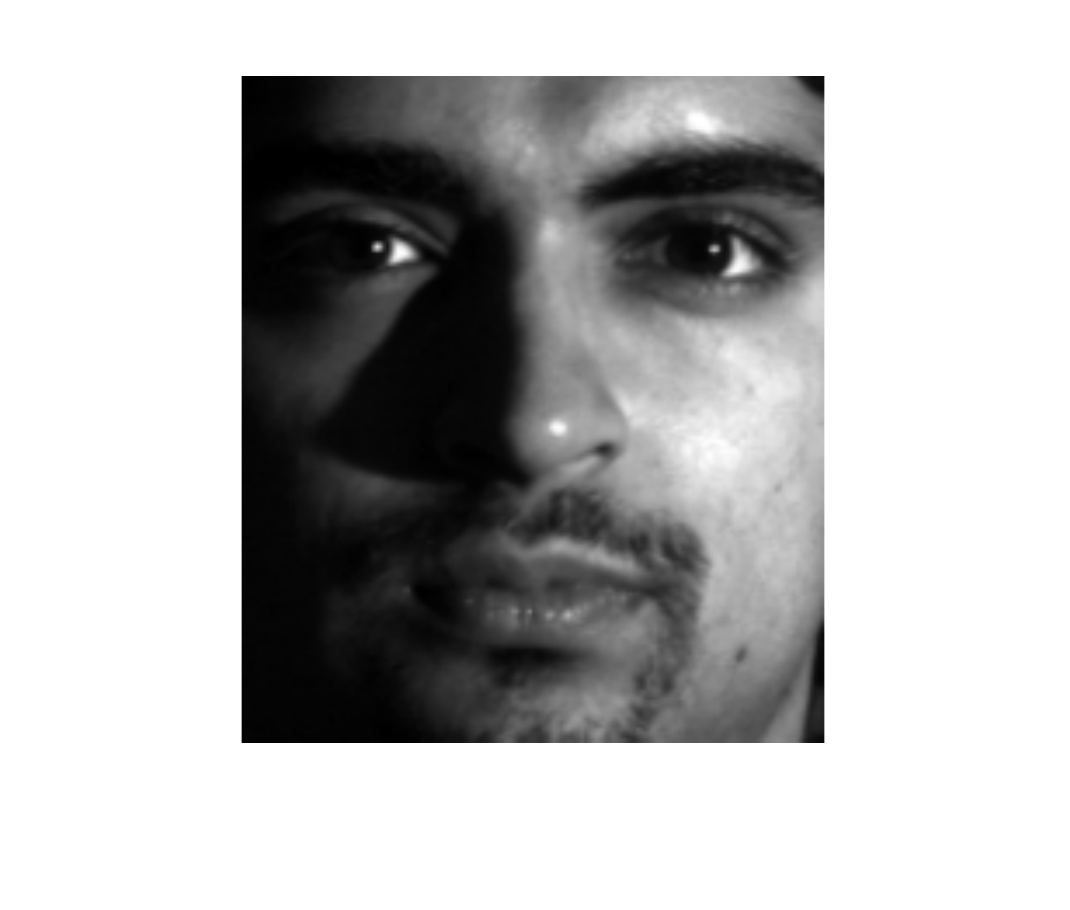}\hspace{1.1mm}
\includegraphics[width=0.054\linewidth, trim = 70 48 70 22 , clip]{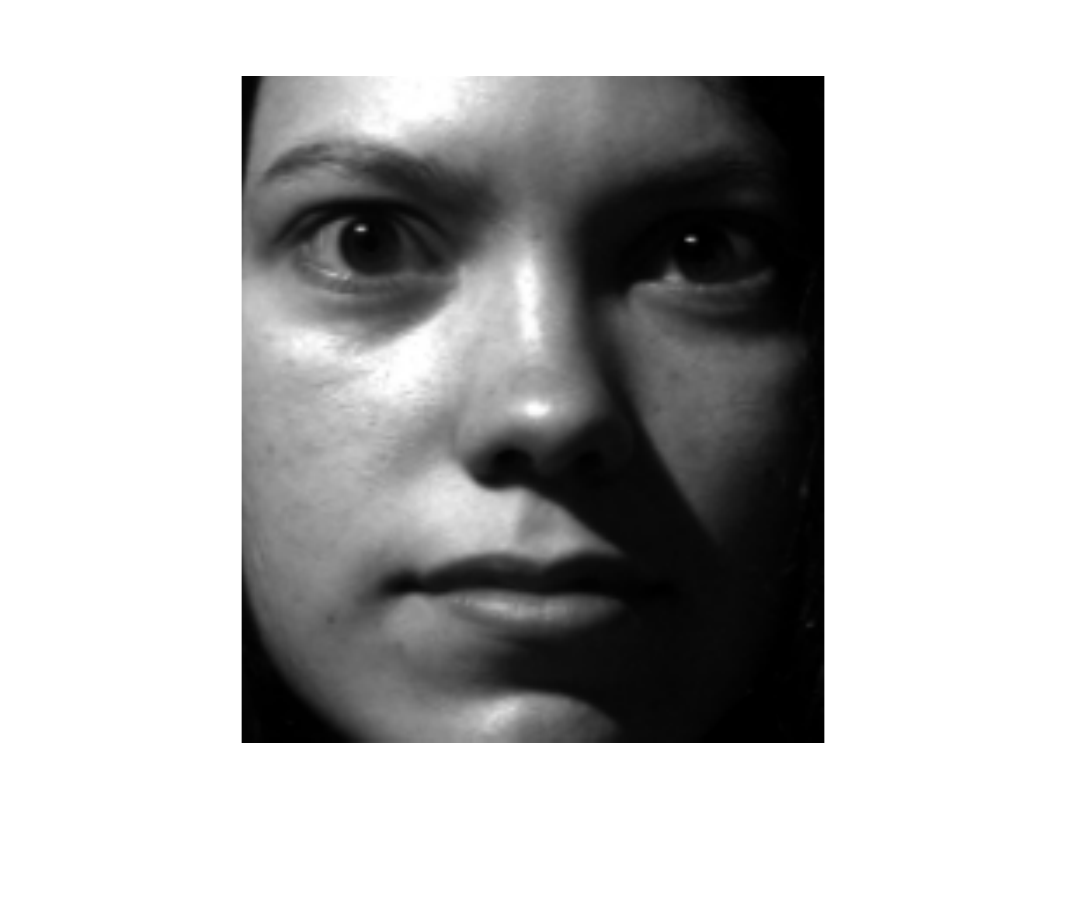}\hspace{1.1mm}
\includegraphics[width=0.054\linewidth, trim = 70 48 70 22 , clip]{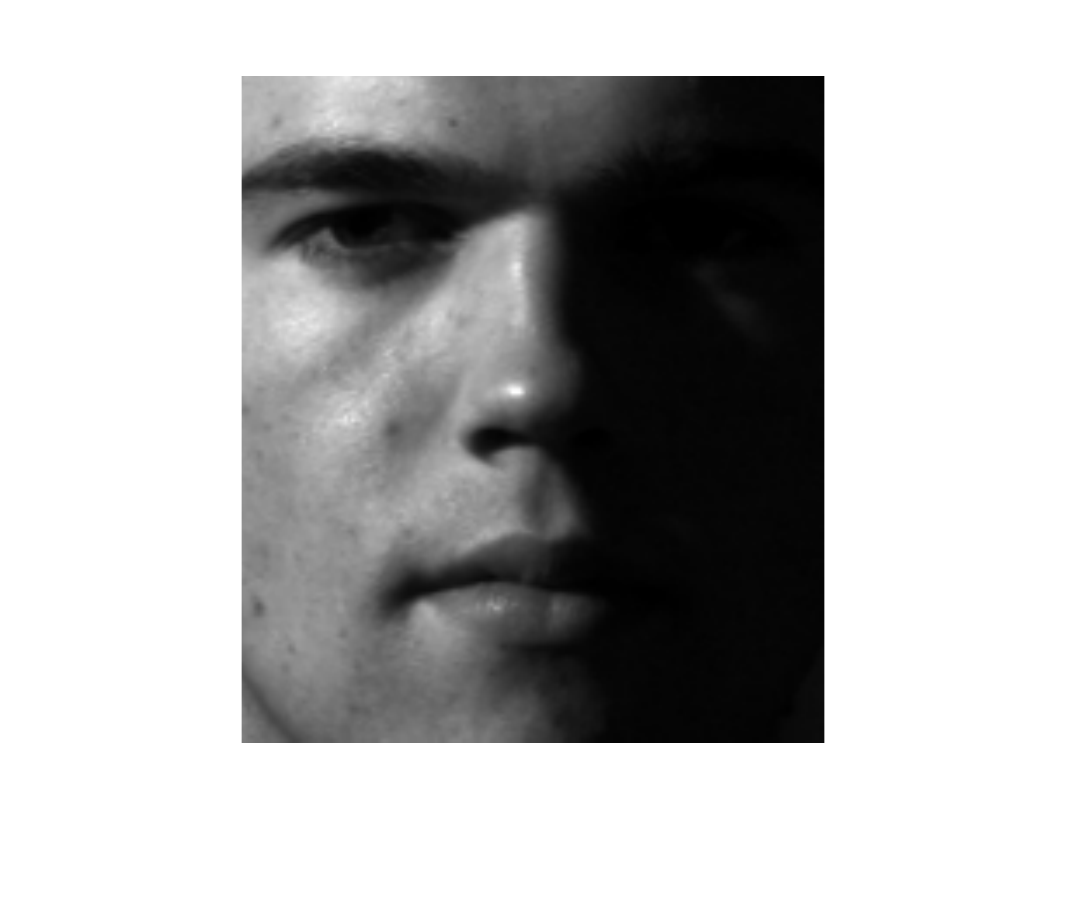}\hspace{1.1mm}
\includegraphics[width=0.054\linewidth, trim = 70 48 70 22 , clip]{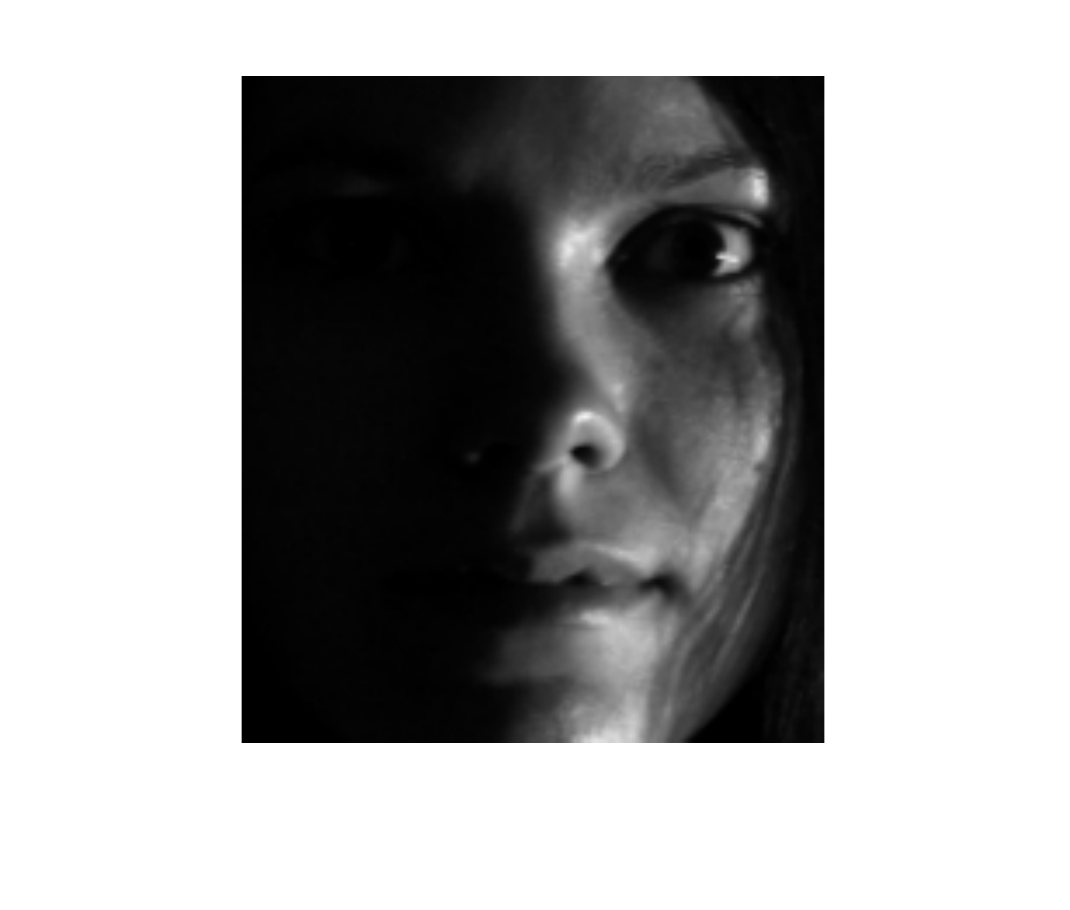}\hspace{1.1mm}
\includegraphics[width=0.054\linewidth, trim = 70 48 70 22 , clip]{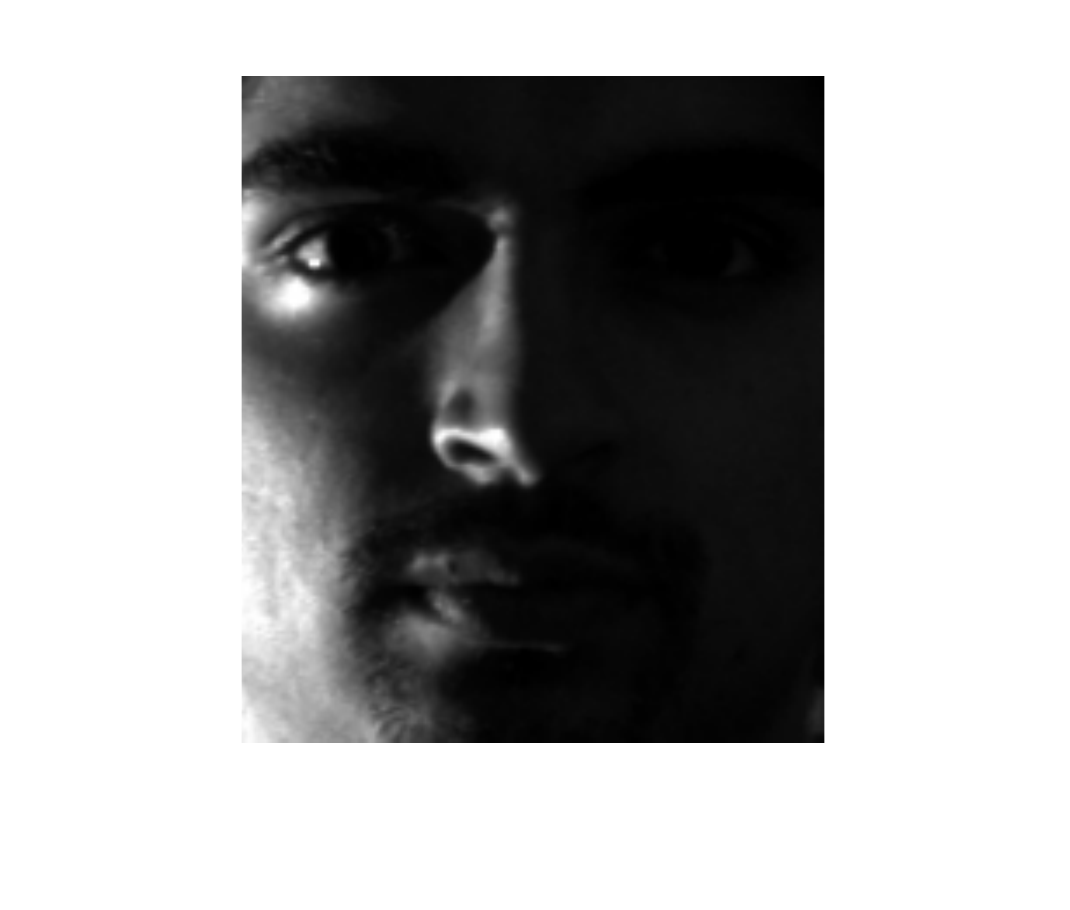}\hspace{1.1mm}
\includegraphics[width=0.054\linewidth, trim = 70 48 70 22 , clip]{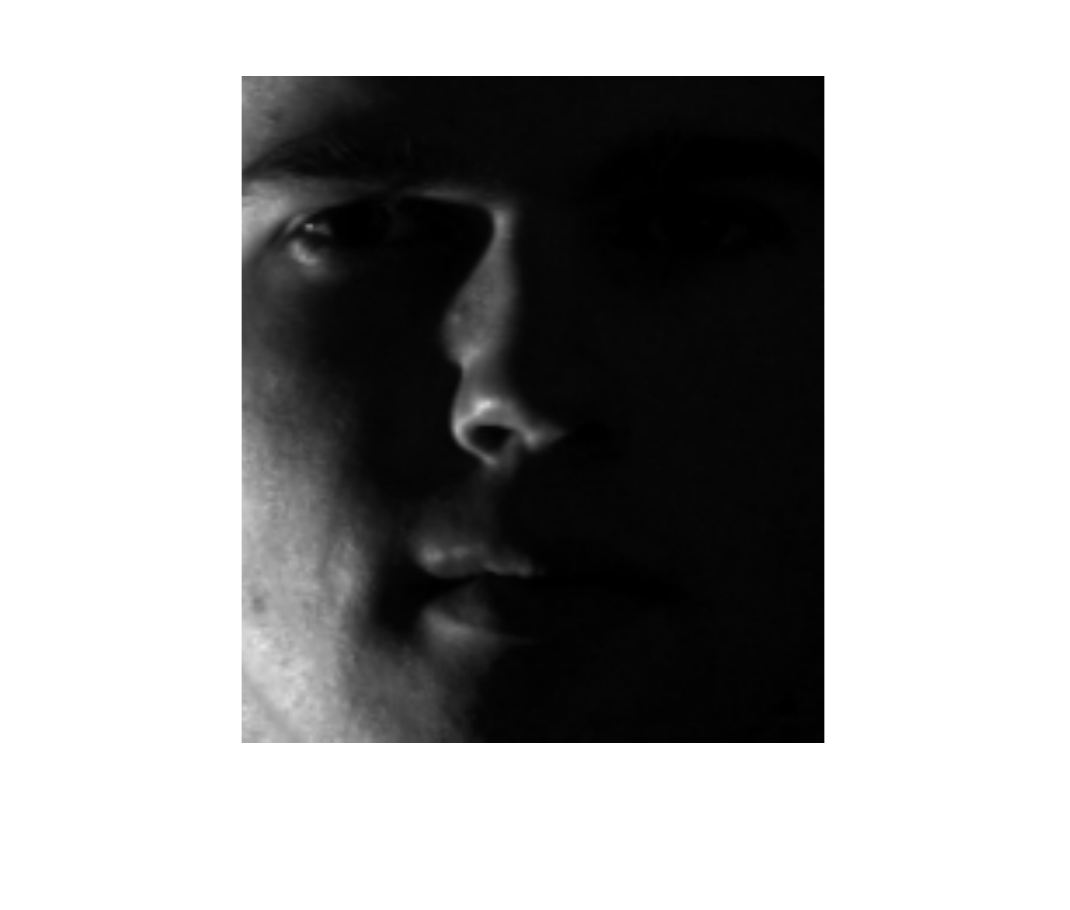}\hspace{1.1mm}
\includegraphics[width=0.054\linewidth, trim = 70 48 70 22 , clip]{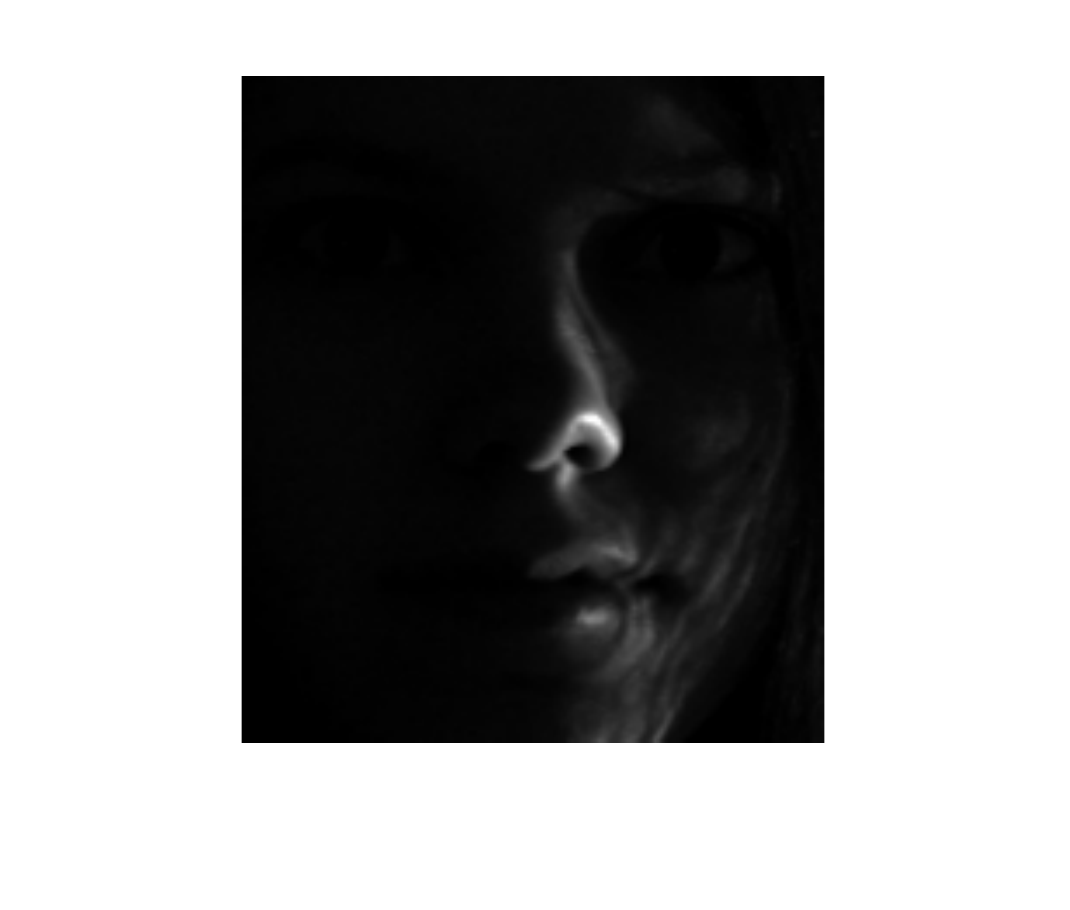}\hspace{1.1mm}
\includegraphics[width=0.054\linewidth, trim = 70 48 70 22 , clip]{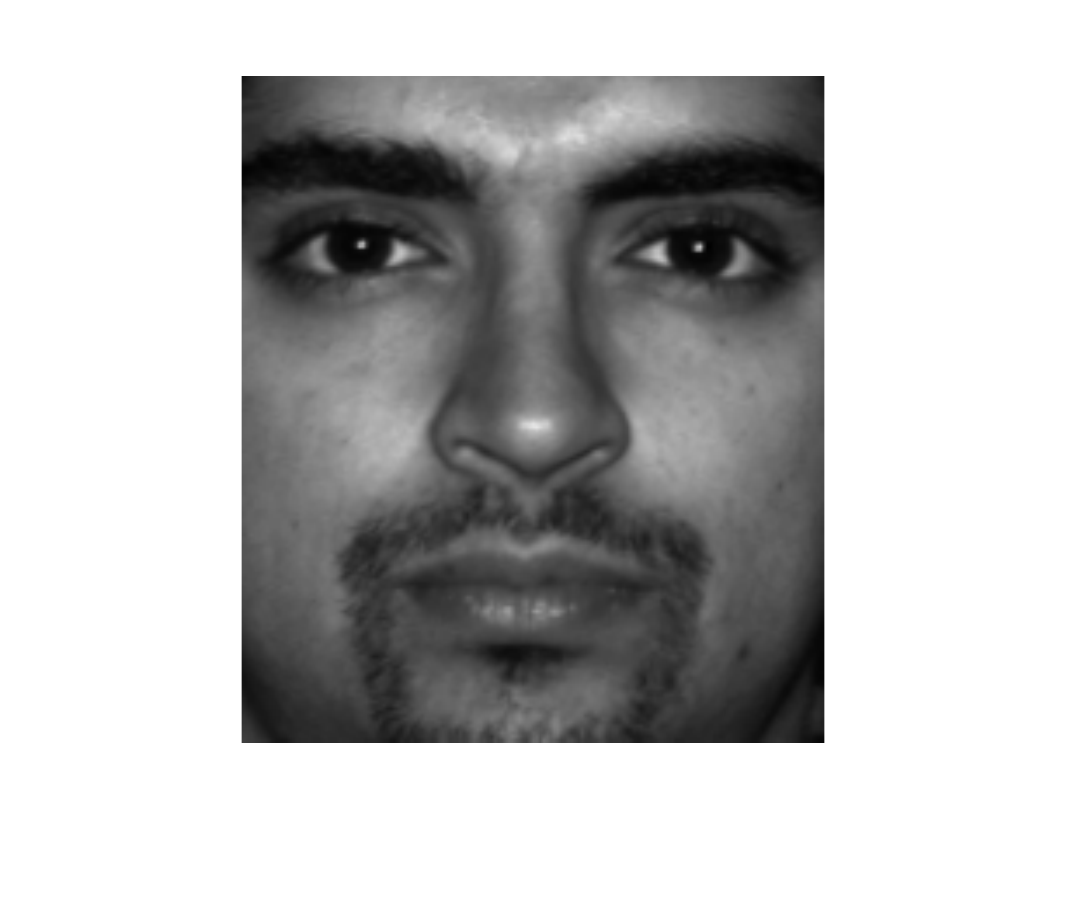}\hspace{1.1mm}
\includegraphics[width=0.054\linewidth, trim = 70 48 70 22 , clip]{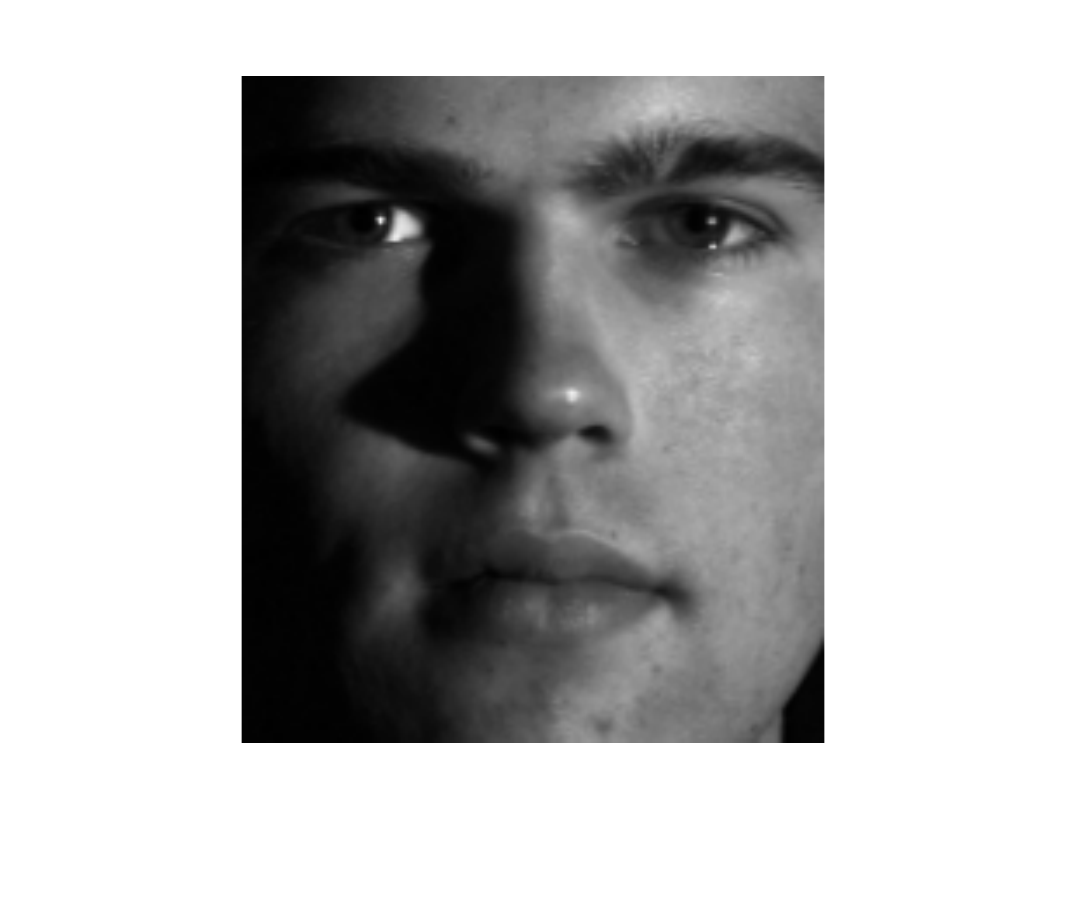}\hspace{1.1mm}
\includegraphics[width=0.054\linewidth, trim = 70 48 70 22 , clip]{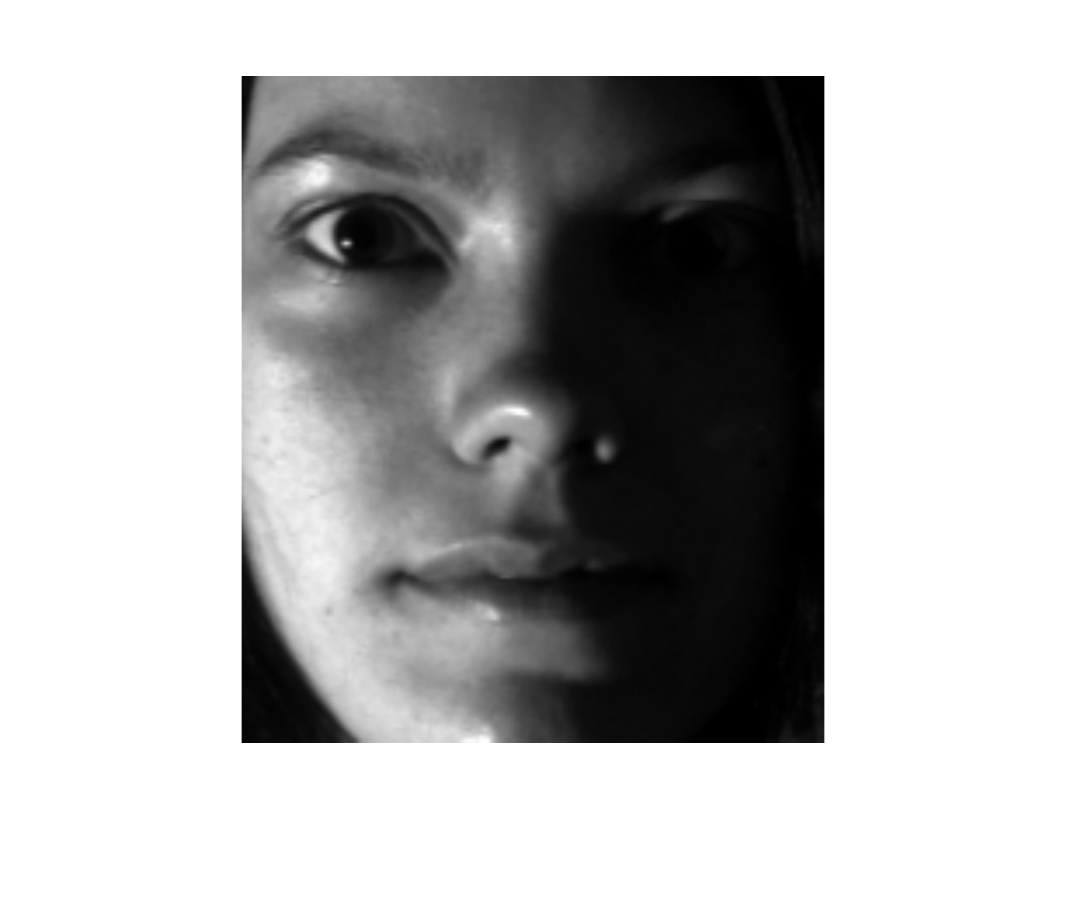}\hspace{1.1mm}
\includegraphics[width=0.054\linewidth, trim = 70 48 70 22 , clip]{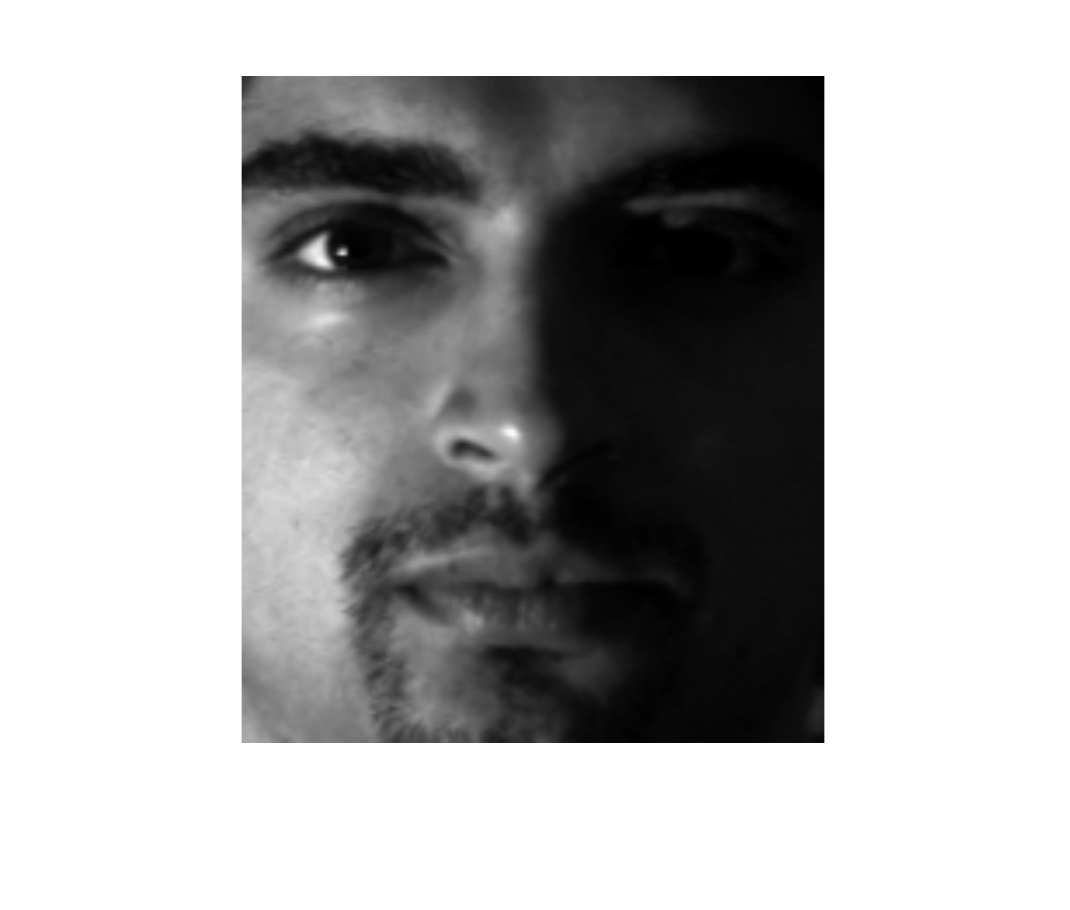}\hspace{1.1mm}
\includegraphics[width=0.054\linewidth, trim = 70 48 70 22 , clip]{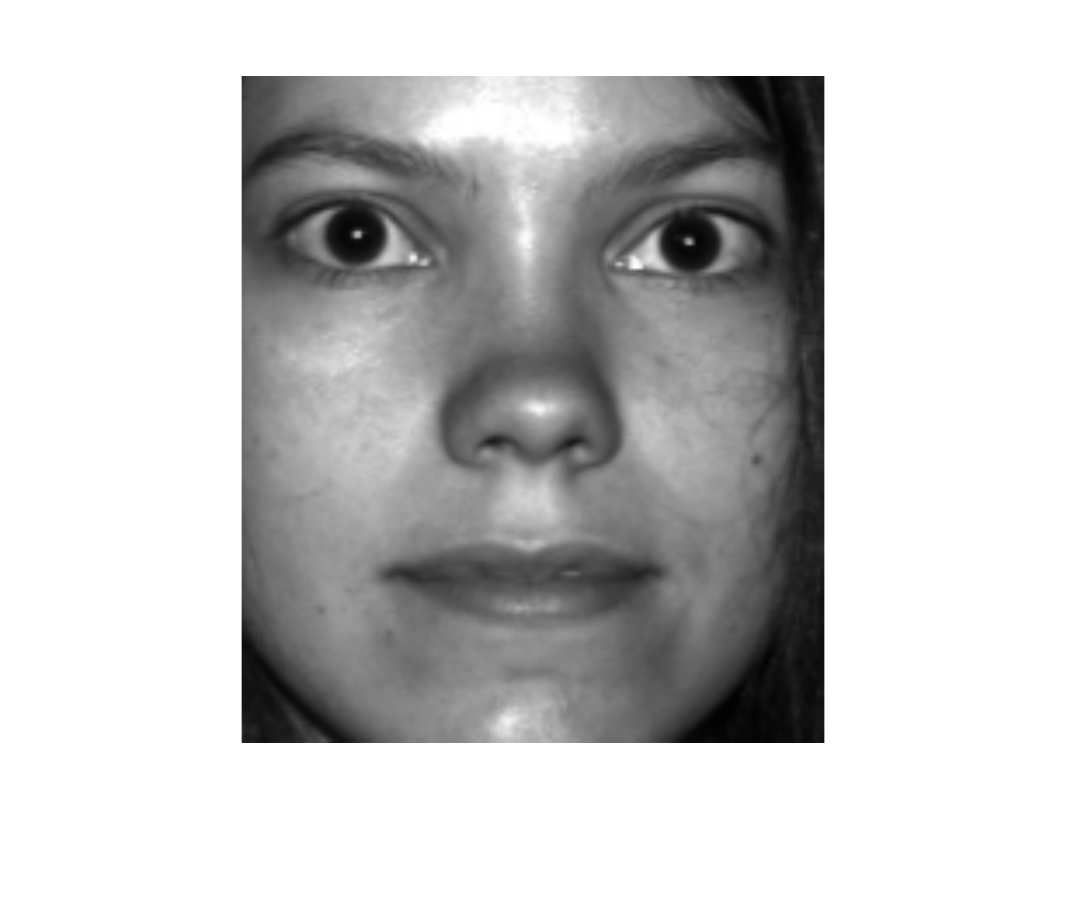}\hspace{1.1mm}
\includegraphics[width=0.054\linewidth, trim = 70 48 70 22 , clip]{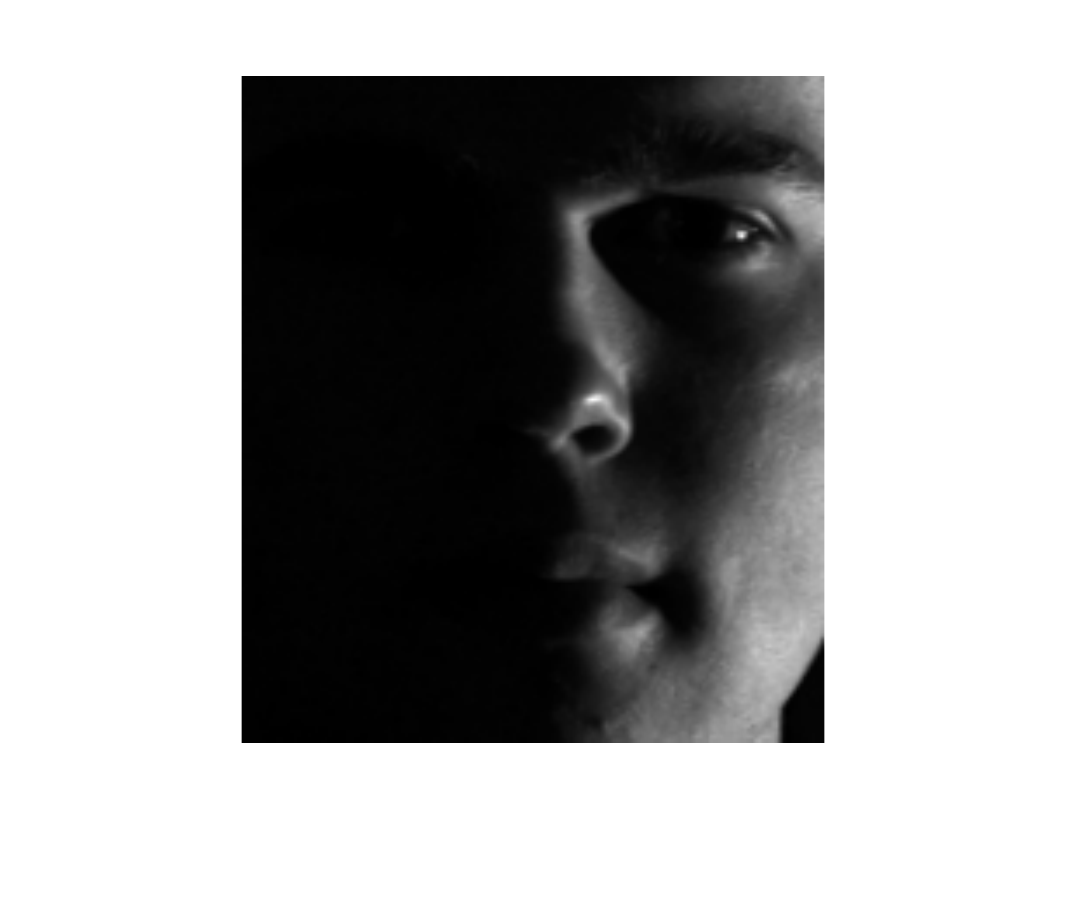}
\\ \vspace{.7mm}
\hspace{1.5mm}\colorbox{red}{\includegraphics[width=0.053\linewidth, trim = 70 48 70 22 , clip]{I09_1}\hspace{-1.5mm}
\includegraphics[width=0.053\linewidth, trim = 70 48 70 22 , clip]{I09_62}\hspace{-1.5mm}
\includegraphics[width=0.053\linewidth, trim = 70 48 70 22 , clip]{I09_23}\hspace{-1.5mm}
\includegraphics[width=0.053\linewidth, trim = 70 48 70 22 , clip]{I09_49}\hspace{-1.5mm}
\includegraphics[width=0.053\linewidth, trim = 70 48 70 22 , clip]{I09_33}}\hspace{1mm}\hfill
\colorbox{green}{\includegraphics[width=0.053\linewidth, trim = 70 48 70 22 , clip]{I28_1}\hspace{-1.5mm}
\includegraphics[width=0.053\linewidth, trim = 70 48 70 22 , clip]{I28_62}\hspace{-1.5mm}
\includegraphics[width=0.053\linewidth, trim = 70 48 70 22 , clip]{I28_17}\hspace{-1.5mm}
\includegraphics[width=0.053\linewidth, trim = 70 48 70 22 , clip]{I28_57}\hspace{-1.5mm}
\includegraphics[width=0.053\linewidth, trim = 70 48 70 22 , clip]{I28_23}}\hspace{1mm}\hfill
\colorbox{blue}{\includegraphics[width=0.053\linewidth, trim = 70 48 70 22 , clip]{I20_1}\hspace{-1.5mm}
\includegraphics[width=0.053\linewidth, trim = 70 48 70 22 , clip]{I20_62}\hspace{-1.5mm}
\includegraphics[width=0.053\linewidth, trim = 70 48 70 22 , clip]{I20_22}\hspace{-1.5mm}
\includegraphics[width=0.053\linewidth, trim = 70 48 70 22 , clip]{I20_33}\hspace{-1.5mm}
\includegraphics[width=0.053\linewidth, trim = 70 48 70 22 , clip]{I20_49}}
\vspace{-1.5mm}
\caption{\small{Face clustering: given face images of multiple subjects (top), the goal is to find images that belong to the same subject (bottom).}}
\label{fig:example-Faces}
\vspace{-1mm}
\end{figure*}

\IEEEPARstart{H}{igh-dimensional} data are ubiquitous in many areas of machine learning, signal and image processing, computer vision, pattern recognition, bioinformatics, etc. For instance, images consist of billions of pixels, videos can have millions of frames, text and web documents are associated with hundreds of thousands of features, etc. The high-dimensionality of the data not only increases the computational time and memory requirements of algorithms, but also adversely affects their performance due to the noise effect and insufficient number of samples with respect to the ambient space dimension, commonly referred to as the ``curse of dimensionality'' \cite{Bellman57}. 
%
%
However, high-dimensional data often lie in low-dimensional structures instead of being uniformly distributed across the ambient space. 
Recovering low-dimensional structures in the data helps to not only reduce the computational cost and memory requirements of algorithms, but also reduce the effect of high-dimensional noise in the data and improve the performance of inference, learning, and recognition tasks.


In fact, in many problems, data in a class or category can be well represented by a low-dimensional subspace of the high-dimensional ambient space. For example, feature trajectories of a rigidly moving object in a video \cite{Tomasi:IJCV92}, face images of a subject under varying illumination \cite{Basri:PAMI03}, and multiple instances of a hand-written digit with different rotations, translations, and thicknesses \cite{Hastie:StatSci98} lie in a low-dimensional subspace of the ambient space. As a result, the collection of data from multiple classes or categories lie in a union of low-dimensional subspaces. 
Subspace clustering (see \cite{Vidal:SPM11-SC} and references therein) refers to the problem of separating data according to their underlying subspaces and finds numerous applications in image processing (\eg, image representation and compression \cite{Hong:TIP06}) and computer vision (\eg, image segmentation \cite{Yang:CVIU08}, motion segmentation \cite{Costeira:IJCV98,Kanatani:ICCV01}, and temporal video segmentation \cite{Vidal:PAMI05}), as illustrated in Figures \ref{fig:example-MS} and \ref{fig:example-Faces}. Since data in a subspace are often distributed arbitrarily and not around a centroid, standard clustering methods \cite{Duda:04} that take advantage of the spatial proximity of the data in each cluster are not in general applicable to subspace clustering. Therefore, there is a need for having clustering algorithms that take into account the multi-subspace structure of the data.

\subsection{Prior Work on Subspace Clustering}
Existing algorithms can be divided into four main categories:  iterative, algebraic, statistical, and spectral clustering-based methods. 

\myparagraph{Iterative methods} Iterative approaches, such as K-subspaces \cite{Tseng:JOTA00, Ho:CVPR03} and median K-flats \cite{Zhang:WSM09} alternate between assigning points to subspaces and fitting a subspace to each cluster. The main drawbacks of such approaches are that they generally require to know the number and dimensions of the subspaces, and that they are sensitive to initialization. 

\myparagraph{Algebraic approaches}
Factorization-based algebraic approaches such as \cite{Costeira:IJCV98, Kanatani:ICCV01, Gear:IJCV98} find an initial segmentation by thresholding the entries of a similarity matrix built from the factorization of the data matrix. These methods are provably correct when the subspaces are independent, but fail when this assumption is violated. In addition, they are sensitive to noise and outliers in the data.
Algebraic-geometric approaches such as Generalized Principal Component Analysis (GPCA) \cite{Vidal:PAMI05,Ma:SIAM08}, fit the data with a polynomial whose gradient at a point gives the normal vector to the subspace containing that point. 
While GPCA can deal with subspaces of different dimensions, 
it is sensitive to noise and outliers, and its complexity increases exponentially in terms of the number and dimensions of subspaces. 

\myparagraph{Statistical methods} Iterative statistical approaches, such as Mixtures of Probabilistic PCA (MPPCA) \cite{Tipping-mixtures:99}, Multi-Stage Learning (MSL) \cite{Kanatani:SMVP04}, or \cite{Gruber-Weiss:CVPR04}, assume that the distribution of the data inside each subspace is Gaussian and alternate between data clustering and subspace estimation by applying Expectation Maximization (EM). 
The main drawbacks of these methods are that they generally need to know the number and dimensions of the subspaces, and that they are sensitive to initialization. 
Robust statistical approaches, such as Random Sample Consensus (RANSAC) \cite{RANSAC}, fit a subspace of dimension $d$ to randomly chosen subsets of $d$ points until the number of inliers is large enough. The inliers are then removed, and the process is repeated to find a second subspace, and so on. RANSAC can deal with noise and outliers, and does not need to know the number of subspaces. However, the dimensions of the subspaces must be known and equal. In addition, the complexity of the algorithm increases exponentially in the dimension of the subspaces. 
%
%
Information-theoretic statistical approaches, such as Agglomerative Lossy Compression (ALC) \cite{Rao:PAMI10}, look for the segmentation of the data that minimizes the coding length needed to fit the points with a mixture of degenerate Gaussians up to a given distortion. As this minimization problem is NP-hard, a suboptimal solution is found by first assuming that each point forms its own group, and then iteratively merging pairs of groups to reduce the coding length. ALC can handle noise and outliers in the data. While, in principle, it does not need to know the number and dimensions of the subspaces, the number of subspaces found by the algorithms 
is dependent on the choice of a distortion parameter. In addition, there is no theoretical proof for the optimality of the agglomerative algorithm.

\myparagraph{Spectral clustering-based methods} Local spectral clustering-based approaches such as Local Subspace Affinity (LSA) \cite{Yan:ECCV06}, Locally Linear Manifold Clustering (LLMC) \cite{Goh:CVPR07}, Spectral Local Best-fit Flats (SLBF) \cite{Zhang:CVPR10}, and \cite{Zelnik-Manor:CVPR03} use local information around each point to build a similarity between pairs of points. The segmentation of the data is then obtained by applying spectral clustering \cite{Ng:NIPS01, vonLuxburg:StatComp2007} to the similarity matrix. These methods have difficulties in dealing with points near the intersection of two subspaces, because the neighborhood of a point can contain points from different subspaces. In addition, they are sensitive to the right choice of the neighborhood size to compute the local information at each point. 

Global spectral clustering-based approaches try to resolve these issues by building better similarities between data points using global information. Spectral Curvature Clustering (SCC) \cite{Chen:IJCV09} uses multi-way similarities that capture the curvature of a collection of points within an affine subspace. SCC can deal with noisy data but requires to know the number and dimensions of subspaces and assumes that subspaces have the same dimension. In addition, the complexity of building the multi-way similarity grows exponentially with the dimensions of the subspaces, hence, in practice, a sampling strategy is employed to reduce the computational cost. Using advances in sparse \cite{Donoho:CPAM06, Candes-Tao:TIT05, Tibshirani:RSS96} and low-rank  \cite{Candes:ACM10, Candes-Recht:FCM08, Recht:SIAM10} recovery algorithms, Sparse Subspace Clustering (SSC) \cite{Elhamifar:CVPR09, Elhamifar:ICASSP10, Candes:ASTAT12}, Low-Rank Recovery (LRR) \cite{Liu:ICML10, Liu:ICCV11, Liu:PAMI12}, and Low-Rank Subspace Clustering (LRSC) \cite{Favaro:CVPR11} algorithms pose the clustering problem as one of finding a sparse or low-rank representation of the data in the dictionary of the data itself. The solution of the corresponding global optimization algorithm is then used to build a similarity graph from which the segmentation of the data is obtained. The advantages of these methods with respect to most state-of-the-art algorithms are that they can handle noise and outliers in data, and that they do not need to know the dimensions and, in principle, the number of subspaces a priori.

\subsection{Paper Contributions}
In this paper, we propose and study an algorithm based on sparse representation techniques, called Sparse Subspace Clustering (SSC), to cluster a collection of data points lying in a union of low-dimensional subspaces. The underlying idea behind the algorithm is what we call the \emph{self-expressiveness} property of the data, which states that each data point in a union of subspaces can be efficiently represented as a linear or affine combination of other points. Such a representation is not unique in general because there are infinitely many ways in which a data point can be expressed as a combination of other points. The key observation is that a \emph{sparse representation} of a data point 
ideally corresponds to a combination of a few points from its own subspace. This motivates solving a global sparse optimization program whose solution is used in a spectral clustering framework to infer the clustering of data. As a result, we can overcome the problems of local spectral clustering-based algorithms, such as choosing the right neighborhood size and dealing with points near the intersection of subspaces, since, for a given data point, the sparse optimization program automatically picks a few other points that are not necessarily close to it but belong to the same subspace.

Since solving the sparse optimization program is in general NP-hard, we consider its $\ell_1$ relaxation. We show that, under mild conditions on the arrangement of subspaces and data distribution, the proposed $\ell_1$-minimization program recovers the desired solution, guaranteeing the success of the algorithm. Our theoretical analysis extends the sparse representation theory to the multi-subspace setting where the number of points in a subspace is arbitrary, possibly much larger than its dimension. Unlike block-sparse recovery problems \cite{Elhamifar:CVPR11, Elhamifar:TSP12, Parvaresh:STSP08, Stojnic:TSP09, Eldar:TIT09, Eldar:TSP10} where the bases for the subspaces are known and given, we do not have the bases for subspaces nor do we know which data points belong to which subspace, making our case more challenging. We only have the sparsifying dictionary for the union of subspaces given by the matrix of data points.

The proposed $\ell_1$-minimization program can be solved efficiently using convex programming tools 
\cite{BoydVandenberghe04,Boyd:STSP07,Boyd:FTML10} and does not require initialization. Our algorithm can directly deal with noise, sparse outlying entries, and missing entries in the data as well as the more general class of affine subspaces by incorporating the data corruption or subspace model into the sparse optimization program. Finally, through experimental results, we show that our algorithm outperforms state-of-the-art subspace clustering methods on the two real-world problems of motion segmentation (Fig. \ref{fig:example-MS}) and face clustering (Fig. \ref{fig:example-Faces}).

\myparagraph{Paper Organization} In Section \ref{sec:ssc}, we motivate and introduce the SSC algorithm for clustering data points in a union of linear subspaces. In Section \ref{sec:practical}, we generalize the algorithm to deal with noise, sparse outlying entries, and missing entries in the data as well as the more general class of affine subspaces. In Section \ref{sec:theory}, we investigate theoretical conditions under which the $\ell_1$-minimization program recovers the desired sparse representations of data points. In Section \ref{sec:connectivity}, we discuss the connectivity of the similarity graph and propose a regularization term to increase the connectivity of points in each subspace. In Section \ref{sec:syntheticexperiments}, we verify our theoretical analysis through experiments on synthetic data. In Section \ref{sec:realexperiments}, we compare the performance of SSC with the state of the art on the two real-world problems of motion segmentation and face clustering. Finally, Section \ref{sec:conclusions} concludes the paper.

\begin{figure*}[t!]
\centering
\includegraphics[width=0.22\linewidth, trim = 2 -4 0 2 , clip]{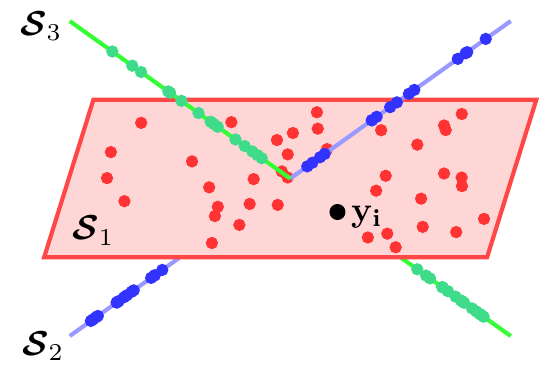}\ \hspace{4mm}
\includegraphics[width=0.242\linewidth, trim = 16 14 14 4 , clip]{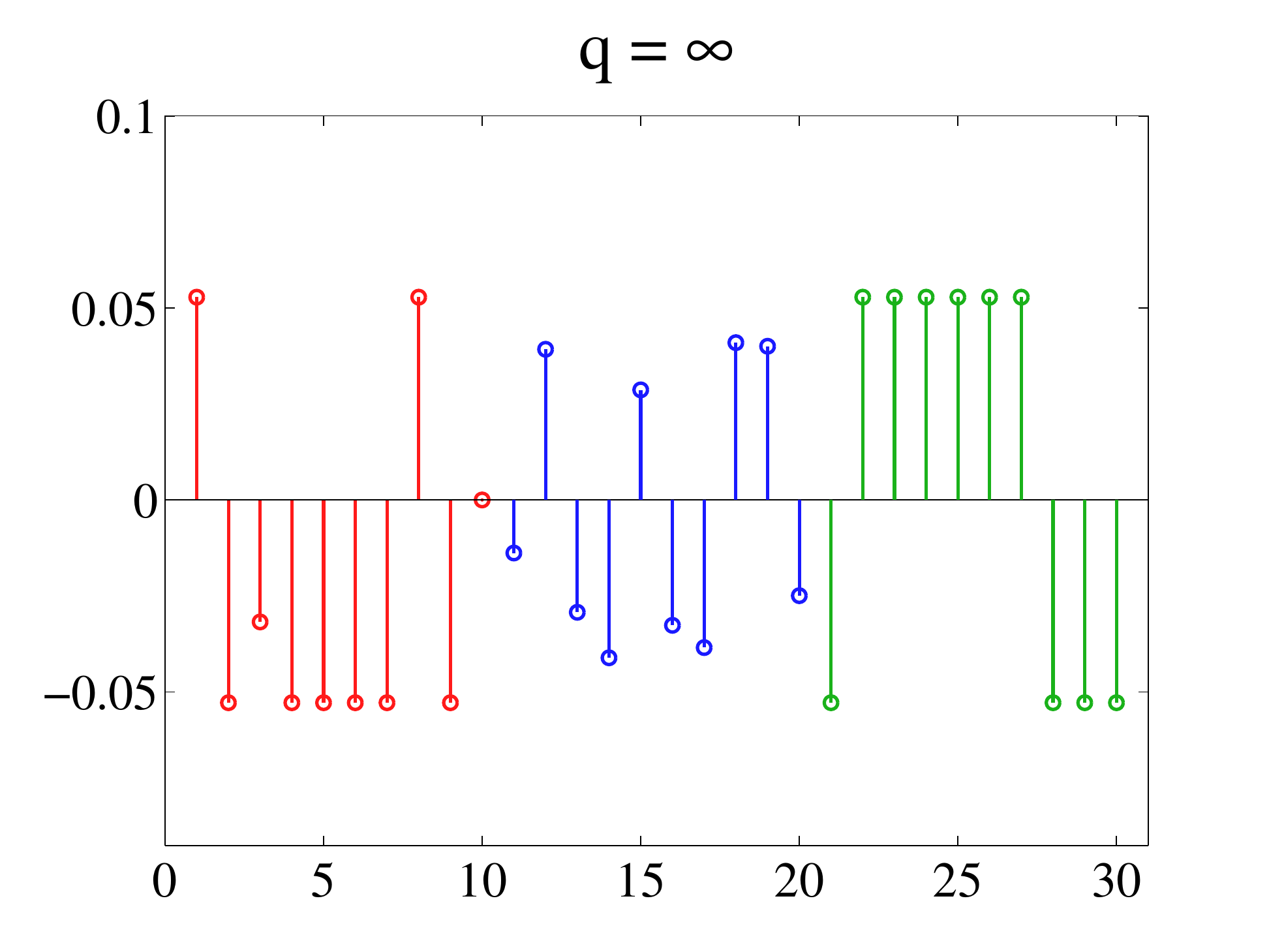}\
\includegraphics[width=0.24\linewidth, trim = 16 14 14 4 , clip]{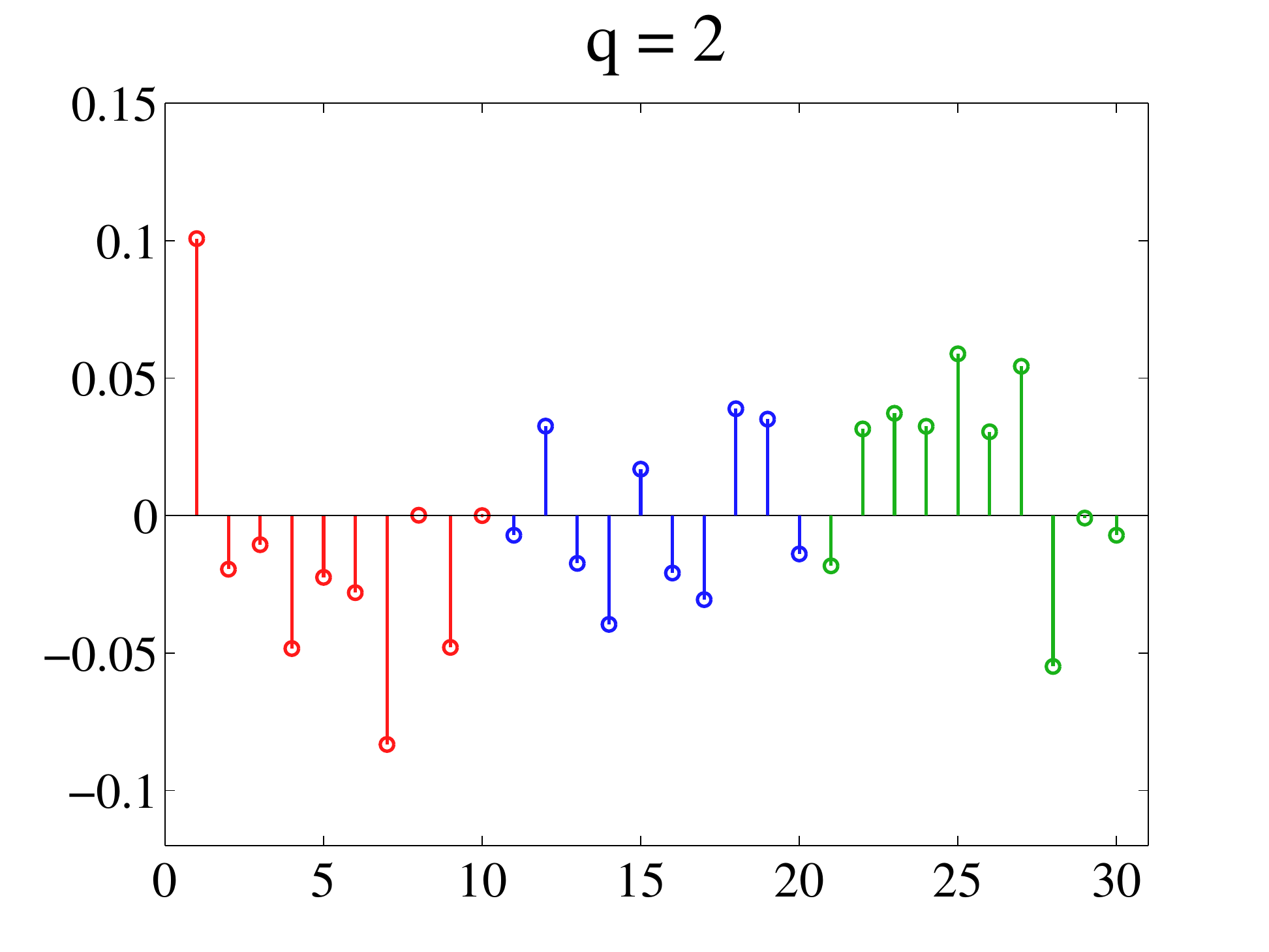}\
\includegraphics[width=0.24\linewidth, trim = 16 14 14 4 , clip]{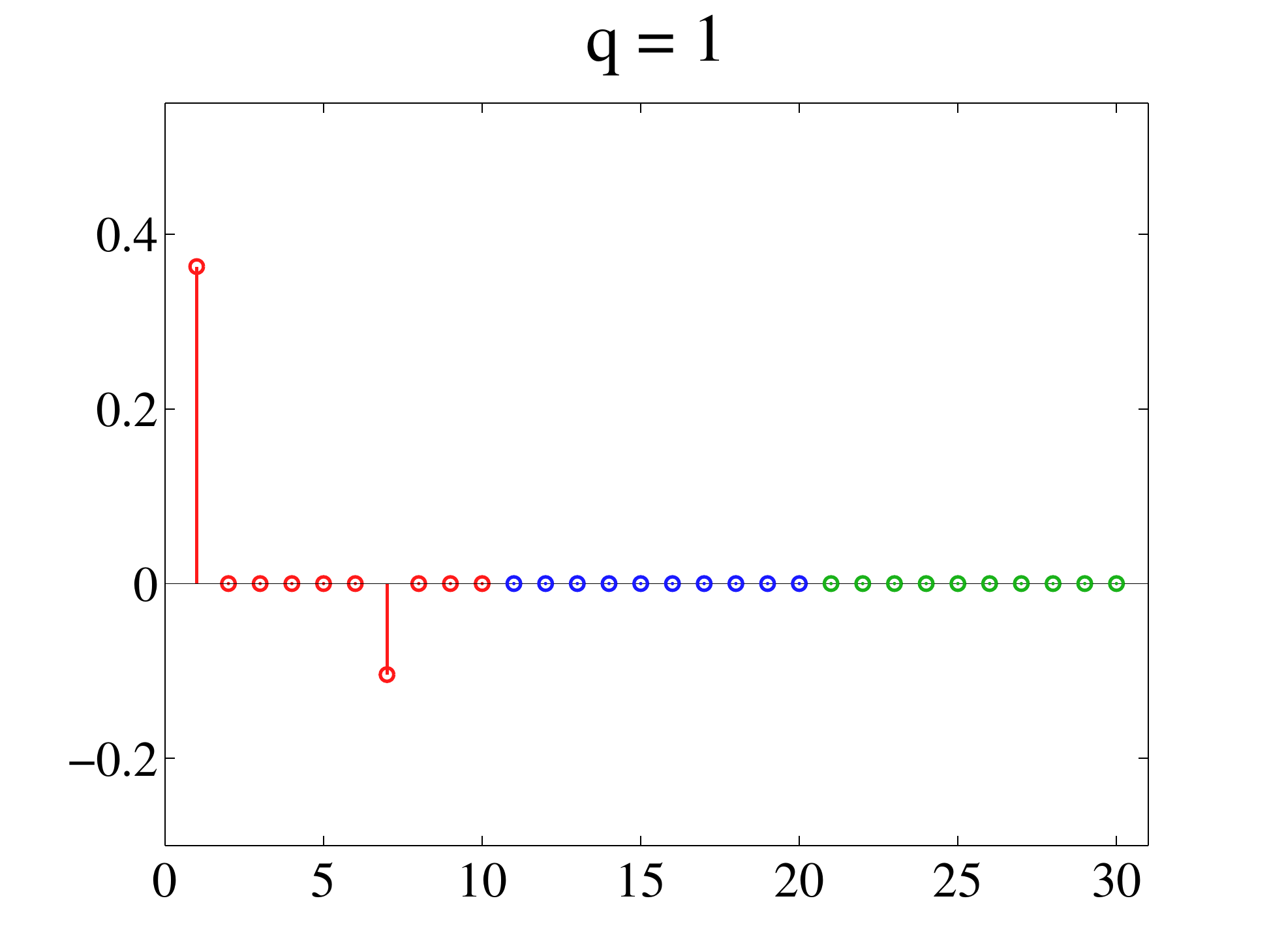}
\vspace{-1mm}
\caption{\small{Three subspaces in $\Re^3$ with $10$ data points in each subspace, ordered such that the fist and the last $10$ points belong to $\S_1$ and $\S_3$, respectively. The solution of the $\ell_q$-minimization program in \eqref{eq:Lq} for $\y_i$ lying in $\S_1$ for $q=1, 2, \infty$ is shown. Note that as the value of $q$ decreases, the sparsity of the solution increases. For $q=1$, the solution corresponds to choosing two other points lying in $\S_1$.}}

\label{fig:L1idea}
\end{figure*}
%

\section{Sparse Subspace Clustering}
\label{sec:ssc}
In this section, we introduce the sparse subspace clustering (SSC) algorithm for clustering a collection of multi-subspace data using sparse representation techniques. We motivate and formulate the algorithm for data points that perfectly lie in a union of linear subspaces. In the next section, we will generalize the algorithm to deal with data nuisances such as noise, sparse outlying entries, and missing entries as well as the more general class of affine subspaces. 


Let $\{ \S_{\ell}\}_{\ell=1}^{n}$ be an arrangement of $n$ linear subspaces of $\Re^{D} $ of dimensions $\{d_{\ell}\}_{\ell=1}^n$. Consider a given collection of $N$ noise-free data points $\{ \y_i \}_{i=1}^{N}$ that lie in the union of the $n$ subspaces. 
Denote the matrix containing all the data points as
\begin{equation}
\Y \triangleq \begin{bmatrix} \y_1 \!& \ldots \!& \y_N \end{bmatrix} = \begin{bmatrix} \Y_1 \!& \ldots \!& \Y_n \end{bmatrix} \boldsymbol{\Gamma}, 
\end{equation}
where $\Y_{\ell} \in \Re^{D \times N_{\ell}}$ is a rank-$d_\ell$ matrix of the $N_{\ell} > d_\ell$ points that lie in $\S_{\ell}$ and $\boldsymbol{\Gamma} \in \Re^{N \times N}$ is an unknown permutation matrix. 
We assume that we do not know a priori the bases of the subspaces nor do we know which data points belong to which subspace. The \emph{subspace clustering} problem refers to the problem of finding the number of subspaces, their dimensions, a basis for each subspace, and the segmentation of the data from $\Y$. 

To address the subspace clustering problem, we propose an algorithm that consists of two steps. In the first step, for each data point, we find a few other points that belong to the same subspace. To do so, we propose a global sparse optimization program whose solution encodes information about the memberships of data points to the underlying subspace of each point. In the second step, we use these information in a spectral clustering framework to infer the clustering of the data.

\subsection{Sparse Optimization Program}
Our proposed algorithm takes advantage of what we refer to as the \emph{self-expressiveness property} of the data, \ie, 
\smallskip
\begin{quote}
each data point in a union of subspaces can be efficiently reconstructed by a combination of other points in the dataset.
\end{quote}
\smallskip
More precisely, each data point for data point $\y_i \in \cup_{\ell=1}^n \S_{\ell}$ can be written as
%
\begin{equation}
\label{eq:selfexpression}
\y_i = \Y \c_i, \quad c_{ii} = 0,
\end{equation}
where $\c_i \triangleq \begin{bmatrix} c_{i1} & c_{i2} & \ldots & c_{iN} \end{bmatrix}^{\top}$ and the constraint $c_{ii} = 0$ eliminates the trivial solution of writing a point as a linear combination of itself. In other words, the matrix of data points $\Y$ is a self-expressive dictionary in which each point can be written as a linear combination of other points. 
However, the representation of $\y_i$ in the dictionary $\Y$ is \emph{not unique} in general. 
This comes from the fact that the number of data points in a subspace is often greater than its dimension, \ie, $N_{\ell} > d_{\ell}$. As a result, each $\Y_{\ell}$, and consequently $\Y$, has a non-trivial nullspace giving rise to infinitely many representations of each data point.

The key observation in our proposed algorithm is that among all solutions of \eqref{eq:selfexpression},
\smallskip
\begin{quote}
there exists a sparse solution, $\c_i$, whose nonzero entries correspond to data points from the same subspace as $\y_i$. We refer to such a solution as a subspace-sparse representation.
\end{quote}
\smallskip
More specifically, a data point $\y_i$ that lies in the $d_{\ell}$-dimensional subspace $\S_{\ell}$ can be written as a linear combination of $d_{\ell}$ other points in general directions from $\S_{\ell}$. 
As a result, ideally, a sparse representation of a data point finds points from the same subspace where the number of the nonzero elements corresponds to the dimension of the underlying subspace.

For a system of equations such as \eqref{eq:selfexpression} with infinitely many solutions, one can restrict the set of solutions 
by minimizing an objective function such as the $\ell_q$-norm of the solution\footnote{The $\ell_q$-norm of $\c_i \in \Re^N$ is defined as $\| \c_i \|_q \triangleq (\sum_{j=1}^{N}{|c_{ij}|^q})^{\frac{1}{q}}$.} as
\begin{equation}
\label{eq:Lq}
\min \| \c_i \|_q \quad \operatorname{s.t.} \quad \y_i = \Y \c_i, \; c_{ii} = 0.
\end{equation}
Different choices of $q$ have different effects in the obtained solution. 
Typically, by decreasing the value of $q$ from infinity toward zero, the sparsity of the solution increases, as shown in Figure \ref{fig:L1idea}. The extreme case of $q=0$ corresponds to the general NP-hard problem \cite{Amaldi:TCS98} of finding the sparsest representation of the given point, as the $\ell_0$-norm counts the number of nonzero elements of the solution. 
Since we are interested in efficiently finding a non-trivial sparse representation of $\y_i$ in the dictionary $\Y$, we consider minimizing the tightest convex relaxation of the $\ell_0$-norm, \ie,
\begin{equation}
\label{eq:L1vec}
\min \| \c_i \|_1 \quad \operatorname{s.t.} \quad \y_i = \Y \c_i, \; c_{ii} = 0,
\end{equation}
which can be solved efficiently using convex programming tools 
\cite{BoydVandenberghe04, Boyd:STSP07, Boyd:FTML10} and is known to prefer sparse solutions \cite{Donoho:CPAM06, Candes-Tao:TIT05, Tibshirani:RSS96}. 

We can also rewrite the sparse optimization program \eqref{eq:L1vec} for all data points $i=1,\dots,N$ in matrix form as
\begin{equation}
\label{eq:L1mat}
\min \| \C \|_1 \quad \operatorname{s.t.} \quad \Y = \Y \C,~~\operatorname{diag}(\C) = \0,
\end{equation}
where $\C \triangleq \begin{bmatrix} \c_1 \!& \c_2 \!& \ldots \!& \c_N \end{bmatrix} \in \Re^{N \times N}$ is the matrix whose $i$-th column corresponds to the sparse representation of $\y_i$, $\c_i$, and $\operatorname{diag}(\C) \in \Re^{N}$ is the vector of the diagonal elements of $\C$. 

Ideally, the solution of \eqref{eq:L1mat} corresponds to subspace-sparse representations of the data points, which we use next to infer the clustering of the data. In Section \ref{sec:theory}, we study conditions under which the convex optimization program in \eqref{eq:L1mat} is guaranteed to recover a subspace-sparse representation of each data point.

\subsection{Clustering using Sparse Coefficients}
After solving the proposed optimization program in \eqref{eq:L1mat}, we obtain a sparse representation for each data point whose nonzero elements ideally correspond to points from the same subspace. The next step of the algorithm is to infer the segmentation of the data into different subspaces using the sparse coefficients.

To address this problem, 
we build a weighted graph $\G=(\V,\edge,\W)$, where $\V$ denotes the set of $N$ nodes of the graph corresponding to $N$ data points and $\edge \subseteq \V \times \V$ denotes the set of edges between nodes. $\W \in \Re^{N \times N}$ is a symmetric non-negative similarity matrix representing the 
weights of the edges, \ie, node $i$ is connected to node $j$ by an edge whose weight is equal to $w_{ij}$. An ideal similarity matrix $\W$, hence an ideal similarity graph $\G$, is one in which nodes that correspond to points from the same subspace are connected to each other and there are no edges between nodes that correspond to points in different subspaces. 

Note that the sparse optimization program ideally recovers to a subspace-sparse representation of each point, \ie, a representation whose nonzero elements correspond to points from the same subspace of the given data point. This provides an immediate choice of the similarity matrix as $\W = |\C| + |\C|^{\top}$. In other words, each node $i$ connects itself to a node $j$ by an edge whose weight is equal to $|c_{ij}| + |c_{ji}|$. The reason for the symmetrization is that, in general, a data point $\y_i \in \S_{\ell}$ can write itself as a linear combination of some points including $\y_j \in \S_{\ell}$. 
However, $\y_j$ may not necessarily choose $\y_i$ in its sparse representation. By this particular choice of the weight, we make sure that nodes $i$ and $j$ get connected to each other if either $\y_i$ or $\y_j$ is in the sparse representation of the other.\footnote{To obtain a symmetric similarity matrix, one can directly impose the constraint of $\C = \C^\top$ in the optimization program. However, this results in increasing the complexity of the optimization program and, in practice, does not perform better than the post-symmetrization of $\C$, as described above. See also \cite{Zass:NIPS06} for other processing approaches of the similarity matrix.}

The similarity graph built this way has ideally $n$ connected components corresponding to the $n$ subspaces, \ie, 
\begin{equation}
\W \!=\! \begin{bmatrix} \W_1 \!\!\!& \cdots \!\!\!& \0 \\ \vdots \!\!\!& \ddots \!\!\!& \vdots \\ \0 \!\!\!& \cdots \!\!\!& \W_n  \end{bmatrix} \!\boldsymbol{\Gamma},
\end{equation}
where $\W_{\ell}$ is the similarity matrix of data points in $\S_{\ell}$. 
%
Clustering of data into subspaces follows then by applying spectral clustering \cite{Ng:NIPS01} to the graph $\G$. More specifically, we obtain the clustering of data by applying the Kmeans algorithm \cite{Duda:04} to the normalized rows of a matrix whose columns are the $n$ bottom eigenvectors of the symmetric normalized Laplacian matrix of the graph. 

\smallskip
\begin{remark}
\label{rem:sim-norm}
An optional step prior to building the similarity graph is to normalize the sparse coefficients as $\c_i \leftarrow \c_i / \| \c_i \|_{\infty}$. This helps to better deal with different norms of data points. More specifically, if a data point with a large Euclidean norm selects a few points with small Euclidean norms, then the values of the nonzero coefficients will generally be large. On the other hand, if a data point with a small Euclidean norm selects a few points with large Euclidean norms, then the values of the nonzero coefficients will generally be small. Since spectral clustering puts more emphasis on keeping the stronger connections in the graph, by the normalization step we make sure that the largest edge weights for all the nodes are of the same scale. 
\end{remark}
\smallskip

Algorithm \ref{alg:SSC-linear} summarizes the SSC algorithm. Note that an advantage of spectral clustering, which will be shown in the experimental results, is that it provides robustness with respect to a few errors in the sparse representations of the data points. In other words, as long as edges between points in different subspaces are weak, spectral clustering 
can find the correct segmentation. 
\vspace{1mm}
\begin{remark}
In principle, SSC does not need to know the number of subspaces. More specifically, under the conditions of the theoretical results in Section \ref{sec:theory}, in the similarity graph there will be no connections between points in different subspaces. Thus, one can determine the number of subspaces by finding the number of graph components, which can be obtained by analyzing the eigenspectrum of the Laplacian matrix of $\G$ \cite{vonLuxburg:StatComp2007}. However, when there are connections between points in different subspaces, other model selection techniques should be employed \cite{Brox:ECCV10}.
\end{remark}
\begin{algorithm}[t]
\caption{\bf : Sparse Subspace Clustering (SSC)}
\textbf{Input:} A set of points $\{\y_i\}_{i=1}^{N}$ lying in a union of $n$ linear subspaces $\{ \S_i \}_{i=1}^{n}$.
\begin{algorithmic}[1]
\State Solve the sparse optimization program \eqref{eq:L1mat} in the case of uncorrupted data or \eqref{eq:L1noiseoutlier1} in the case of corrupted data.
\State Normalize the columns of $\C$ as $\c_i \leftarrow \frac{\c_i}{\| \c_i \|_{\infty}}$. 
\State Form a similarity graph with $N$ nodes representing the data points. Set the weights on the edges between the nodes by $\W = | \C | + | \C |^{\top}$.%
%
%
\vspace{3pt}
\State Apply spectral clustering \cite{Ng:NIPS01} to the similarity graph. 

\end{algorithmic}
\textbf{Output:} Segmentation of the data: $\Y_1, \Y_2, \ldots, \Y_n$.
\label{alg:SSC-linear}
\end{algorithm}
\section{Practical Extensions}
\label{sec:practical}
In real-world problems, data are often corrupted by noise and sparse outlying entries due to measurement/process noise and ad-hoc data collection techniques. In such cases, the data do not lie perfectly in a union of subspaces. For instance, in the motion segmentation problem, because of the malfunctioning of the tracker, feature trajectories can be corrupted by noise or can have entries with large errors \cite{Rao:PAMI10}. Similarly, in clustering of human faces, images can be corrupted by errors due to specularities, cast shadows, and occlusions \cite{Wright:PAMI09}. On the other hand, data points may have missing entries, \eg, when the tracker loses track of some feature points in a video due to occlusions \cite{Vidal:CVPR04-multiaffine}. Finally, data may lie in a union of affine subspaces, a more general model which includes linear subspaces as a particular case. 

In this section, we generalize the SSC algorithm for clustering data lying perfectly in a union of linear subspaces, to deal with the aforementioned challenges. Unlike state-of-the-art methods, which require to run a separate algorithm first to correct the errors in the data \cite{Rao:PAMI10, Vidal:CVPR04-multiaffine}, we deal with these problems in a unified framework by incorporating a model for the corruption into the sparse optimization program. Thus, the sparse coefficients again encode information about memberships of data to subspaces, which are used in a spectral clustering framework, as before.

\subsection{Noise and Sparse Outlying Entries}
In this section, we consider clustering of data points that are contaminated with sparse outlying entries and noise. Let 
%
%
\begin{equation}
\label{eq:corrupting}
\y_i = \y_i^0 + \e_i^0 + \z_i^0
\end{equation}
be the $i$-th data point that is obtained by corrupting an error-free point $\y_i^0$, which perfectly lies in a subspace, with a vector of sparse outlying entries $\e_i^0 \in \Re^D$ that has only a few large nonzero elements, \ie, $\| \e_i^0 \|_0 \leq k$ for some integer $k$, and with a noise $\z_i^0 \in \Re^D$ whose norm is bounded as $\| \z_i^0 \|_2 \leq \zeta$ for some $\zeta > 0$. Since error-free data points perfectly lie in a union of subspaces, using the self-expressiveness property, we can reconstruct $\y_i^0 \in \S_{\ell}$ in terms of other error-free points as
\begin{equation}
\label{eq:noisefreeL1}
\y_i^0 = \sum_{j \neq i}{c_{ij} \y_j^0}.
\end{equation}
Note that the above equation has a sparse solution since $\y_i^0$ can be expressed as a linear combination of at most $d_{\ell}$ other points from $\S_{\ell}$. Rewriting $\y_i^0$ using \eqref{eq:corrupting} in terms of the corrupted point $\y_i$, the sparse outlying entries vector $\e_i^0$, and the noise vector $\z_i^0$ and substituting it into \eqref{eq:noisefreeL1}, we obtain 
%
%
%
%
\begin{equation}
\label{eq:noisyL1}
\y_i = \sum_{j \neq i}{c_{ij} \y_j} + \e_i + \z_i,
\end{equation}
where the vectors $\e_i \in \Re^{D}$ and $\z_i \in \Re^{D}$ are defined as
\begin{eqnarray}
\label{eq:newoutlier}
\e_i &\triangleq  \e_i^0 - \sum_{j \neq i}{c_{ij} \e_j^0},\\
\label{eq:newnoise}
\z_i &\triangleq \z_i^0 - \sum_{j \neq i}{c_{ij} \z_j^0}.
\end{eqnarray}
Since \eqref{eq:noisefreeL1} has a sparse solution $\c_i$, $\e_i$ and $\z_i$ also correspond to vectors of sparse outlying entries and noise, respectively. More precisely, when a few $c_{ij}$ are nonzero, $\e_i$ is a vector of sparse outlying entries since it is a linear combination of a few vectors of outlying entries in \eqref{eq:newoutlier}. Similarly, when a few $c_{ij}$ are nonzero and do not have significantly large magnitudes\footnote{One can show that, under broad conditions, sum of $|c_{ij}|$ is bounded above by the square root of the dimension of the underlying subspace of $\y_i$. Theoretical guarantees of the proposed optimization program in the case of corrupted data is the subject of the current research.}, $\z_i$ is a vector of noise since it is linear combination of a few noise vectors in~\eqref{eq:newnoise}. 


Collecting $\e_i$ and $\z_i$ as columns of the matrices $\E$ and $\Z$, respectively, we can rewrite \eqref{eq:noisyL1} in matrix form as
\begin{equation}
\label{eq:noisyL1Mat}
\Y = \Y \C + \E + \Z, \quad \diag(\C) = \0. 
\end{equation}
Our objective is then to find a solution $(\C,\E,\Z)$ for \eqref{eq:noisyL1Mat}, where $\C$ corresponds to a sparse coefficient matrix, $\E$ corresponds to a matrix of sparse outlying entries, and $\Z$ is a noise matrix. To do so, we propose to solve the following optimization program
\begin{eqnarray}
\label{eq:L1noiseoutlier1}
\begin{split}
&\min \;\; \| \C \|_1 + \lambda_{e} \| \E \|_1 + \frac{\lambda_z}{2} \| \Z \|_F^2  \\ &\st \;\;\; \Y = \Y \C + \E + \Z, \;\; \diag(\C) = \0,
\end{split}
\end{eqnarray}
where the $\ell_1$-norm promotes sparsity of the columns of $\C$ and $\E$, while the Frobenius norm promotes having small entries in the columns of $\Z$. The two parameters $\lambda_e > 0$ and $\lambda_z > 0$ balance the three terms in the objective function. Note that the optimization program in \eqref{eq:L1noiseoutlier1} is convex with respect to the optimization variables $(\C,\E,\Z)$, hence, can be solved efficiently using convex programming tools. 

When data are corrupted only by noise, we can eliminate $\E$ from the optimization program in \eqref{eq:L1noiseoutlier1}. On the other hand, when the data are corrupted only by sparse outlying entries, we can eliminate $\Z$ in \eqref{eq:L1noiseoutlier1}. In practice, however, $\E$ can also deal with small errors due to noise. 
The following proposition suggests setting $\lambda_z = \alpha_z / \mu_z$ and $\lambda_e = \alpha_e / \mu_e$, where $\alpha_z , \alpha_e > 1$ and 
\begin{equation}
\label{eq:muz-mue}
\mu_z \triangleq \min_{i} \max_{j \neq i} | \y_i^{\top} \y_j |, \quad\; \mu_e \triangleq \min_{i} \max_{j \neq i} \| \y_j \|_1.
\end{equation}
The proofs of all theoretical results in the paper are provided in the supplementary material.
%
\vspace{1.5mm}
\begin{proposition}
\label{prop:lambdaE_setting}
Consider the optimization program \eqref{eq:L1noiseoutlier1}. Without the term $\Z$, if $\lambda_e \leq 1 / \mu_e$, then there exists at least one data point $\y_\ell$ for which in the optimal solution we have $(\c_\ell,\e_\ell) = (\0,\y_\ell)$. Also, without the term $\E$, if $\lambda_z \leq 1 / \mu_z$, then there exists at least one data point $\y_\ell$ for which $(\c_\ell,\z_\ell) = (\0,\y_\ell)$.
%
%
%
\end{proposition}
\vspace{1.5mm}
%


After solving the proposed optimization programs, we use $\C$ to build a similarity graph and infer the clustering of data using spectral clustering. Thus, by incorporating the corruption model of data into the sparse optimization program, we can deal with clustering of corrupted data, as before, without explicitly running a separate algorithm to correct the errors in the data \cite{Rao:PAMI10, Vidal:CVPR04-multiaffine}.
\subsection{Missing Entries}
We consider now the clustering of incomplete data, where some of the entries of a subset of the data points are missing. Note that when only a small fraction of the entries of each data point is missing, clustering of incomplete data can be cast as clustering of data with sparse outlying entries. More precisely, one can fill in the missing entries of each data point with random values, hence obtain data points with sparse outlying entries. Then clustering of the data follows by solving \eqref{eq:L1noiseoutlier1} and applying spectral clustering to the graph built using the obtained sparse coefficients. However, the drawback of this approach is that it disregards the fact that we know the locations of the missing entries in the data matrix.

It is possible, in some cases, to cast the clustering of data with missing entries as clustering of complete data. To see this, consider a collection of data points $\{ \y_i \}_{i=1}^{N}$ in $\Re^D$. Let $J_i \subset \{1,\ldots, D\}$ denote indices of the known entries of $\y_i$ and define $J \triangleq \bigcap_{i=1}^{N} J_i$. Thus, for every index in $J$, all data points have known entries. When the size of $J$, denoted by $|J|$, is not small relative to the ambient space dimension, $D$, we can project the data, hence, the original subspaces, into a subspace spanned by the columns of the identity matrix indexed by $J$ and apply the SSC algorithm to the obtained complete data. In other words, we can only keep the rows of $\Y$ indexed by $J$, obtain a new data matrix of complete data $\bar{\Y} \in \Re^{|J| \times N}$, and solve the sparse optimization program \eqref{eq:L1noiseoutlier1}.
We can then infer the clustering of the data by applying spectral clustering to the graph built using the sparse coefficient matrix. Note that the approach described above is based on the assumption that $J$ is nonempty. Addressing the problem of subspace clustering with missing entries when $J$ is empty or has a small size is the subject of the future research.

\subsection{Affine Subspaces}
In some real-world problems, the data lie in a union of affine rather than linear subspaces. For instance, the motion segmentation problem involves clustering of data that lie in a union of $3$-dimensional affine subspaces \cite{Tomasi:IJCV92,Vidal:CVPR04-multiaffine}. 
A naive way to deal with this case is to ignore the affine structure of the data and perform clustering as in the case of linear subspaces. This comes from the fact that a $d_{\ell}$-dimensional affine subspace $\S_{\ell}$ can be considered as a subset of a $(d_{\ell}+1)$-dimensional linear subspace that includes $\S_{\ell}$ and the origin. However, this has the drawback of possibly increasing the dimension of the intersection of two subspaces, which in some cases can result in indistinguishability of subspaces from each other. For example, two different lines $x = -1$ and $x = +1$ in the $x$-$y$ plane form the same $2$-dimensional linear subspace after including the origin, hence become indistinguishable.


To directly deal with affine subspaces, we use the fact that any data point $\y_i$ in an affine subspace $\S_\ell$ of dimension $d_\ell$ can be written as an affine combination of $d_\ell+1$ other points from $\S_\ell$. In other words, a sparse solution of
%
%
\begin{equation}
\y_i = \Y \c_i, \quad \1^{\top} \c_i = 1, \;\; c_{ii} = 0,
\end{equation}
corresponds to $d_{\ell}+1$ other points that belong to $\S_{\ell}$ containing $\y_i$. Thus, to cluster data points lying close to a union of affine subspaces, we propose to solve the sparse optimization program 
%
\begin{eqnarray}
\label{eq:L1affine}
\begin{split}
&\min \;\; \| \C \|_1 + \lambda_{e} \| \E \|_1 + \frac{\lambda_z}{2} \| \Z \|_F^2  \\ &\st \;\;\; \Y = \Y \C + \E + \Z,~~\1^{\top} \C = \1^{\top},~\diag(\C) = \0,
\end{split}
\end{eqnarray}
which, in comparison to \eqref{eq:L1noiseoutlier1} for the case of linear subspaces, includes additional linear equality constraints. Note that \eqref{eq:L1affine} can deal with linear subspaces as well since a linear subspace is also an affine subspace. 

%
\begin{figure}[t!]
\centering
\includegraphics[width=0.46\linewidth, trim = 0 3 0 0 , clip]{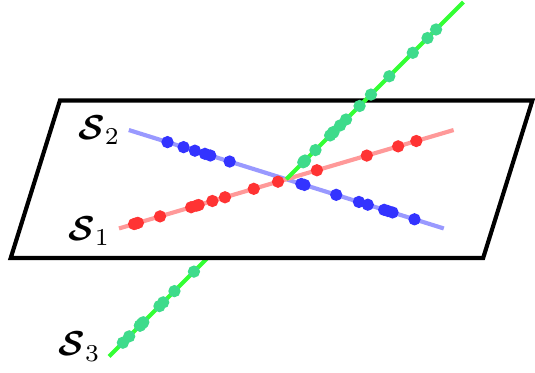}\ \hspace{1mm}
\includegraphics[width=0.46\linewidth, trim = 0 -13 0 0 , clip]{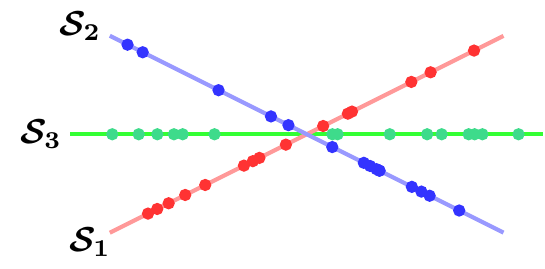}
\vspace{-1mm}
\caption{\small{Left: the three $1$-dimensional subspaces are independent as they span the $3$-dimensional space and the sum of their dimensions is also $3$. Right: the three $1$-dimensional are disjoint as any two subspaces intersect at the origin.}}
\label{fig:independent-disjoint}
\end{figure}
%

\section{Subspace-Sparse Recovery Theory}
\label{sec:theory}
The underlying assumption for the success of the SSC algorithm is that the proposed optimization program recovers a subspace-sparse representation of each data point, \ie, a representation whose nonzero elements correspond to the subspace of the given point. In this section, we investigate conditions under which, for data points that lie in a union of linear subspaces, the sparse optimization program in \eqref{eq:L1vec} recovers subspace-sparse representations of data points. We investigate recovery conditions for two classes of subspace arrangements: \emph{independent} and \emph{disjoint} {subspace models} \cite{Elhamifar:ICASSP10}.
\vspace{1.5mm}
\begin{definition}
\label{def:independent}
\emph{A collection of subspaces $\{ \S_i\}_{i=1}^{n}$ is said to be \emph{independent} if $\dim(\oplus_{i=1}^{n} {\S_i}) = \sum_{i=1}^{n} {\dim(\S_i)}$, where $\oplus$ denotes the direct sum operator.}
\end{definition}
\vspace{1.5mm}
As an example, the three $1$-dimensional subspaces shown in Figure \ref{fig:independent-disjoint} (left) are independent since they span a $3$-dimensional space and the sum of their dimensions is also $3$. On the other hand, the subspaces shown in Figure \ref{fig:independent-disjoint} (right) are not independent since they span a $2$-dimensional space while the sum of their dimensions is $3$.
\vspace{1.5mm}
\begin{definition}
\label{def:disjoint}
\emph{A collection of subspaces $\{ \S_i\}_{i=1}^{n}$ is said to be \emph{disjoint} if every pair of subspaces intersect only at the origin. In other words, for every pair of subspaces we have $\dim(\S_i \oplus \S_j) = \dim(\S_i) + \dim(\S_j)$.}
\end{definition}
\smallskip
As an example, both subspace arrangements shown in Figure \ref{fig:independent-disjoint} are disjoint since each pair of subspaces intersect at the origin. Note that, based on the above definitions, the notion of disjointness is weaker than independence as an independent subspace model is always disjoint while the converse is not necessarily true. An important notion that can be used to characterize two disjoint subspaces is the smallest principal angle, defined as follows.
\smallskip
\begin{definition}
\label{def:subspaceangle}
\emph{The \emph{smallest principal angle} between two subspaces $\S_i$ and $\S_j$, denoted by $\theta_{ij}$, is defined as
\begin{equation}
\label{eq:smallprincipalangle}
\cos(\theta_{ij}) \triangleq \max_{\v_i \in \S_i,\v_j \in \S_j} { \frac{\v_i^{\top} \v_j}{\| \v_i \|_2 \| \v_j \|_2} }.
\end{equation}
}
\end{definition}
\vspace{1.5mm}
Note that two disjoint subspaces intersect at the origin, hence their smallest principal angle is greater than zero and $\cos(\theta_{ij}) \in [0,1)$.
\subsection{Independent Subspace Model}
In this section, we consider data points that lie in a union of independent subspaces, which is the underlying model of many subspace clustering algorithms. We show that the $\ell_1$-minimization program in \eqref{eq:L1vec} and more generally the $\ell_q$-minimization in \eqref{eq:Lq} for $q < \infty$ always recover subspace-sparse representations of the data points. More specifically, we show the following result.
%
\vspace{1.5mm}
\begin{theorem}
\label{thm:independent}
\emph{Consider a collection of data points drawn from $n$ independent subspaces $\{\S_i\}_{i=1}^n$ of dimensions $\{ d_i \}_{i=1}^{n}$. Let $\Y_i$ denote $N_i$ data points in $\S_i$, where $\rank(\Y_i) = d_i$, and let $\Y_{-i}$ denote data points in all subspaces except $\S_i$. Then, for every $\S_i$ and every nonzero $\y$ in $\S_i$, the $\ell_q$-minimization program
\begin{equation}
\label{eq:Lq2}
\begin{bmatrix} \c^* \\ \c^*_{-} \end{bmatrix} = \argmin \left \| \begin{bmatrix} \c \\ \c_{-} \end{bmatrix} \right \|_q ~~ \operatorname{s.t.} ~~ \y = [ \Y_i ~~ \Y_{-i} ] \begin{bmatrix} \c \\ \c_{-} \end{bmatrix},
\end{equation}
for $q < \infty$, recovers a subspace-sparse representation, \ie, $\c^* \neq \0$ and $\c^*_{-} = \0$.
}
\end{theorem}
\vspace{1.5mm}
Note that the subspace-sparse recovery holds without any assumption on the distribution of the data points in each subspace, other than $\rank(\Y_i) = d_i$. This comes at the price of having a more restrictive model for the subspace arrangements. Next, we will show that for the more general class of disjoint subspaces, under appropriate conditions on the relative configuration of the subspaces as well as the distribution of the data in each subspace, the $\ell_1$-minimization in \eqref{eq:L1vec} recovers subspace-sparse representations of the data points. 
\subsection{Disjoint Subspace Model}
We consider now the more general class of disjoint subspaces and investigate conditions under which the optimization program in \eqref{eq:L1vec} recovers a subspace-sparse representation of each data point. To that end, we consider a vector $\x$ in the intersection of $\S_i$ with $\oplus_{j \neq i}{\S_j}$ and let the optimal solution of the $\ell_1$-minimization when we restrict the dictionary to data points from $\S_i$ be
\begin{equation}
\label{eq:L1samesubspace}
\a_i = \argmin \| \a \|_1 \quad \operatorname{s.t.} \quad \x = \Y_i \; \a .
\end{equation}
We also let the optimal solution of the $\ell_1$-minimization when we restrict the dictionary to points from all subspaces except $\S_i$ be
\begin{equation}
\label{eq:L1allsubspace}
\a_{-i} = \argmin \| \a \|_1 \quad \operatorname{s.t.} \quad \x = \Y_{-i} \; \a. \footnote{In fact, $\a_{i}$ and $\a_{-i}$ depend on $\x$, $\Y_i$, and $\Y_{-i}$. Since this dependence is clear from the context, we drop the arguments in $\a_i(\x,\Y_i)$ and $\a_{-i}(\x,\Y_{-i})$.}
\end{equation}
%
We show in the supplementary material that the SSC algorithm succeeds in recovering subspace-sparse representations of data points in each $\S_i$, if for every nonzero $\x$ in the intersection of $\S_i$ with $\oplus_{j \neq i}{\S_j}$, the $\ell_1$-norm of the solution of \eqref{eq:L1samesubspace} is strictly smaller than the $\ell_1$-norm of the solution of \eqref{eq:L1allsubspace}, \ie, 
\begin{equation}
\label{eq:suffcondgeneral}
\forall \, \x \in \S_i \cap (\oplus_{j \neq i}{\S_j}), \x \neq \0 \implies \| \a_i \|_1 < \| \a_{-i} \|_1.
\end{equation}
More precisely, we show the following result. 
\smallskip
\begin{theorem}
\label{thm:sufficient-condition}
\emph{Consider a collection of data points drawn from $n$ disjoint subspaces $\{ \S_i \}_{i=1}^n$ of dimensions $\{ d_i \}_{i=1}^{n}$. Let $\Y_i$ denote $N_i$ data points in $\S_i$, where $\rank(\Y_i) = d_i$, and let $\Y_{-i}$ denote data points in all subspaces except $\S_i$. The $\ell_1$-minimization 
\begin{equation}
\label{eq:L1}
\begin{bmatrix} \c^* \\ \c^*_{-} \end{bmatrix} = \argmin \left \| \begin{bmatrix} \c \\ \c_{-} \end{bmatrix} \right \|_1 ~~ \operatorname{s.t.} ~~ \y = [ \Y_i ~~ \Y_{-i} ] \begin{bmatrix} \c \\ \c_{-} \end{bmatrix},
\end{equation}
recovers a subspace-sparse representation of every nonzero $\y$ in $\S_i$, \ie, $\c^* \neq \0$ and $\c^*_{-} = \0$, if and only if \eqref{eq:suffcondgeneral} holds.
}
\end{theorem}
\smallskip

While the necessary and sufficient condition in \eqref{eq:suffcondgeneral} guarantees a successful subspace-sparse recovery via the $\ell_1$-minimization program, it does not explicitly show the relationship between the subspace arrangements and the data distribution for the success of the $\ell_1$-minimization program. To establish such a relationship, we show that $\| \a_i \|_1 \leq \beta_i$, where $\beta_i$ depends on the singular values of data points in $\S_i$, and $\beta_{-i} \leq \| \a_{-i} \|_1$, where $\beta_{-i}$ depends on the subspace angles between $\S_i$ and other subspaces. Then, the sufficient condition $\beta_i < \beta_{-i}$ establishes the relationship between the subspace angles and the data distribution under which the $\ell_1$-minimization is successful in subspace-sparse recovery, since it implies that 
\begin{equation}
\| \a_i \|_1 \leq \beta_i < \beta_{-i} \leq \| \a_{-i} \|_1,
\end{equation}
\ie, the condition of Theorem \ref{thm:sufficient-condition} holds. 

%
\begin{figure*}[t]
\centering
\includegraphics[width=0.22\linewidth, trim = 0 0 0 0 , clip]{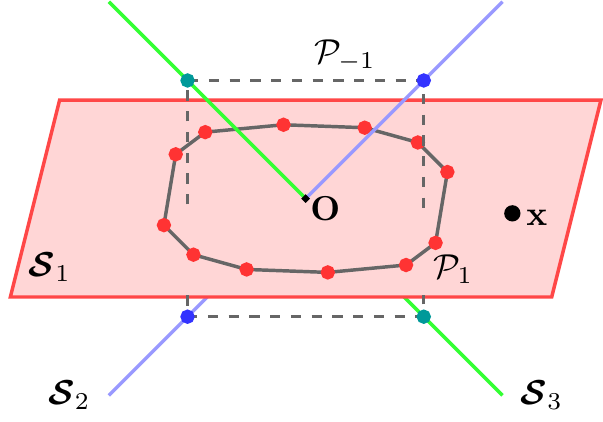}\ \hspace{12mm}
\includegraphics[width=0.22\linewidth, trim = 0 0 0 0 , clip]{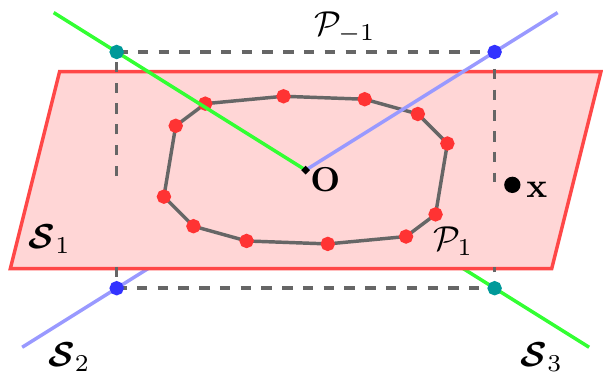}\ \hspace{12mm}
\includegraphics[width=0.22\linewidth, trim = 0 0 0 0 , clip]{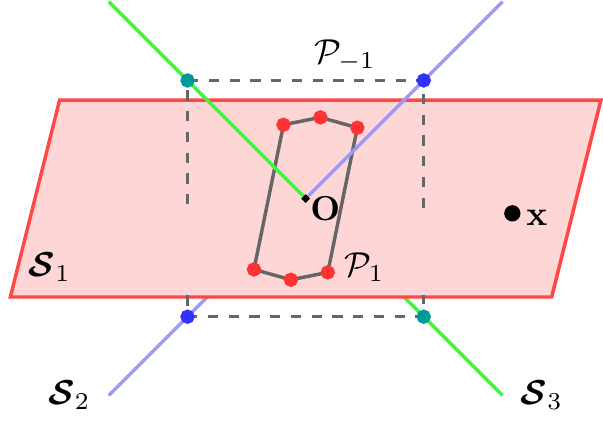}
\vspace{-2mm}
\caption{\small{Left: for any nonzero $\x$ in the intersection of $\S_1$ and $\S_2 \oplus \S_3$, the polytope $\alpha \P_1$ reaches $\x$ for a smaller $\alpha$ than $\alpha \P_{-1}$, hence, subspace-sparse recovery holds. Middle: when the subspace angle decreases, the polytope $\alpha \P_{-1}$ reaches $\x$ for a smaller $\alpha$ than $\alpha \P_1$. Right: when the distribution of the data in $\S_1$ becomes nearly degenerate, in this case close to a line, the polytope $\alpha \P_{-1}$ reaches $\x$ for a smaller $\alpha$ than $\alpha \P_1$. In both cases, in the middle and right, the subspace-sparse recovery does not hold for points at the intersecion.}}
\label{fig:L1geometrical}
\end{figure*}
\vspace{1.5mm}
\begin{theorem}
\label{thm:sufficient-condition2}
\emph{Consider a collection of data points drawn from $n$ disjoint subspaces $\{ \S_i \}_{i=1}^n$ of dimensions $\{d_i\}_{i=1}^n$. Let $\mathbb{W}_i$ be the set of all full-rank submatrices $\tilde{\Y}_i \in \Re^{D \times d_i}$ of $\Y_i$, where $\rank(\Y_i) = d_i$. 
If the condition
\begin{equation}
\label{eq:suffcond2}
\max_{\tilde{\Y}_i \in \mathbb{W}_i} \sigma_{d_i}(\tilde{\Y}_i ) > \sqrt{d_i} \, \| \Y_{-i} \|_{1,2} \, \max_{j \neq i} {\cos( \theta_{ij} )}   
\end{equation}
holds, then for every nonzero $\y$ in $\S_i$, the $\ell_1$-minimization in \eqref{eq:L1} recovers a subspace-sparse solution, \ie, $\c^* \neq \0$ and $\c_{-}^* = \0$.\footnote{The induced norm $\| \Y_{-i} \|_{1,2}$ denotes the maximum $\ell_2$-norm of the columns of $\Y_{-i}$.}
}
\end{theorem}
\vspace{1.5mm}
%
%
%
Loosely speaking, the sufficient condition in Theorem \ref{thm:sufficient-condition2} states that if the smallest principal angle between each $\S_i$ and any other subspace is larger than a certain value that depends on the data distribution in $\S_i$, then the subspace-sparse recovery holds. This bound can be rather high when the norms of the data points are oddly distributed, \eg, when the maximum norm of data points in $\S_i$ is much smaller than the maximum norm of data points in all other subspaces. Since the segmentation of the data does not change when data points are scaled, we can apply SSC to linear subspaces after normalizing the data points to have unit Euclidean norms. In this case, the sufficient condition in \eqref{eq:suffcond2} reduces to
%
\begin{equation}
\label{eq:suffcondnormalized}
\max_{\tilde{\Y}_i \in \mathbb{W}_i} \sigma_{d_i}(\tilde{\Y}_i ) > \sqrt{d_i} \, \max_{j \neq i} {\cos( \theta_{ij} )}.   
\end{equation}
%
%
%
\begin{remark}
For independent subspaces, the intersection of a subspace with the direct sum of other subspaces is the origin, hence, the condition in \eqref{eq:suffcondgeneral} always holds. As a result, from Theorem \ref{thm:sufficient-condition}, the $\ell_1$-minimization always recovers subspace-sparse representations of data points in independent subspaces.
\end{remark}
\begin{remark}
The condition in \eqref{eq:suffcondgeneral} is closely related to the nullspace property in the sparse recovery literature \cite{DonohoElad:PNAS03,Gribonval:TIT03, Stojnic:TSP09, VandenBerg:TIT10}. The key difference, however, is that we only require the inequality in \eqref{eq:suffcondgeneral} to hold for the optimal solutions of \eqref{eq:L1samesubspace} and \eqref{eq:L1allsubspace} instead of any feasible solution. Thus, while the inequality can be violated for many feasible solutions, it can still hold for the optimal solutions, guaranteeing successful subspace-sparse recovery from Theorem \ref{thm:sufficient-condition}. Thus, our result can be thought of as a generalization of the nullspace property to the multi-subspace setting where the number of points in each subspace is arbitrary.
\end{remark}
%

\subsection{Geometric interpretation}
In this section, we provide a geometric interpretation of the subspace-sparse recovery conditions in \eqref{eq:suffcondgeneral} and \eqref{eq:suffcond2}. To do so, it is necessary to recall the relationship between the $\ell_1$-norm of the optimal solution of 
\begin{equation}
\label{eq:L1polytope}
\min \| \a \|_1 \quad \st \quad \x = \B \a,
\end{equation}
and the symmetrized convex polytope of the columns of $\B$ \cite{Donoho:TechRep05}. More precisely, if we denote the columns of $\B$ by $\b_i$ and define the symmetrized convex hull of the columns of $\B$ by
\begin{equation}
\P \triangleq \operatorname{conv}(\pm \b_1, \pm \b_2, \cdots),
\end{equation}
then the $\ell_1$-norm of the optimal solution of \eqref{eq:L1polytope} corresponds to the smallest $\alpha > 0$ such that the scaled polytope $\alpha \P$ reaches $\x$ \cite{Donoho:TechRep05}. Let us denote the symmetrized convex polytopes of  $\Y_i$ and $\Y_{-i}$ by $\P_i$ and $\P_{-i}$, respectively. Then the condition in \eqref{eq:suffcondgeneral} has the following geometric interpretation: 
\smallskip
\begin{quote}
the subspace-sparse recovery in $\S_i$ holds if and only if for any nonzero $\x$ in the intersection of $\S_i$ and $\oplus_{j \neq i}{\S_j}$, 
$\alpha \P_i$ reaches $\x$ before $\alpha \P_{-i}$, i.e., for a smaller $\alpha$.
\end{quote}
\smallskip

As shown in the left plot of Figure \ref{fig:L1geometrical}, for $\x$ in the intersection of $\S_1$ and $\S_2 \oplus \S_3$, the polytope $\alpha \P_1$ reaches $\x$ before $\alpha \P_{-1}$, hence the subspace-sparse recovery condition holds. On the other hand, when the principal angles between $\S_1$ and other subspaces decrease, as shown in the middle plot of Figure \ref{fig:L1geometrical}, the subspace-sparse recovery condition does not hold since the polytope $\alpha \P_{-1}$ reaches $\x$ before $\alpha \P_1$. Also, as shown in the right plot of Figure \ref{fig:L1geometrical}, when the distribution of the data in $\S_1$ becomes nearly degenerate, in this case close to a $1$-dimensional subspace orthogonal to the direction of $\x$, then the subspace-sparse recovery condition does not hold since $\alpha \P_{-1}$ reaches $\x$ before $\alpha \P_{1}$. Note that the sufficient condition in \eqref{eq:suffcond2} translates the relationship between the polytopes, mentioned above, explicitly in terms of a relationship between the subspace angles and the singular values of the data.

\section{Graph Connectivity}
\label{sec:connectivity}
In the previous section, we studied conditions under which the proposed $\ell_1$-minimization program recovers subspace-sparse representations of data points. As a result, in the similarity graph, the points that lie in different subspaces do not get connected to each other. On the other hand, our extensive experimental results on synthetic and real data show that data points in the same subspace always form a connected component of the graph, hence, for $n$ subspaces the similarity graph has $n$ connected components. \cite{Hartley:CVPR11} has theoretically verified the connectivity of points in the same subspace for $2$ and $3$ dimensional subspaces. However, it has shown that, for subspaces of dimensions greater than or equal to $4$, under odd distribution of the data, it is possible that points in the same subspace form multiple components of the graph.

\begin{figure}[t!]
\centering
\includegraphics[width=0.44\linewidth, trim = 70 42 48 30 , clip]{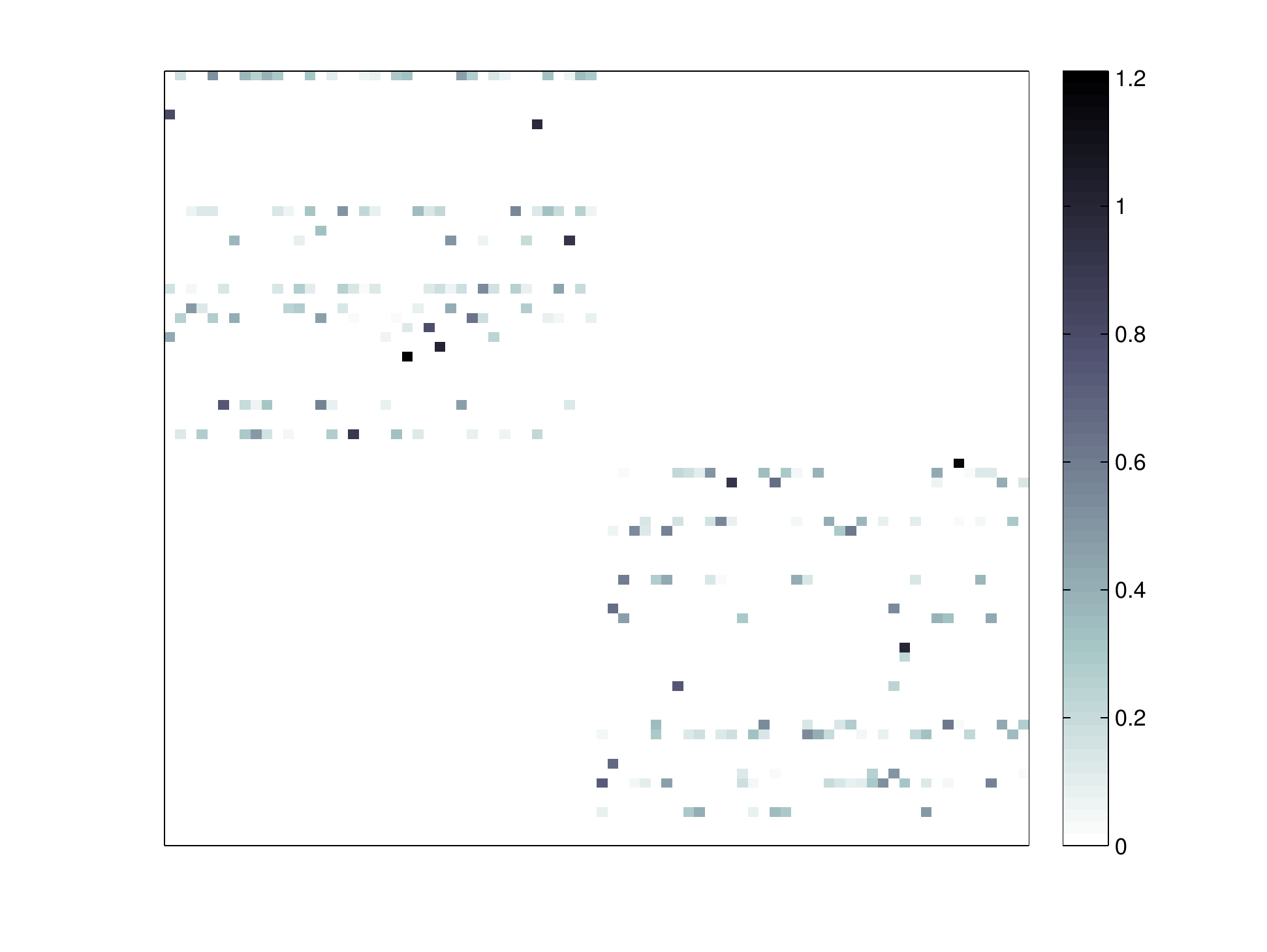}\ \hspace{2mm}
\includegraphics[width=0.44\linewidth, trim = 70 42 48 30 , clip]{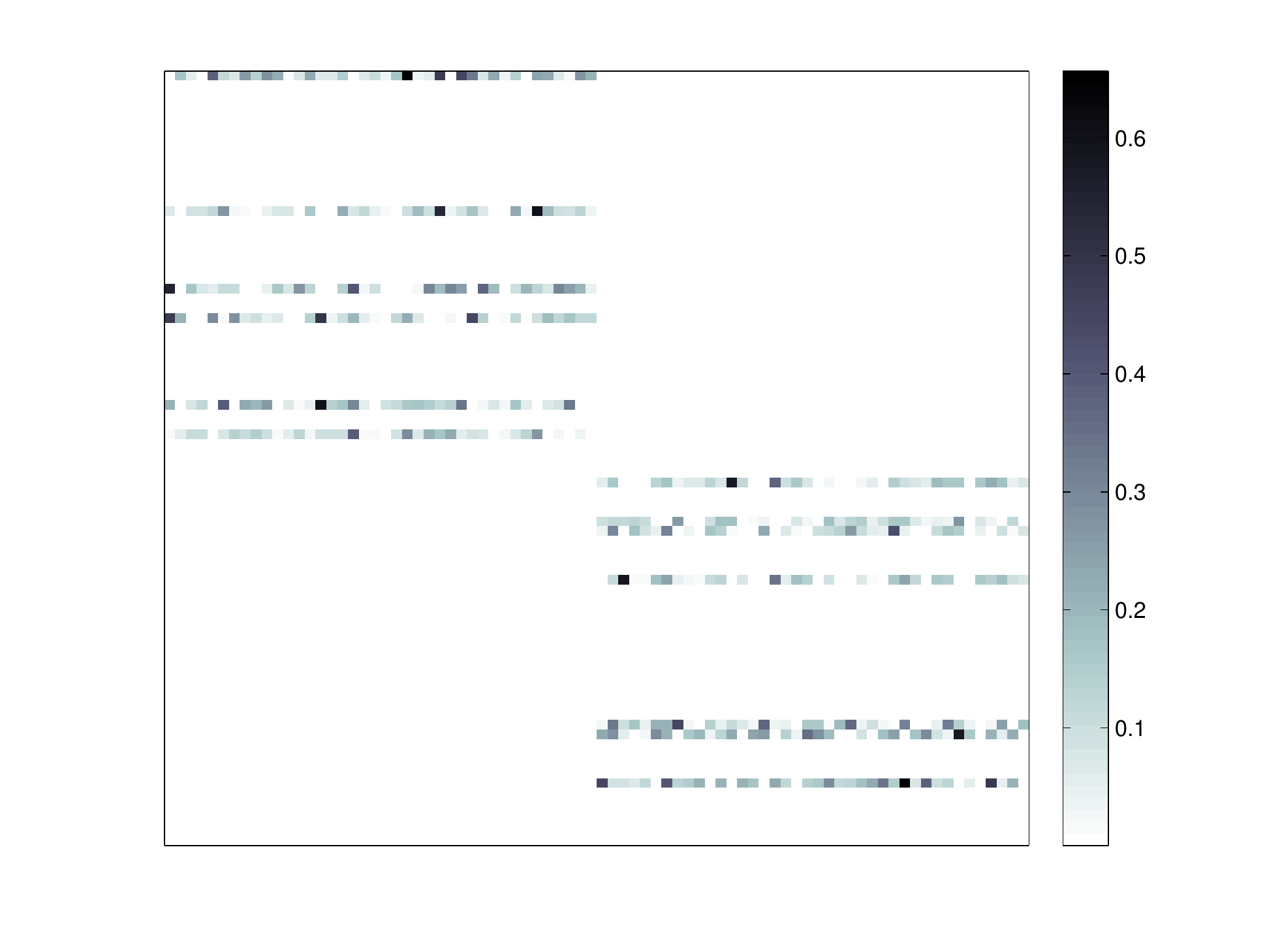}
\vspace{-1mm}
\caption{\small{Coefficient matrix obtained from the solution of \eqref{eq:L1Lrow} for data points in two subspaces. Left: $\lambda_r = 0$. Right: $\lambda_r = 10$. Increasing $\lambda_r$ results in concentration of the nonzero elements in a few rows of the coefficient matrix, hence choosing a few common data points.}}
\label{fig:LLLconnectivity2}
\vspace{-1mm}
\end{figure}
\begin{figure*}[t!]
\centering
\includegraphics[width=0.23\linewidth, trim = 0 -1mm 0 0 , clip]{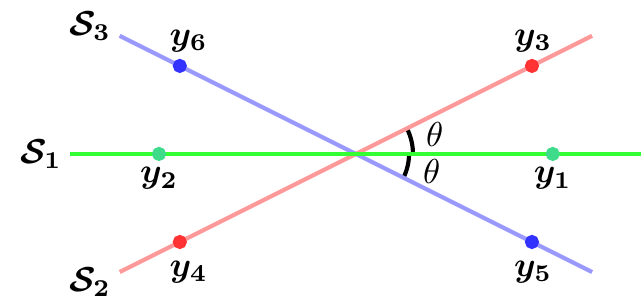}\ \hspace{4mm}
\includegraphics[width=0.27\linewidth, trim = 5 0 0 0 , clip]{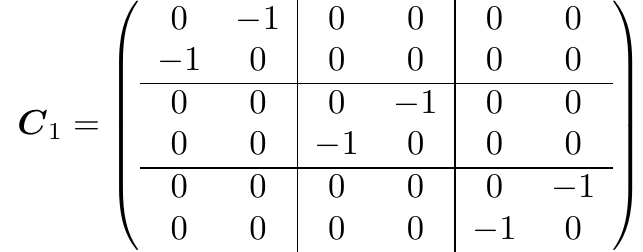}\ \hspace{2mm}
\includegraphics[width=0.325\linewidth, trim = 5 0 0 0, clip]{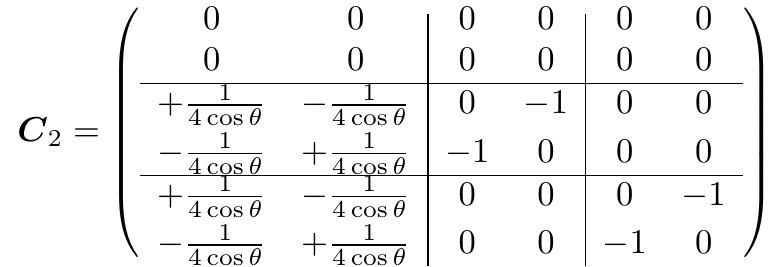}
\vspace{-.7mm}
\caption{\small{Left: three $1$-dimensional subspaces in $\Re^2$ with normalized data points. Middle: $\C_1$ corresponds to the solution of \eqref{eq:L1Lrow} for $\lambda_r=0$. The similarity graph of $\C_1$ has three components corresponding to the three subspaces. Right: $\C_2$ corresponds to the solution of \eqref{eq:L1Lrow} for $\lambda_r \rightarrow +\infty$ and $\theta \in (0,\frac{4 \pi}{10})$. The similarity graph of $\C_2$ has only one connected component.}}
\label{fig:LLLconnectivity}
\end{figure*}

In this section, we consider a regularization term in the sparse optimization program that promotes connectivity of the points within each subspace.\footnote{Another approach to deal with the connectivity issue is to analyze the subspaces corresponding to the components of the graph and merge the components whose associated subspaces have a small distance from each other, \ie, have a small principal angle. However, the result can be sensitive to the choice of the dimension of the subspaces to fit to each component as well as the threshold value on the principal angles to merge the subspaces.} We use the idea that if data points in each subspace choose a few \emph{common} points from the same subspace in their sparse representations, then they form a single component of the similarity graph. Thus, we add to the sparse optimization program the regularization term
\begin{equation}
\label{eq:regularizationL0}
\| \C \|_{r,0} \triangleq \sum_{i=1}^{N}{\operatorname{I}(\| \c^i \|_2 > 0)},
\end{equation}
where $\operatorname{I}(\cdot)$ denotes the indicator function and $\c^i$ denotes the $i$-th row of $\C$. 
Hence, minimizing \eqref{eq:regularizationL0} corresponds to minimizing the number of nonzero rows of $\C$ \cite{Elhamifar:CVPR12, Jenatton:JMLR11, Tropp:SP06}, \ie, choosing a few common data points in the sparse representation of each point. Since a minimization problem that involves \eqref{eq:regularizationL0} is in general NP-hard, we consider its convex relaxation as
\begin{equation}
\label{eq:regularizationL1}
\| \C \|_{r,1} \triangleq \sum_{i=1}^{N}{\| \c^i \|_2}.
\end{equation}
Thus, to increase the connectivity of data points from the same subspace in the similarity graph, we propose to solve
%
\begin{equation}
\label{eq:L1Lrow}
\min \| \C \|_1+ \lambda_r \| \C \|_{r,1} ~~ \st ~~ \Y = \Y \C,~ \diag(\C) = \0,
\end{equation}
where $\lambda_r > 0$ sets the trade-off between the sparsity of the solution and the connectivity of the graph. Figure \ref{fig:LLLconnectivity2} shows how adding this regularization term promotes selecting common points in sparse representations. The following example demonstrates the reason for using the row-sparsity term as a regularizer but not as an objective function instead of the $\ell_1$-norm.

\vspace{1mm}
\begin{example}
Consider the three $1$-dimensional subspaces in $\Re^2$, shown in Figure \ref{fig:LLLconnectivity}, where the data points have unit Euclidean norms and the angle between $\S_1$ and $\S_2$ as well as between $\S_1$ and $\S_3$ is equal to $\theta$. Note that in this example, the sufficient condition in \eqref{eq:suffcond2} holds for all values of $\theta \in (0,\frac{\pi}{2})$. As a result, the solution of \eqref{eq:L1Lrow} with $\lambda_r = 0$ recovers a subspace-sparse representation for each data point, which in this example is uniquely given by $\C_1$ shown in Figure \ref{fig:LLLconnectivity}. Hence, the similarity graph has exactly $3$ connected components corresponding to the data points in each subspace. Another feasible solution of \eqref{eq:L1Lrow} is given by $\C_2$, shown in Figure \ref{fig:LLLconnectivity}, where the points in $\S_1$ choose points from $\S_2$ and $\S_3$ in their representations. Hence, the similarity graph has only one connected component. Note that for a large range of subspace angles $\theta \in (0,\frac{4 \pi}{10})$ we have
\begin{equation}
\| \C_2 \|_{r,1} = \sqrt{16 + 2/\cos^2(\theta)} \, < \, \| \C_1 \|_{r,1} = 6.
\end{equation}
As a result, for large values of $\lambda_r$, \ie, when we only minimize the second term of the objective function in \eqref{eq:L1Lrow}, we cannot recover subspace-sparse representations of the data points. This suggests using the row-sparsity regularizer with a small value of $\lambda_r$.
\end{example}

\section{Experiments with Synthetic Data}
\label{sec:syntheticexperiments}
In Section \ref{sec:theory}, we showed that the success of the $\ell_1$-minimization for subspace-sparse recovery depends on the principal angles between subspaces and the distribution of the data in each subspace. 
In this section, we verify this relationship through experiments on synthetic data.

We consider three disjoint subspaces $\{ \S_i \}_{i=1}^{3}$ of the same dimension $d$ embedded in the $D$-dimensional ambient space. To make the problem hard enough so that every data point in a subspace can also be reconstructed as a linear combination of points in other subspaces, we generate subspace bases $\{\U_i \in \Re^{D \times d}\}_{i=1}^3$ such that each subspace lies in the direct sum of the other two subspaces, \ie, $\rank( \begin{bmatrix} \U_1 \!&\! \U_2 \!&\! \U_3 \end{bmatrix} ) = 2 d$. In addition, we generate the subspaces such that the smallest principal angles $\theta_{12}$ and $\theta_{23}$ are equal to $\theta$. Thus, we can verify the effect of the smallest principal angle in the subspace-sparse recovery by changing the value of $\theta$. To investigate the effect of the data distribution in the subspace-sparse recovery, we generate the same number of data points, $N_g$, in each subspace at random and change the value of $N_g$. Typically, as the number of data points in a subspace increases, the probability of the data being close to a degenerate subspace decreases.\footnote{To remove the effect of different scalings of data points, \ie, to consider only the effect of the principal angle and number of points, we normalize the data points.} 

After generating three $d$-dimensional subspaces associated to $(\theta,N_g)$, we solve the $\ell_1$-minimization program in \eqref{eq:L1vec} for each data point and measure two different errors. First, denoting the sparse representation of $\y_i \in \S_{k_i}$ by $\c_i^{\top} \triangleq \begin{bmatrix} \c_{i1}^{\top} & \c_{i2}^{\top} & \c_{i3}^{\top} \end{bmatrix}$, with $\c_{ij}$ corresponding to points in $\S_j$, we measure the \emph{subspace-sparse recovery error} by
%
\begin{equation}
\label{eq:ssrerror}
\text{ssr error} = \frac{1}{3N_g} \sum_{i=1}^{3N_g}{ (1 - \frac{\| \c_{ik_i} \|_1}{\| \c_i \|_1}) } \in [0,1],
\end{equation}
where each term inside the summation indicates the fraction of the $\ell_1$-norm of $\c_i$ that comes from points in other subspaces. The error being zero corresponds to $\y_i$ choosing points only in its own subspace, while the error being equal to one corresponds to $\y_i$ choosing points from other subspaces. Second, after building the similarity graph using the sparse coefficients and applying spectral clustering, we measure the \emph{subspace clustering error} by
\begin{equation}
\label{eq:clusteringerror}
\text{subspace clustering error} = \frac{\# \text{ of misclassified points}}{\text{total } \# \text{ of points}}.
\end{equation}
\begin{figure}[t!]
\centering
\includegraphics[width=0.49\linewidth, trim = 46 2 52 6 , clip]{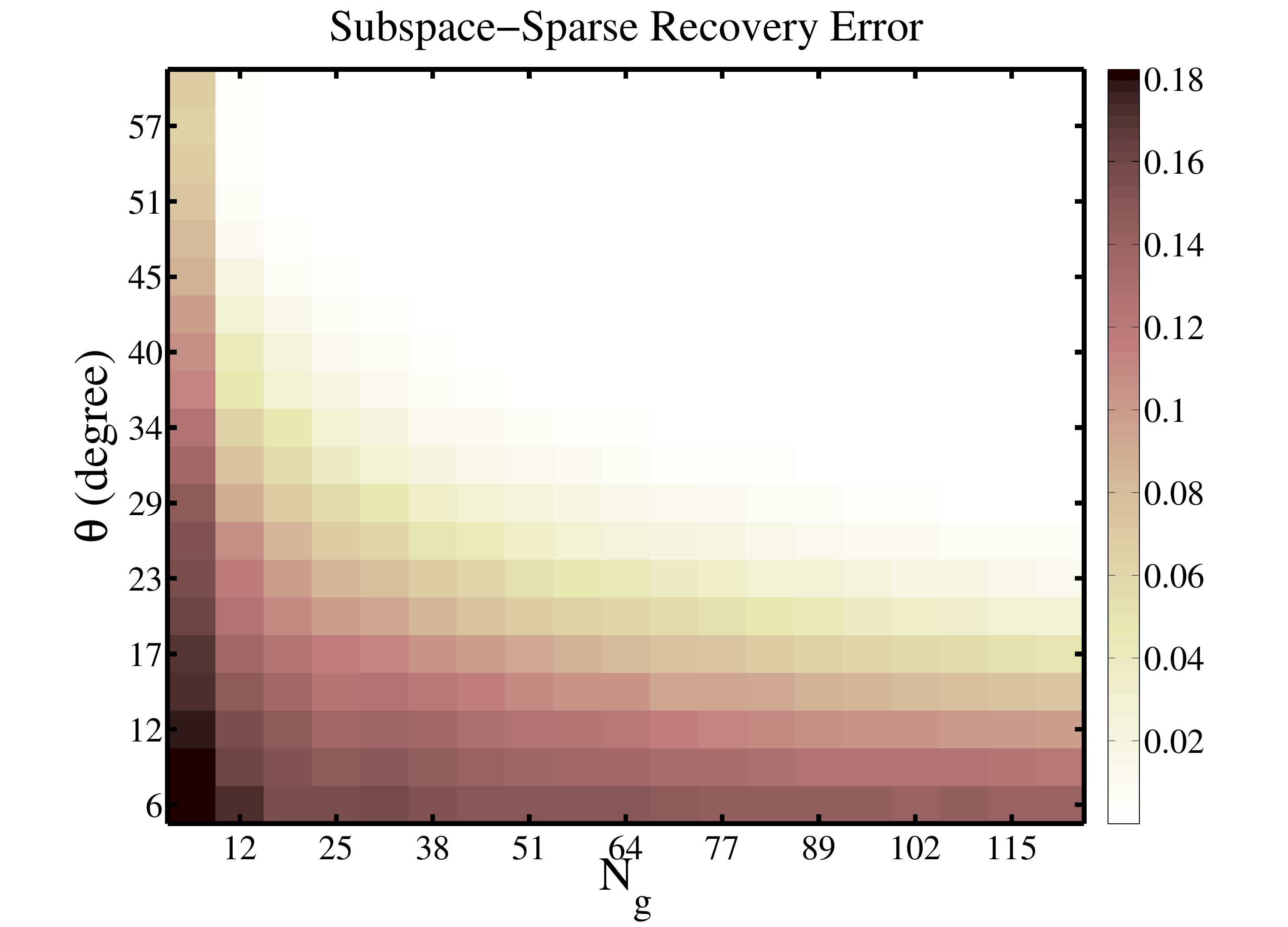}\hspace{0mm}
\includegraphics[width=0.49\linewidth, trim = 46 2 52 6 , clip]{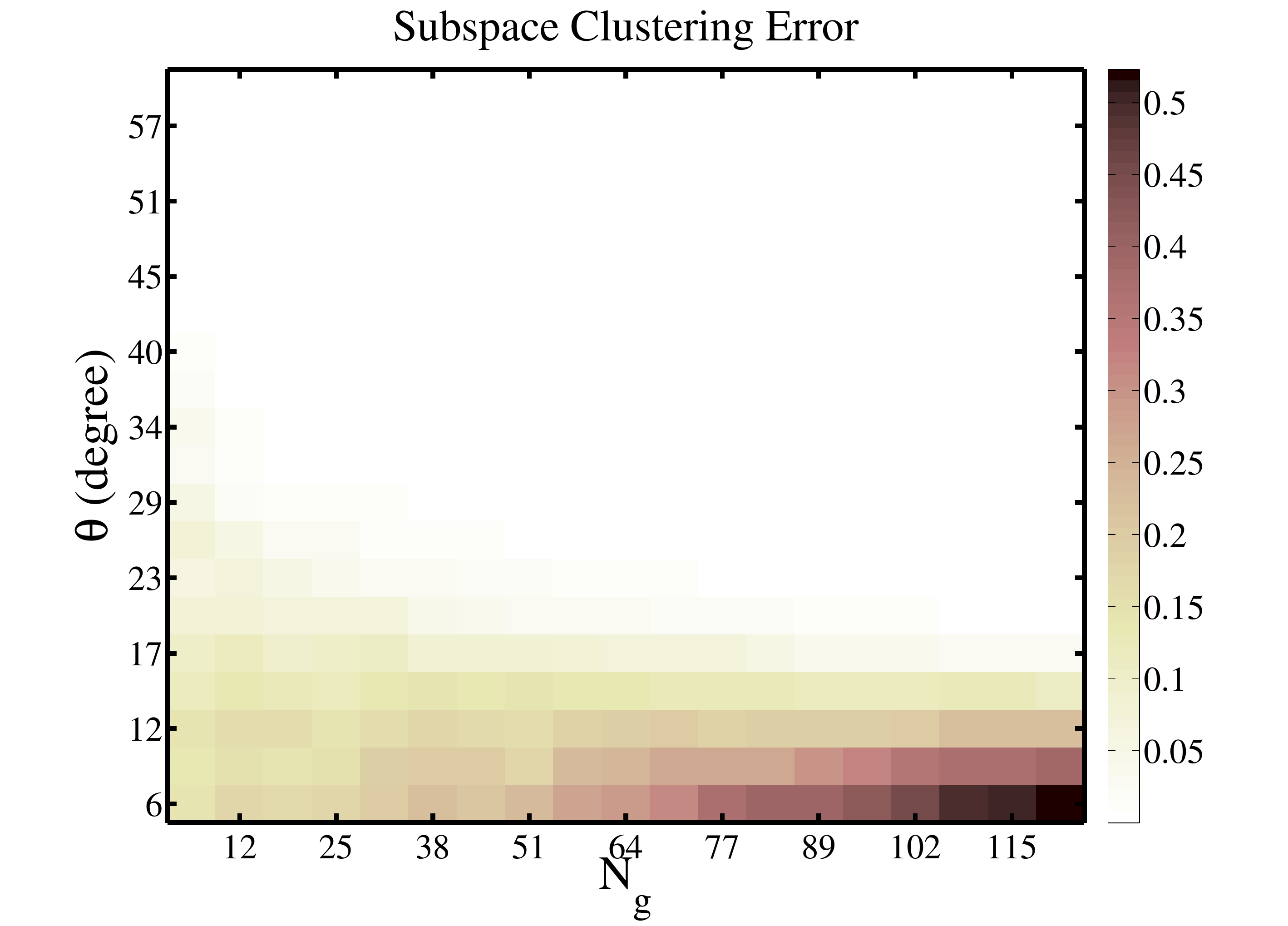}
\vspace{-7.5mm}
\caption{\small{Subspace-sparse recovery error (left) and subspace clustering error (right) for three disjoint subspaces. Increasing the number of points or smallest principal angle decreases the errors.}}
\label{fig:ssr-clustering-error}
\vspace{-1mm}
\end{figure}

In our experiments, we set the dimension of the ambient space to $D = 50$. We change the smallest principal angle between subspaces as 
$\theta \in [6,60]$ degrees and change the number of points in each subspace as $N_g \in [d+1,32d]$. For each pair $(\theta,N_g)$ we compute the average of the errors in \eqref{eq:ssrerror} and \eqref{eq:clusteringerror} over $100$ trials (randomly generated subspaces and data points). The results for $d=4$ are shown in Figure \ref{fig:ssr-clustering-error}. Note that when either $\theta$ or $N_g$ is small, both the subspace-sparse recovery error and the clustering error are large, as predicted by our theoretical analysis. On the other hand, when $\theta$ or $N_g$ increases, the errors decrease, and for $(\theta,N_g)$ sufficiently large we obtain zero errors. 
The results also verify that the success of the clustering relies on the success of the $\ell_1$-minimization in recovering subspace-sparse representations of data points. 
Note that for small $\theta$ as we increase $N_g$, the subspace-sparse recovery error is large and slightly decreases, while the clustering error increases. This is due to the fact that increasing the number of points, the number of undesirable edges between different subspaces in the similarity graph increases, making the spectral clustering more difficult. Note also that, for the values of $(\theta,N_g)$ where the subspace-sparse recovery error is zero, \ie, points in different subspaces are not connected to each other in the similarity graph, the clustering error is also zero. This implies that, in such cases, the similarity graph has exactly three connected components, \ie, data points in the same subspace form a single component of the graph.

\begin{figure}[t!]
\centering
\includegraphics[width=0.485\linewidth, trim = 28 10 45 15 , clip]{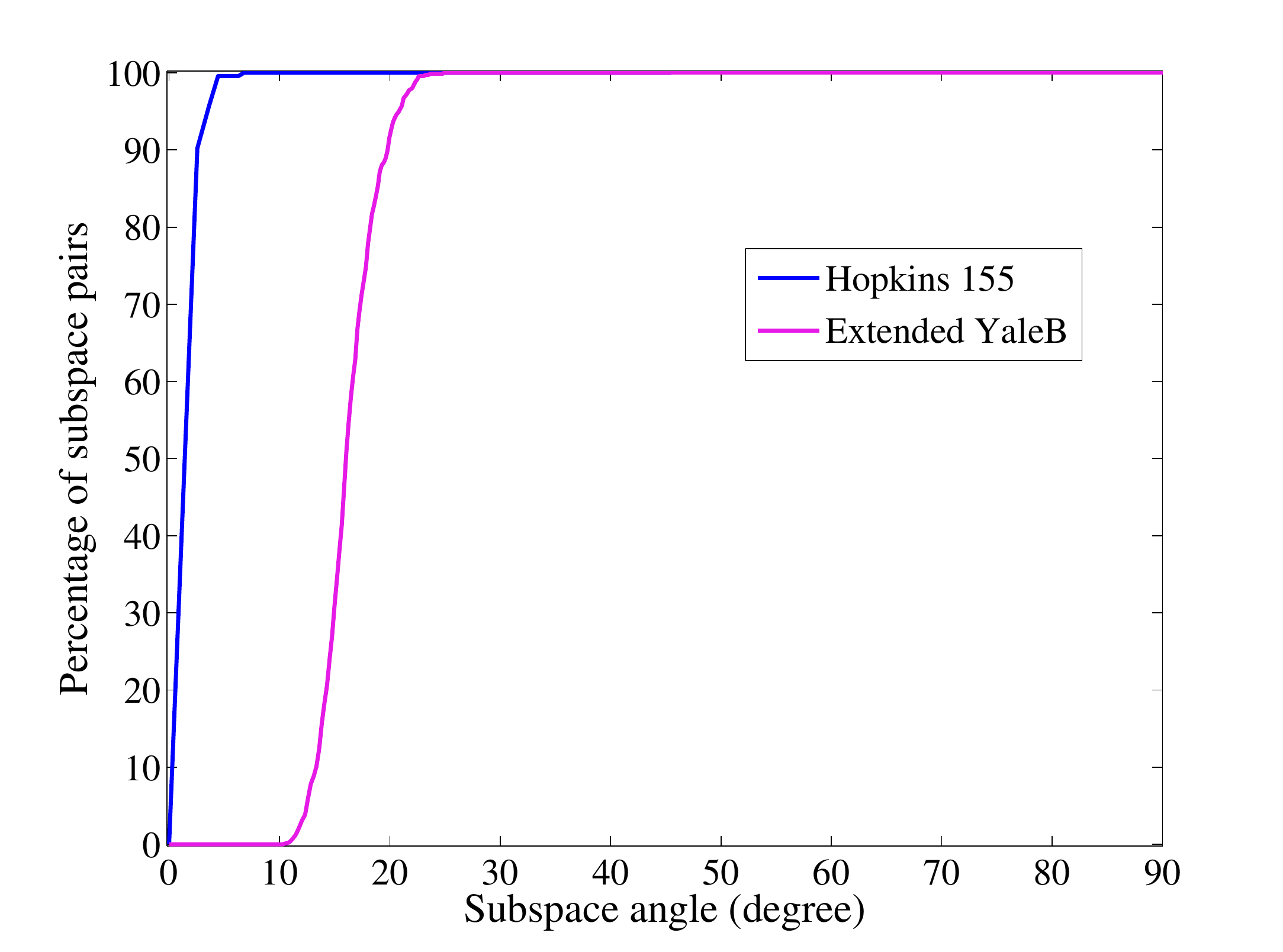}\hspace{0mm}
\includegraphics[width=0.485\linewidth, trim = 28 10 45 15 , clip]{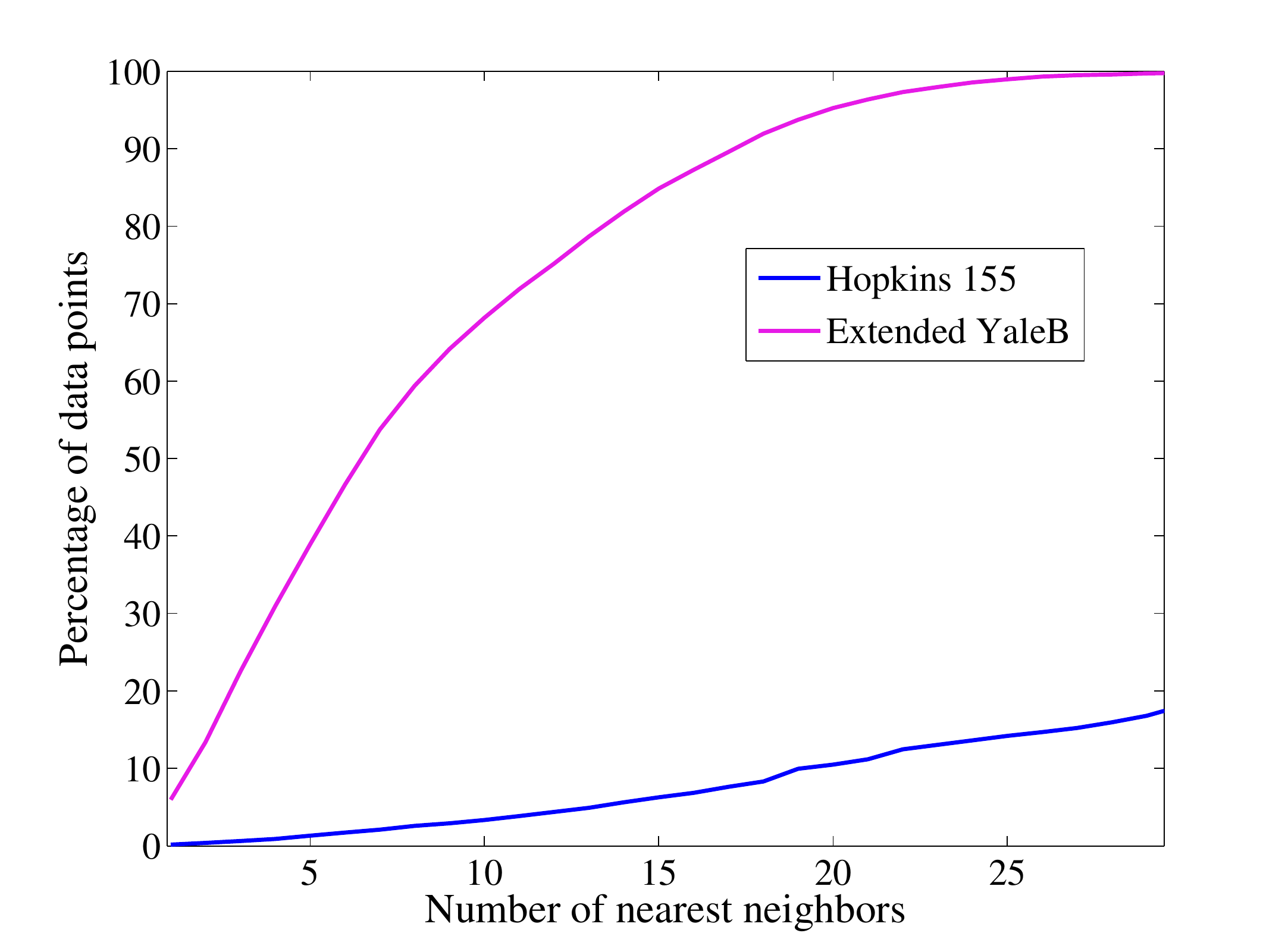}
\vspace{-2mm}
\caption{\small{Left: percentage of pairs of subspaces whose smallest principal angle is smaller than a given value. Right: average percentage of data points in pairs of subspaces that have one or more of their $K$-nearest neighbors in the other subspace.}}
\label{fig:Hopkins-YaleB-Stats}
\vspace{0mm}
\end{figure}
\begin{figure}[t!]
\centering
\includegraphics[width=0.485\linewidth, trim = 35 32 52 5 , clip]{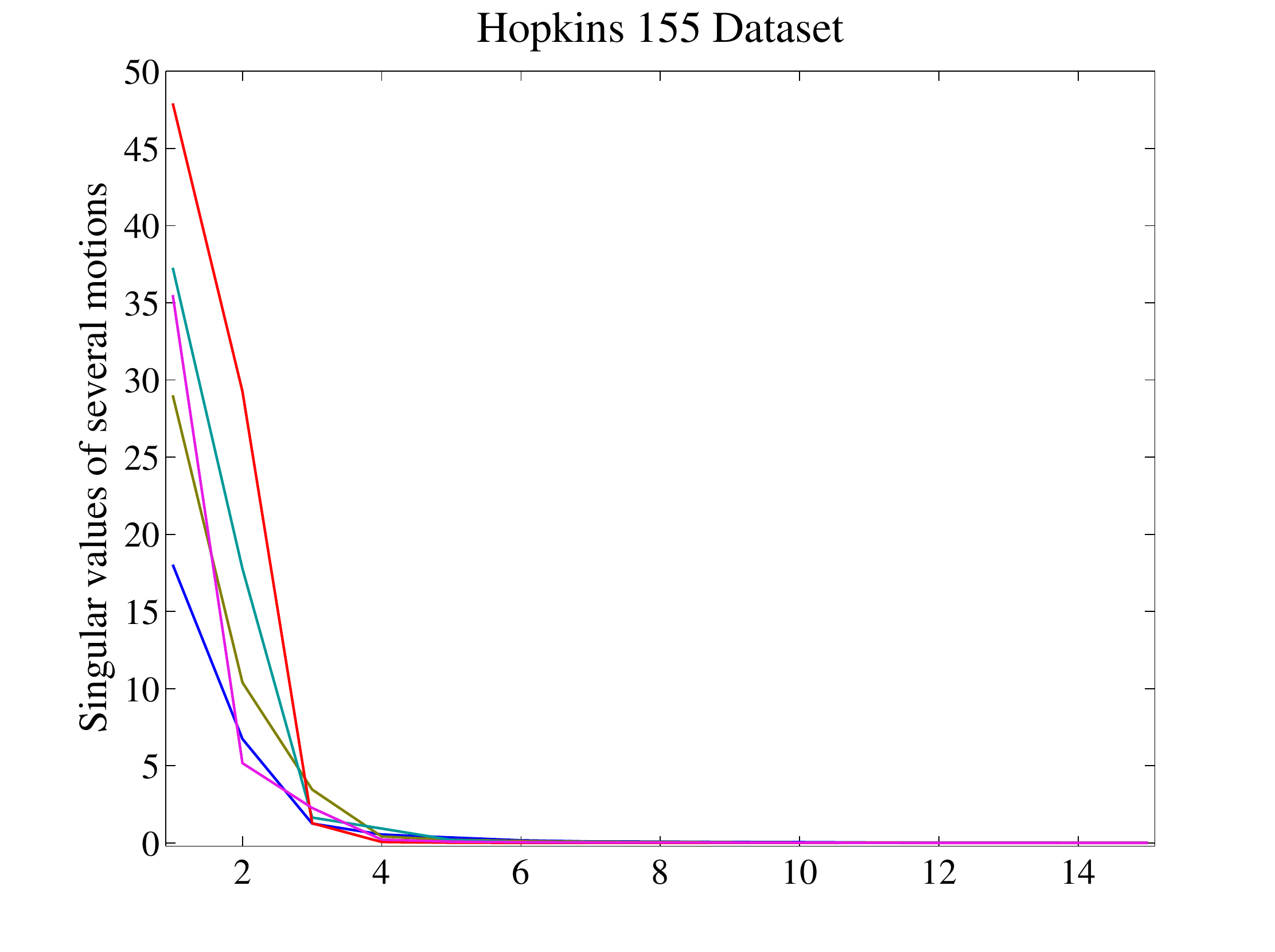}\hspace{0mm}
\includegraphics[width=0.485\linewidth, trim = 35 32 52 5 , clip]{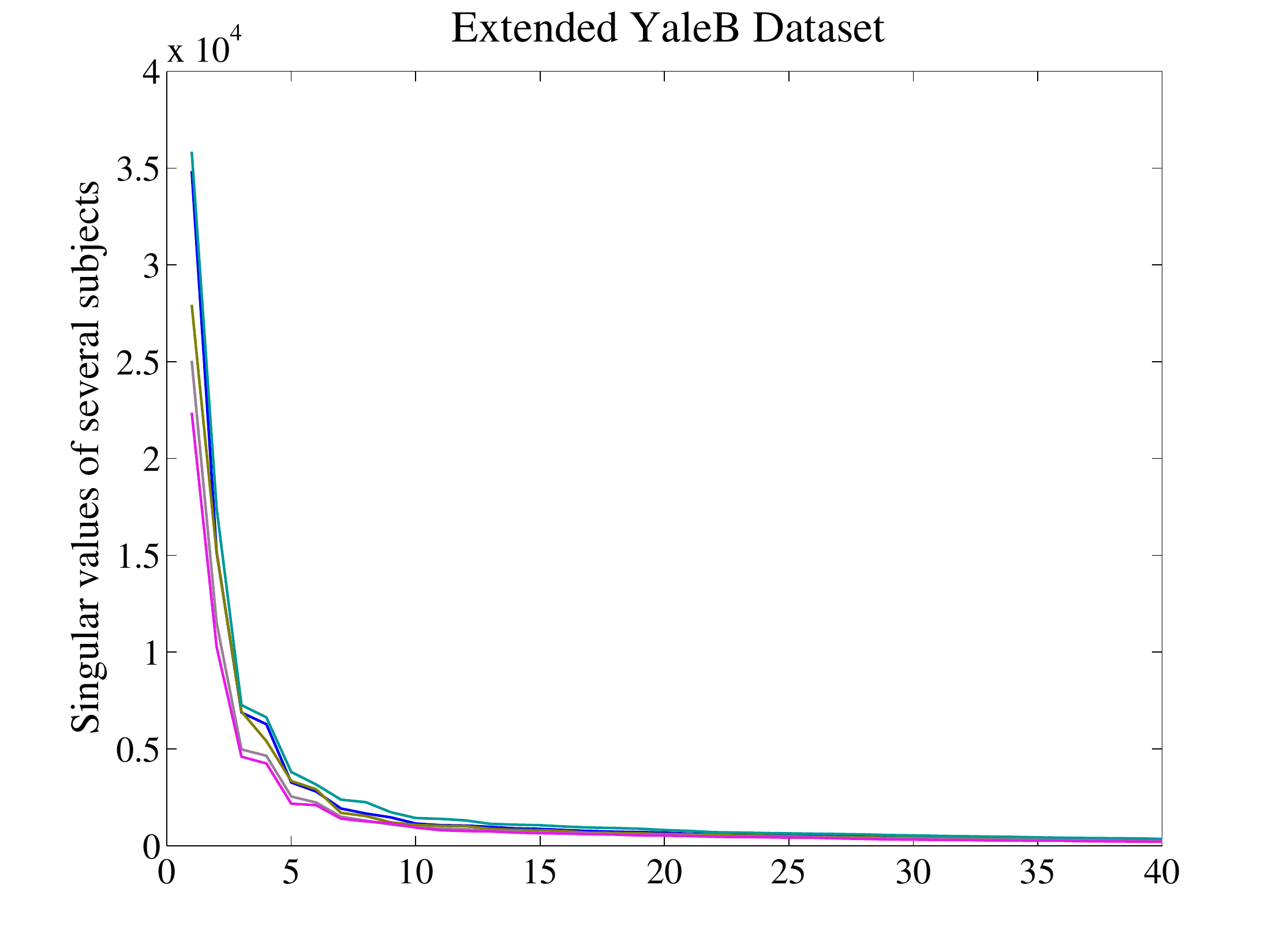}
\vspace{-2mm}
\caption{\small{Left: singular values of several motions in the Hopkins~155~dataset. Each motion corresponds to a subspace of dimension at most $4$. Right: singular values of several faces in the Extended~Yale~B~dataset. Each subject corresponds to a subspace of dimension around $9$.}}
\label{fig:Hopkins-YaleB-SVD}
\vspace{-1mm}
\end{figure}
\section{Experiments with Real Data}
\label{sec:realexperiments}

In this section, we evaluate the performance of the SSC algorithm in dealing with two real-world problems: segmenting multiple motions in videos (Fig. \ref{fig:example-MS}) and clustering images of human faces (Fig. \ref{fig:example-Faces}). We compare the performance of SSC with the best state-of-the-art subspace clustering algorithms: LSA \cite{Yan:ECCV06}, SCC \cite{Chen:IJCV09}, LRR \cite{Liu:ICML10}, and LRSC \cite{Favaro:CVPR11}. 

\myparagraph{Implementation details} We implement the SSC optimization algorithm in \eqref{eq:L1noiseoutlier1} using an Alternating Direction Method of Multipliers (ADMM) framework \cite{Boyd:FTML10, Gabay:CMA76} whose derivation is provided in the supplementary material. 
For the motion segmentation experiments, we use the noisy variation of the optimization program \eqref{eq:L1noiseoutlier1}, \ie, without the term $\E$, with the affine constraint, and choose $\lambda_z = 800/\mu_z$ in all the experiments ($\mu_z$ is defined in \eqref{eq:muz-mue}). For the face clustering experiments, we use the sparse outlying entries variation of the optimization program \eqref{eq:L1noiseoutlier1}, \ie, without the term $\Z$, and choose $\lambda_e = 20/\mu_e$ in all the experiments ($\mu_e$ is defined in \eqref{eq:muz-mue}). It is also worth mentioning that SSC performs better with the ADMM approach than with general interior point solvers \cite{Boyd:STSP07}, which typically return many small nonzero coefficients, degrading the spectral clustering result. 

For the state-of-the-art algorithms, we use the codes provided by their authors. For LSA, we use $K=8$ nearest neighbors and dimension $d=4$, to fit local subspaces, for motion segmentation and use $K=7$ nearest neighbors and dimension $d=5$ for face clustering. For SCC, we use dimension $d=3$, for the subspaces, for motion segmentation and $d=9$ for face clustering. For LRR, we use $\lambda = 4$ for motion segmentation and $\lambda = 0.18$ for face clustering. Note that the LRR algorithm according to \cite{Liu:ICML10}, similar to SSC, applies spectral clustering to a similarity graph built directly from the solution of its proposed optimization program. However, the code of the algorithm applies a heuristic post-processing step, similar to \cite{Lauer:ICCV09}, to the low-rank solution prior to building the similarity graph \cite{Liu:PAMI12}. Thus, to compare the effectiveness of sparse versus low-rank objective function and to investigate the effect of the post-processing step of LRR, we report the results for both cases of without (LRR) and with (LRR-H) the heuristic post-processing step.\footnote{The original published code of LRR contains the function ``compacc.m" for computing the misclassification rate, which is erroneous. We have used the correct code for computing the misclassification rate and as a result, the reported performance for LRR-H is different from the published results in \cite{Liu:ICML10} and \cite{Liu:PAMI12}.} 
For LRSC, we use the method in \cite[Lemma~1]{Favaro:CVPR11} with parameter $\tau=420$ for motion segmentation, and an ALM variant of the method in \cite[Section~4.2]{Favaro:CVPR11} with parameters $\alpha = 3\tau = 0.5*(1.25/\sigma_1(\Y))^2$,  $\gamma =  0.008$ and $\rho = 1.5$ for face clustering. Finally, as LSA and SCC need to know the number of subspaces a priori and the estimation of the number of subspaces from the eigenspectrum of the graph Laplacian in the noisy setting is often unreliable, in order to have a fair comparison, we provide the number of subspaces as an input to all the algorithms. 

\myparagraph{Datasets and some statistics} For the motion segmentation problem, we consider the Hopkins~$155$~dataset \cite{Tron:CVPR07}, which consists of $155$ video sequences of $2$ or $3$ motions corresponding to $2$ or $3$ low-dimensional subspaces in each video \cite{Tomasi:IJCV92, Boult:WMU91}. For the face clustering problem, we consider the Extended~Yale~B~dataset \cite{Kriegman:PAMI05}, which consists of face images of $38$ human subjects, where images of each subject lie in a low-dimensional subspace \cite{Basri:PAMI03}.

Before describing each problem in detail and presenting the experimental results, we present some statistics on the two datasets that help to better understand the challenges of subspace clustering and the performance of different algorithms. First, we compute the smallest principal angle for each pair of subspaces, which in the motion segmentation problem corresponds to a pair of motions in a video and in the face clustering problem corresponds to a pair of subjects. Then, we compute the percentage of the subspace pairs whose smallest principal angle is below a certain value, which ranges from $0$ to $90$ degrees. Figure \ref{fig:Hopkins-YaleB-Stats} (left) shows the corresponding graphs for the two datasets. As shown, subspaces in both datasets have relatively small principal angles. In the Hopkins-$155$~dataset, principal angles between subspaces are always smaller than $10$ degrees, while in the Extended~Yale~B~dataset, principal angles between subspaces are between $10$ and $20$ degrees. Second, for each pair of subspaces, we compute the percentage of data points that have one or more of their $K$-nearest neighbors in the other subspace.
Figure \ref{fig:Hopkins-YaleB-Stats} (right) shows the average percentages over all possible pairs of subspaces in each dataset. 
As shown, in the Hopkins-$155$~dataset for almost all data points, their few nearest neighbors belong to the same subspace. On the other hand, for the Extended~Yale~B~dataset, there is a relatively large number of data points whose nearest neighbors come from the other subspace. This percentage rapidly increases as the number of nearest neighbors increases. As a result, from the two plots in Figure \ref{fig:Hopkins-YaleB-Stats}, we can conclude that in the Hopkins~$155$~dataset the challenge is that subspaces have small principal angles, while in the Extended~Yale~B~dataset, beside the principal angles between subspaces being small, the challenge is that data points in a subspace are very close to other subspaces. 

\begin{table}[t!]
\caption{\footnotesize Clustering error ($\%$) of different algorithms on the Hopkins~155 dataset with the $2F$-dimensional data points.} \centering
\vspace{-1.5mm}
\begin{small}
\begin{tabular}{|@{\;\,}c@{\;\,}|@{\;\,}c@{\;\,}|@{\;\,}c@{\;\,}|@{\;\,}c@{\;\,}|@{\;\,}c@{\;\,}|@{\;\,}c@{\;\,}|@{\;\,}c@{\;\,}|}
\hline
Algorithms & LSA & SCC & LRR & LRR-H & LRSC & SSC\\
\hline
\multicolumn{6}{l}{\textsl{2 Motions}}\\
\hline
Mean &    $4.23$ & $2.89$ & $4.10$ & $2.13$ & $3.69$ & $\textcolor{black}{\textbf{1.52}} \, (2.07)$\\
Median & $0.56$ & $\textcolor{black}{\textbf{0.00}}$ & $0.22$ & $\textcolor{black}{\textbf{0.00}}$ & $0.29$ & $\textcolor{black}{\textbf{0.00}} \, (0.00)$\\
\hline
\multicolumn{6}{l}{\textsl{3 Motions}}\\
\hline
Mean &  $7.02$ & $8.25$ & $9.89$ & $\textcolor{black}{\textbf{4.03}}$ & $7.69$ & $4.40 \, (5.27)$\\
Median & $1.45$ & $\textcolor{black}{\textbf{0.24}}$ & $6.22$ & $1.43$ & $3.80$ & $0.56 \, (0.40)$\\
\hline
\multicolumn{6}{l}{\textsl{All}}\\ 
\hline
Mean &  $4.86$ & $4.10$ & $5.41$ & $2.56$ & $4.59$ & $\textcolor{black}{\textbf{2.18}} \, (2.79)$\\
Median & $0.89$ & $\textcolor{black}{\textbf{0.00}}$ & $0.53$ & $\textcolor{black}{\textbf{0.00}}$ & $0.60$ & $\textcolor{black}{\textbf{0.00}} \, (0.00)$\\
\hline
\end{tabular}
\end{small}
\label{tab:motseg-2F}
\vspace{-2mm}
\end{table}
%

\subsection{Motion Segmentation}

Motion segmentation refers to the problem of segmenting a video sequence of multiple rigidly moving objects into multiple spatiotemporal regions that correspond to different motions in the scene (Fig. \ref{fig:example-MS}). This problem is often solved by extracting and tracking a set of $N$ feature points $\{ \x_{fi} \in \Re^2 \}_{i=1}^{N}$ through the frames $f = 1, \ldots, F$ of the video. Each data point $\y_i$, which is also called a feature trajectory, corresponds to a $2F$-dimensional vector obtained by stacking the feature points $\x_{fi}$ in the video as
\begin{equation}
\label{eq:featuretraj}
\y_i \triangleq \begin{bmatrix} \x_{1i}^{\top} & \x_{2i}^{\top} & \cdots & \x_{Fi}^{\top} \end{bmatrix}^{\top} \in \Re^{2F}.
\end{equation}
Motion segmentation refers to the problem of separating these feature trajectories according to their underlying motions. Under the affine projection model, all feature trajectories associated with a single rigid motion lie in an affine subspace of $\Re^{2F}$ of dimension at most $3$, or equivalently lie in a linear subspace of $\Re^{2F}$ of dimension at most $4$ \cite{Tomasi:IJCV92, Boult:WMU91}. Therefore, feature trajectories of $n$ rigid motions lie in a union of $n$ low-dimensional subspaces of $\Re^{2F}$. Hence, motion segmentation reduces to clustering of data points in a union of subspaces. 

\begin{table}[t!]
\caption{\footnotesize Clustering error ($\%$) of different algorithms on the Hopkins~155 dataset with the $4n$-dimensional data points obtained by applying PCA.} \centering
\vspace{-1.5mm}
\begin{small}
\begin{tabular}{|@{\;\,}c@{\;\,}|@{\;\,}c@{\;\,}|@{\;\,}c@{\;\,}|@{\;\,}c@{\;\,}|@{\;\,}c@{\;\,}|@{\;\,}c@{\;\,}|@{\;\,}c@{\;\,}|}
\hline
Algorithms & LSA & SCC & LRR & LRR-H & LRSC & SSC\\
\hline
\multicolumn{6}{l}{\textsl{2 Motions}}\\ 
\hline
Mean &  $3.61$ & $3.04$ & $4.83$ & $3.41$ & $3.87$ & $\textcolor{black}{\textbf{1.83}} \, (2.14)$\\
Median & $0.51$ & $\textcolor{black}{\textbf{0.00}}$ & $0.26$ & $\textcolor{black}{\textbf{0.00}}$ & $0.26$ & $\textcolor{black}{\textbf{0.00}} \, (0.00)$\\
\hline
\multicolumn{6}{l}{\textsl{3 Motions}}\\ 
\hline
Mean &  $7.65$ & $7.91$ & $9.89$ & $4.86$ & $7.72$ & $\textcolor{black}{\textbf{4.40}} \, (5.29)$\\
Median & $1.27$ & $1.14$ & $6.22$ & $1.47$ & $3.80$ & $\textcolor{black}{\textbf{0.56}} \, (0.40)$\\
\hline
\multicolumn{6}{l}{\textsl{All}}\\ 
\hline 
Mean & $4.52$ & $4.14$ & $5.98$ & $3.74$ & $4.74$ & $\textcolor{black}{\textbf{2.41}} \, (2.85)$\\
Median & $0.57$ & $\textcolor{black}{\textbf{0.00}}$ & $0.59$ & $\textcolor{black}{\textbf{0.00}}$ & $0.58$ & $\textcolor{black}{\textbf{0.00}} \, (0.00)$\\
\hline
\end{tabular}
\end{small}
\label{tab:motseg-4n}
\vspace{-2mm}
\end{table}

In this section, we evaluate the performance of the SSC algorithm as well as that of state-of-the-art subspace clustering methods for the problem of motion segmentation. To do so, we consider the Hopkins~$155$~dataset \cite{Tron:CVPR07} that consists of $155$ video sequences, where $120$ of the videos have two motions and $35$ of the videos have three motions. On average, in the dataset, each sequence of $2$ motions has $N = 266$ feature trajectories and $F = 30$ frames, while each sequence of $3$ motions has $N = 398$ feature trajectories and $F = 29$ frames. The left plot of Figure \ref{fig:Hopkins-YaleB-SVD} shows the singular values of several motions in the dataset. Note that the first four singular values are nonzero and the rest of the singular values are very close to zero, corroborating the $4$-dimensionality of the underlying linear subspace of each motion.\footnote{If we subtract the mean of the data points in each motion from them, the singular values drop at $3$, showing the $3$-dimensionality of the affine subspaces.} In addition, it shows that the feature trajectories of each video can be well modeled as data points that almost perfectly lie in a union of linear subspaces of dimension at most $4$.

The results of applying subspace clustering algorithms to the dataset when we use the original $2F$-dimensional feature trajectories and when we project the data into a $4n$-dimensional subspace ($n$ is the number of subspaces) using PCA are shown in Table \ref{tab:motseg-2F} and Table \ref{tab:motseg-4n}, respectively. From the results, we make the following conclusions:

\smallskip\noindent\textbf{--} In both cases, SSC obtains a small clustering error outperforming the other algorithms. This suggests that the separation of  different motion subspaces in terms of their principal angles and the distribution of the feature trajectories in each motion subspace are sufficient for the success of the sparse optimization program, hence clustering. The numbers inside parentheses show the clustering errors of SSC without normalizing the similarity matrix, \ie, without step 2 in Algorithm \ref{alg:SSC-linear}. Notice that, as explained in Remark \ref{rem:sim-norm}, the normalization step helps to improve the clustering results. However, this improvement is small (about $0.5\%$), \ie, SSC performs well with or without the post-processing of $\C$. 

\smallskip\noindent\textbf{--} Without post-processing of its coefficient matrix, LRR has higher errors than other algorithms. On the other hand, post-processing of the low-rank coefficient matrix significantly improves the clustering performance (LRR-H). 

\smallskip\noindent\textbf{--} LRSC tries to find a noise-free dictionary for data while finding their low-rank representation. This helps to improve over LRR. Also, note that the errors of LRSC are higher than the reported ones in \cite{Favaro:CVPR11}. This comes from the fact that \cite{Favaro:CVPR11} has used the erroneous Ócompacc.mÓ function from [32] to compute the errors.

\smallskip\noindent\textbf{--} The clustering performances of different algorithms when using the $2F$-dimensional feature trajectories or the $4n$-dimensional PCA projections are close. This comes from the fact that the feature trajectories of $n$ motions in a video almost perfectly lie in a $4n$-dimensional linear subspace of the $2F$-dimensional ambient space. Thus, projection using PCA onto a $4n$-dimensional subspace preserves the structure of the subspaces and the data, hence, for each algorithm, the clustering error in Table \ref{tab:motseg-2F} is close to the error in Table \ref{tab:motseg-4n}.

\begin{figure}
\centering
\includegraphics[width=0.59\linewidth, trim = 28 8 38 30 , clip]{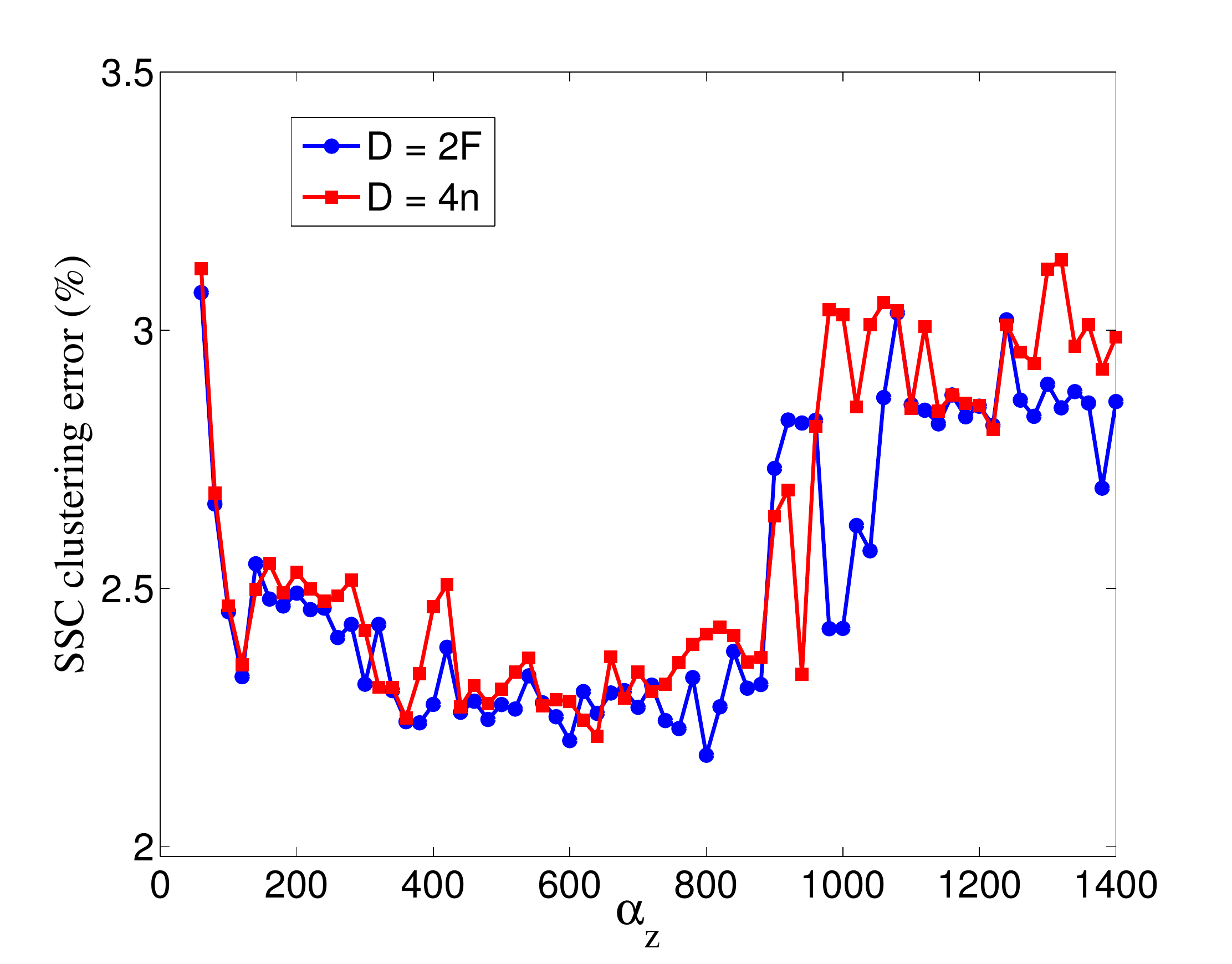}
\vspace{-2mm}
\caption{\small{Clustering error ($\%$) of SSC as a function of $\alpha_z$ in the regularization parameter $\lambda_z = \alpha_z / \mu_z$ for the two cases of clustering of $2F$-dimensional data and $4n$-dimensional data obtained by PCA.}}
\label{fig:ssc-MS-regChange}
\vspace{-2mm}
\end{figure}

In Figure \ref{fig:ssc-MS-regChange} we show the effect of the regularization parameter $\lambda_z = \alpha_z / \mu_z$ in the clustering performance of SSC over the entire Hopkins 155 dataset. Note that the clustering errors of SSC as a function of $\alpha_z$ follow a similar pattern using both the $2F$-dimensional data and the $4n$-dimensional data. Moreover, in both cases the clustering error is less than $2.5\%$ in both cases for a large range of values of $\alpha_z$.

%


Finally, notice that the results of SSC in Tables \ref{tab:motseg-2F}-\ref{tab:motseg-4n} do not coincide with those reported in \cite{Elhamifar:CVPR09}. This is mainly due to the fact in \cite{Elhamifar:CVPR09} we used random projections for dimensionality reduction, while here we use PCA or the original $2F$-dimensional data. In addition, in \cite{Elhamifar:CVPR09} we used a CVX solver to compute a subspace-sparse representation, while here we use an ADMM solver. Also, notice that we have improved the overall clustering error of LSA, for the the case of $4n$-dimensional data, from $4.94\%$, reported in \cite{Tron:CVPR07, Elhamifar:CVPR09}, to $4.52\%$. This is due to using $K=8$ nearest neighbors here instead of $K=5$ in \cite{Tron:CVPR07}.

\subsection{Face Clustering}
Given face images of multiple subjects, acquired with a fixed pose and varying illumination, we consider the problem of clustering images according to their subjects (Fig. \ref{fig:example-Faces}). It has been shown that, under the Lambertian assumption, images of a subject with a fixed pose and varying illumination lie close to a linear subspace of dimension $9$ \cite{Basri:PAMI03}. Thus, the collection of face images of multiple subjects lie close to a union of $9$-dimensional subspaces.

In this section, we evaluate the clustering performance of SSC as well as the state-of-the-art methods on the Extended~Yale~B dataset \cite{Kriegman:PAMI05}. The dataset consists of $192 \times 168$ pixel cropped face images of $n=38$ individuals, where there are $N_i=64$ frontal face images for each subject acquired under various lighting conditions. To reduce the computational cost and the memory requirements of all algorithms, we downsample the images to $48 \times 42$ pixels and treat each $2,016$-dimensional vectorized image as a data point, hence, $D = 2,016$. The right plot in Figure \ref{fig:Hopkins-YaleB-SVD} shows the singular values of  data points of several subjects in the dataset. Note that the singular value curve has a knee around $9$, corroborating the approximate $9$-dimensionality of the face data in each subject. In addition, the singular values gradually decay to zero, showing that the data are corrupted by errors. Thus, the face images of $n$ subjects can be modeled as corrupted data points lying close to a union of $9$-dimensional subspaces.

To study the effect of the number of subjects in the clustering performance of different algorithms, we devise the following experimental setting: we divide the $38$ subjects into $4$ groups, where the first three groups correspond to subjects $1$ to $10$, $11$ to $20$, $21$ to $30$, and the fourth group corresponds to subjects $31$ to $38$. For each of the first three groups we consider all choices of $n \in \{2, 3, 5, 8, 10\}$ subjects and for the last group we consider all choices of $n \in \{2, 3, 5, 8\}$.\footnote{Note that choosing $n$ out of $38$ leads to extremely large number of trials. Thus, we have devised the above setting in order to have a repeatable experiment with a reasonably large number of trials for each $n$.} Finally, we apply clustering algorithms for each trial, \ie, each set of $n$ subjects.

\subsubsection{Applying RPCA separately on each subject}
As shown by the SVD plot of the face data in Figure \ref{fig:Hopkins-YaleB-SVD} (right), the face images do not perfectly lie in a linear subspace as they are corrupted by errors. In fact, the errors correspond to the cast shadows and specularities in the face images and can be modeled as sparse outlying entries. As a result, it is important for a subspace clustering algorithm to effectively deal with data with sparse corruptions. 

In order to validate the fact that corruption of faces is due to sparse outlying errors and show the importance of dealing with corruptions while clustering, we start by the following experiment. We apply the Robust Principal Component Analysis (RPCA) algorithm \cite{Candes:ACM10} to remove the sparse outlying entries of the face data in each subject. Note that \emph{in practice}, we do not know the clustering of the data beforehand, hence cannot apply the RPCA to the faces of each subject. However, as we will show, this experiment illustrates some of the challenges of the face clustering and validates several conclusions about the performances of different algorithms. 

Table \ref{tab:facerec-DSRPCAEach} shows the clustering error of different algorithms after applying RPCA to the data points in each subject and removing the sparse outlying entries, \ie, after bringing the data points back to their low-dimensional subspaces. From the results, we make the following conclusions:

\smallskip\noindent\textbf{--} The clustering error of SSC is very close to zero for different number of subjects suggesting that SSC can deal well with face clustering if the face images are corruption free. In other words, while the data in different subspaces are very close to each other, as shown in Figure \ref{fig:Hopkins-YaleB-Stats} (right), the performance of the SSC is more dependent on the principal angles between subspaces which, while small, are large enough for the success of SSC.

\smallskip\noindent\textbf{--} The LRR and LRSC algorithms have also low clustering errors (LRSC obtains zero errors) showing the effectiveness of removing sparse outliers in the clustering performance. On the other hand, while LRR-H has a low clustering error for $2$, $3$, and $5$ subjects, it has a  relatively large error for $8$ and $10$ subjects, showing that the post processing step on the obtained low-rank coefficient matrix not always improves the result of LRR.

\smallskip\noindent\textbf{--} For LSA and SCC, the clustering error is relatively large and the error increases as the number of subjects increases. This comes from the fact that, as shown in Figure \ref{fig:Hopkins-YaleB-Stats} (right), for face images, the neighborhood of each data point contains points that belong to other subjects and, in addition, the number of neighbors from other subjects increases as we increase the number of subjects.

\begin{table}[t!]
\caption{\footnotesize Clustering error ($\%$) of different algorithms on the Extended~Yale~B dataset after applying RPCA separately to the data points in each subject.} \centering
\vspace{-1.5mm}
\begin{small}
\begin{tabular}{|@{\;\,}c@{\;\,}|@{\;\,}c@{\;\,}|@{\;\,}c@{\;\,}|@{\;\,}c@{\;\,}|@{\;\,}c@{\;\,}|@{\;\,}c@{\;\,}|@{\;\,}c@{\;\,}|}
\hline
Algorithm & LSA & SCC & LRR & LRR-H & LRSC & SSC\\
\hline
\multicolumn{6}{l}{\textsl{2 Subjects}}\\
\hline
Mean &  $6.15$ & $1.29$ & $0.09$ & $0.05$ & $\textcolor{black}{\textbf{0.00}}$ & $0.06$\\
Median &  $\textcolor{black}{\textbf{0.00}}$ & $\textcolor{black}{\textbf{0.00}}$ & $\textcolor{black}{\textbf{0.00}}$ & $\textcolor{black}{\textbf{0.00}}$ & $\textcolor{black}{\textbf{0.00}}$ & $\textcolor{black}{\textbf{0.00}}$\\
\hline
\multicolumn{6}{l}{\textsl{3 Subjects}}\\
\hline
Mean &  $11.67$ & $19.33$ & $0.12$ & $0.10$  & $\textcolor{black}{\textbf{0.00}}$ & $0.08$\\
Median &  $2.60$ & $8.59$ & $\textcolor{black}{\textbf{0.00}}$ & $\textcolor{black}{\textbf{0.00}}$  & $\textcolor{black}{\textbf{0.00}}$ & $\textcolor{black}{\textbf{0.00}}$\\
\hline 
\multicolumn{6}{l}{\textsl{5 Subjects}}\\
\hline
Mean &  $21.08$ & $47.53$ & $0.16$ & $0.15$  & $\textcolor{black}{\textbf{0.00}}$ & $0.07$\\
Median &  $19.21$ & $47.19$ & $\textcolor{black}{\textbf{0.00}}$ & $\textcolor{black}{\textbf{0.00}}$  & $\textcolor{black}{\textbf{0.00}}$ & $\textcolor{black}{\textbf{0.00}}$\\
\hline 
\multicolumn{6}{l}{\textsl{8 Subjects}}\\
\hline
Mean &  $30.04$ & $64.20$ & $4.50$ & $11.57$  & $\textcolor{black}{\textbf{0.00}}$ & $0.06$\\
Median &  $29.00$ & $63.77$ & $0.20$ & $15.43$  & $\textcolor{black}{\textbf{0.00}}$ & $\textcolor{black}{\textbf{0.00}}$\\
\hline
\multicolumn{6}{l}{\textsl{10 Subjects}}\\
\hline 
Mean &  $35.31$ & $63.80$ & $0.15$ & $13.02$  & $\textcolor{black}{\textbf{0.00}}$ & $0.89$\\
Median &  $30.16$ & $64.84$ & $\textcolor{black}{\textbf{0.00}}$ & $13.13$  & $\textcolor{black}{\textbf{0.00}}$ & $0.31$\\
\hline
\end{tabular}
\end{small}
\label{tab:facerec-DSRPCAEach}
\vspace{-2mm}
\end{table}

\subsubsection{Applying RPCA simultaneously on all subjects}
In practice, we cannot apply RPCA separately to the data in each subject because the clustering is unknown. In this section, we deal with sparse outlying entries in the data by applying the RPCA algorithm to the collection of all data points for each trial prior to clustering. The results are shown in Table \ref{tab:facerec-DSRPCA} from which we make the following conclusions:

\smallskip\noindent\textbf{--} The clustering error for SSC is low for all different number of subjects. Specifically, SSC obtains $2.09\%$ and $11.46\%$ for clustering of data points in $2$ and $10$ subjects, respectively.     

\smallskip\noindent\textbf{--} Applying RPCA to all data points simultaneously may not be as effective as applying RPCA to data points in each subject separately. This comes from the fact that RPCA tends to bring the data points into a common low-rank subspace which can result in decreasing the principal angles between subspaces and decreasing the distances between data points in different subjects. This can explain the increase in the clustering error of all clustering algorithms with respect to the results in Table \ref{tab:facerec-DSRPCAEach}.

\begin{table}[t!]
\caption{\footnotesize Clustering error ($\%$) of different algorithms on the Extended~Yale~B dataset after applying RPCA simultaneously  to all the data in each trial.} \centering
\vspace{-1.5mm}
\begin{small}
\begin{tabular}{|@{\;\,}c@{\;\,}|@{\;\,}c@{\;\,}|@{\;\,}c@{\;\,}|@{\;\,}c@{\;\,}|@{\;\,}c@{\;\,}|@{\;\,}c@{\;\,}|@{\;\,}c@{\;\,}|}
\hline
Algorithm & LSA & SCC & LRR &  LRR-H  & LRSC & SSC\\
\hline
\multicolumn{6}{l}{\textsl{2 Subjects}}\\  
\hline
Mean &  $32.53$ & $9.29$ & $7.27$ & $5.72$ & $5.67$ & $\textcolor{black}{\textbf{2.09}}$\\
Median &  $47.66$ & $7.03$ & $6.25$ & $3.91$ & $4.69$ & $\textcolor{black}{\textbf{0.78}}$\\
\hline
\multicolumn{6}{l}{\textsl{3 Subjects}}\\  
\hline
Mean &  $53.02$ & $32.00$ & $12.29$ & $10.01$ & $8.72$ & $\textcolor{black}{\textbf{3.77}}$\\
Median &  $51.04$ & $37.50$ & $11.98$ & $9.38$ & $8.33$ & $\textcolor{black}{\textbf{2.60}}$\\
\hline
\multicolumn{6}{l}{\textsl{5 Subjects}}\\   
\hline
Mean &  $58.76$ & $53.05$ & $19.92$ & $15.33$ & $10.99$ & $\textcolor{black}{\textbf{6.79}}$\\
Median &  $56.87$ & $51.25$ & $19.38$ & $15.94$ & $10.94$ & $\textcolor{black}{\textbf{5.31}}$\\
\hline
\multicolumn{6}{l}{\textsl{8 Subjects}}\\   
\hline
Mean &  $62.32$ & $66.27$ & $31.39$ & $28.67$ & $16.14$ & $\textcolor{black}{\textbf{10.28}}$\\
Median &  $62.50$ & $64.84$ & $33.30$ & $31.05$ & $14.65$ & $\textcolor{black}{\textbf{9.57}}$\\
\hline
\multicolumn{6}{l}{\textsl{10 Subjects}}\\   
\hline
Mean &  $62.40$ & $63.07$ & $35.89$ & $32.55$ & $21.82$ & $\textcolor{black}{\textbf{11.46}}$\\
Median &  $62.50$ & $60.31$ & $34.06$ & $30.00$ & $25.00$ & $\textcolor{black}{\textbf{11.09}}$\\
\hline
\end{tabular}
\end{small}
\label{tab:facerec-DSRPCA}
\vspace{-1mm}
\end{table}

\subsubsection{Using original data points}
Finally, we apply the clustering algorithms to the original data points without pre-processing the data. The results are shown in Table \ref{tab:facerec-DS} from which we make the following conclusions:

 \smallskip\noindent\textbf{--} The SSC algorithm obtains a low clustering error for all numbers of subjects, obtaining $1.86\%$ and $10.94\%$ clustering error for $2$ and $10$ subjects, respectively. In fact, the error is smaller than when applying RPCA to all data points. 
This is due to the fact that SSC directly incorporates the corruption model of the data by sparse outlying entries into the sparse optimization program, giving it the ability to perform clustering on the corrupted data.
 
 \smallskip\noindent\textbf{--} While LRR also has a regularization term to deal with the corrupted data, the clustering error is relatively large especially as the number of subjects increases. This can be due to the fact that there is not a clear relationship between corruption of each data point and the LRR regularization term in general \cite{Liu:ICML10}. On the other hand, the post processing step of LRR-H on the low-rank coefficient matrix helps to significantly reduce the clustering error, although it is larger than the SSC error. 

 \smallskip\noindent\textbf{--} As LRSC tries to recover error-free data points while finding their low-rank representation, it obtains smaller errors than LRR.
 
 \smallskip\noindent\textbf{--} LSA and SCC do not have an explicit way to deal with corrupted data. This together with the fact that the face images of each subject have relatively a large number of neighbors in other subjects, as shown in Figure \ref{fig:Hopkins-YaleB-Stats} (right), result in low performances of these algorithms.

\begin{table}[t!]
\caption{\footnotesize Clustering error ($\%$) of different algorithms on the Extended~Yale~B dataset without pre-processing the data.} \centering
\vspace{-1.5mm}
\begin{small}
\begin{tabular}{|@{\;\,}c@{\;\,}|@{\;\,}c@{\;\,}|@{\;\,}c@{\;\,}|@{\;\,}c@{\;\,}|@{\;\,}c@{\;\,}|@{\;\,}c@{\;\,}|@{\;\,}c@{\;\,}|}
\hline
Algorithm & LSA & SCC & LRR & LRR-H & LRSC & SSC\\
\hline
\multicolumn{6}{l}{\textsl{2 Subjects}}\\
\hline
Mean &  $32.80$ & $16.62$ & $9.52$ & $2.54$ & $5.32$ & $\textcolor{black}{\textbf{1.86}}$\\
Median &  $47.66$ & $7.82$ & $5.47$ & $0.78$ & $4.69$ & $\textcolor{black}{\textbf{0.00}}$\\
\hline
\multicolumn{6}{l}{\textsl{3 Subjects}}\\
\hline
Mean &  $52.29$ & $38.16$ & $19.52$ & $4.21$ & $8.47$ & $\textcolor{black}{\textbf{3.10}}$\\
Median &  $50.00$ & $39.06$ & $14.58$ & $2.60$ & $7.81$ & $\textcolor{black}{\textbf{1.04}}$\\
\hline
\multicolumn{6}{l}{\textsl{5 Subjects}}\\ 
\hline
Mean &  $58.02$ & $58.90$ & $34.16$ & $6.90$ & $12.24$ & $\textcolor{black}{\textbf{4.31}}$\\
Median &  $56.87$ & $59.38$ & $35.00$ & $5.63$ & $11.25$ & $\textcolor{black}{\textbf{2.50}}$\\
\hline
\multicolumn{6}{l}{\textsl{8 Subjects}}\\  
\hline
Mean &  $59.19$ & $66.11$ & $41.19$ & $14.34$ & $23.72$ & $\textcolor{black}{\textbf{5.85}}$\\
Median &  $58.59$ & $64.65$ & $43.75$ & $10.06$ & $28.03$ & $\textcolor{black}{\textbf{4.49}}$\\
\hline
\multicolumn{6}{l}{\textsl{10 Subjects}}\\  
\hline 
Mean &  $60.42$ & $73.02$ & $38.85$ & $22.92$ & $30.36$ & $\textcolor{black}{\textbf{10.94}}$\\
Median &  $57.50$ & $75.78$ & $41.09$ & $23.59$ & $28.75$ & $\textcolor{black}{\textbf{5.63}}$\\
\hline
\end{tabular}
\end{small}
\label{tab:facerec-DS}
\vspace{-3mm}
\end{table}

\subsubsection{Computational time comparison}
\begin{figure}[h!]
\centering
\includegraphics[width=0.67\linewidth, trim = 5 5 10 28 , clip]{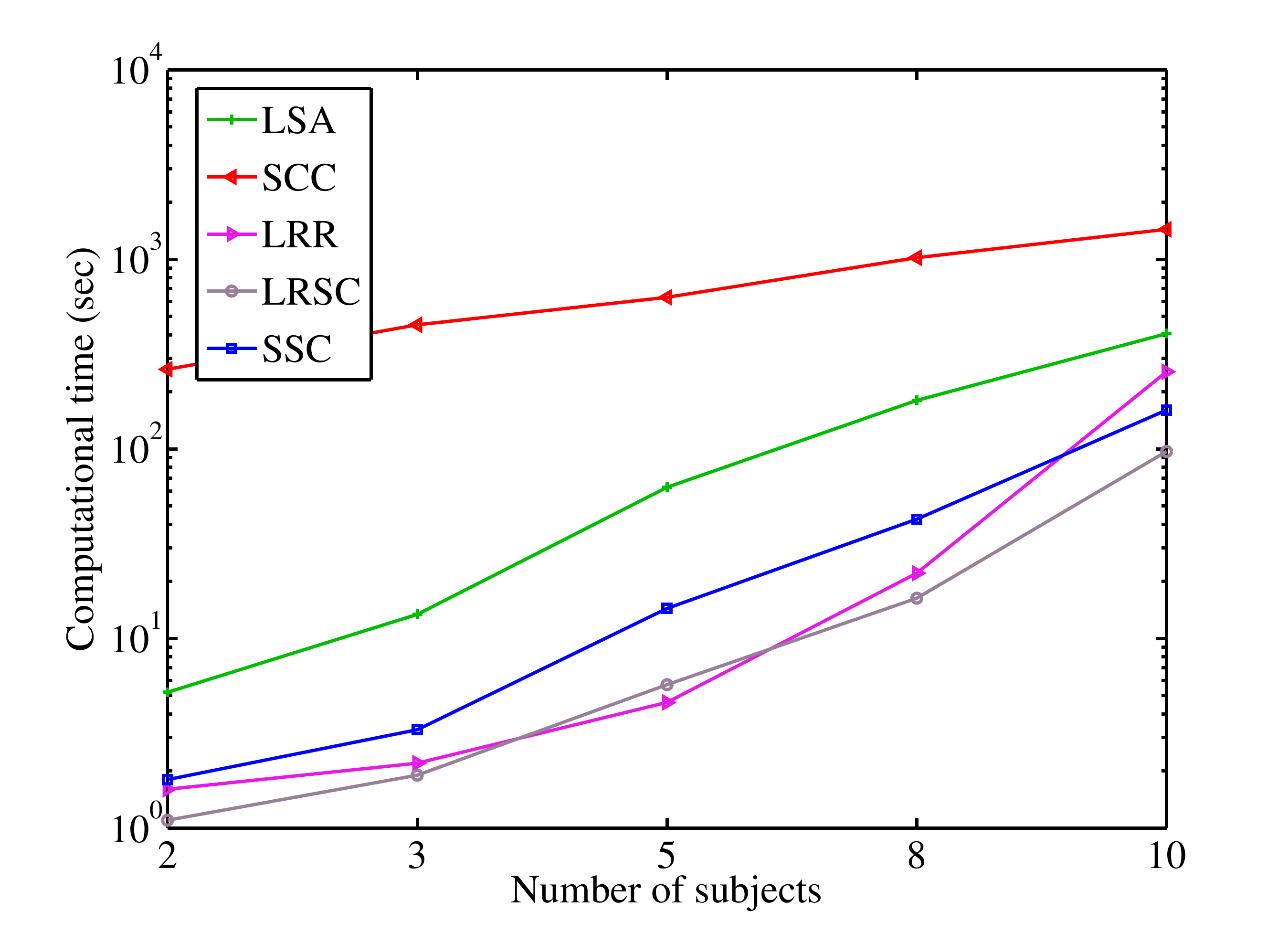}
\vspace{-3mm}
\caption{\small{Average computational time (sec.) of the algorithms on the Extended~Yale~B~dataset as a function of the number of subjects.}}
\label{fig:computation}
\vspace{-3mm}
\end{figure}

The average computational time of each algorithm as a function of the number of subjects (or equivalently the number of data points) is shown in Figure \ref{fig:computation}. Note that the computational time of SCC is drastically higher than other algorithms. This comes from the fact that the complexity of SCC increases exponentially in the dimension of the subspaces, which in this case is $d = 9$. On the other hand, SSC, LRR and LRSC use fast and efficient convex optimization techniques which keeps their computational time lower than other algorithms. 
The exact computational times are provided in the supplementary materials. 

\section{Conclusions  and Future Work}
\label{sec:conclusions}
We studied the problem of clustering a collection of data points that lie in or close to a union of low-dimensional subspaces. We proposed a subspace clustering algorithm based on sparse representation techniques, called SSC, that finds a sparse representation of each point in the dictionary of the other points, builds a similarity graph using the sparse coefficients, and obtains the segmentation of the data using spectral clustering. 
We showed that, under appropriate conditions on the arrangement of subspaces and the distribution of data, the algorithm succeeds in recovering the desired sparse representations of data points. A key advantage of the algorithm is its ability to directly deal with data nuisances, such as noise, sparse outlying entries, and missing entries as well as the more general class of affine subspaces by incorporating the corresponding models into the sparse optimization program. 
Experiments on real data such as face images and motions in videos showed the effectiveness of our algorithm and its superiority over the state of the art. 

Interesting avenues of research, which we are currently investigating, include theoretical analysis of the subspace-sparse recovery in the presence of noise, sparse outlying entries, and missing entries in the data. As our extensive experiments on synthetic and real data show, the points in each subspace, in general, form a single component of the similarity graph. Theoretical analysis of the connectivity of the similarity graph for points in the same subspace in a probabilistic framework would provide a better understanding for this observation. Finally, making the two steps of solving a sparse optimization program and spectral clustering applicable to very large datasets is an interesting and a practical subject for the future work.


%

%
%
%
  \section*{Acknowledgment}
%
The authors would like to thank the financial support of grants NSF-ISS 0447739 and NSF-CSN 0931805.

\ifCLASSOPTIONcaptionsoff
  \newpage
\fi


{
\footnotesize
\bibliographystyle{IEEEtran}
\bibliography{biblio/math,biblio/sparse,biblio/vidal,biblio/learning,biblio/vision,biblio/recognition,biblio/segmentation,biblio/geometry}
}

\medskip
\appendix

\section*{Proof of Proposition \ref{prop:lambdaE_setting}}

In this section, we prove the result of Proposition \ref{prop:lambdaE_setting} in the paper regarding the optimization program
\begin{eqnarray}
\label{eq:L1noiseoutlier1ap}
\begin{split}
&\min \;\; \| \C \|_1 + \lambda_{e} \| \E \|_1 + \frac{\lambda_z}{2} \| \Z \|_F^2  \\ &\st \;\;\; \Y = \Y \C + \E + \Z, \;\; \diag(\C) = \0.
\end{split}
\end{eqnarray}
The result of the proposition suggests to set the regularization parameters as 
\begin{equation}
\lambda_e = \alpha_e / \mu_e, \quad\; \lambda_z = \alpha_z / \mu_z,
\end{equation}
where $\alpha_e , \alpha_z > 1$ and $\mu_e$ and $\mu_z$ are defined as 
\begin{equation}
\mu_e \triangleq \min_{i} \max_{j \neq i} \| \y_j \|_1, \quad\; \mu_z \triangleq \min_{i} \max_{j \neq i} | \y_i^{\top} \y_j |.
\end{equation}
We use the following Lemma in the theoretical proof of the proposition. Proof of this Lemma can be found in \cite{Boyd:STSP07}.

\vspace{1.5mm}
\begin{lemma}
\label{lem:lassoap}
Consider the optimization program
\begin{equation}
\min \;\; \| \c \|_1 + \frac{\lambda}{2} \| \y - \A \c \|_2^2.
\end{equation}
For $\lambda < \| \A^\top \y \|_{\infty}$, we have $\c = \0$.
\end{lemma}

\vspace{1.5mm}
\textbf{Proposition \ref{prop:lambdaE_setting}:}
Consider the optimization program \eqref{eq:L1noiseoutlier1ap}. Without the term $\Z$, if $\lambda_e \leq 1 / \mu_e$, then there exists at least one data point $\y_\ell$ for which in the optimal solution we have $(\c_\ell,\e_\ell) = (\0,\y_\ell)$. Also, without the term $\E$, if $\lambda_z \leq 1 / \mu_z$, then there exists at least one data point $\y_\ell$ for which $(\c_\ell,\z_\ell) = (\0,\y_\ell)$.
\vspace{1.5mm}
\begin{proof}
Note that solving the optimization program \eqref{eq:L1noiseoutlier1ap} is equivalent to solving $N$ optimization programs as
\begin{eqnarray}
\label{eq:L1noiseoutlier1-indvidualap}
\begin{split}
&\min \;\; \| \c_i \|_1 + \lambda_{e} \| \e_i \|_1 + \frac{\lambda_z}{2} \| \z_i \|_2^2  \\ &\st \;\;\; \y_i = \Y \c_i + \e_i + \z_i, \;\; c_{ii} = 0,
\end{split}
\end{eqnarray}
where $\c_i$, $\e_i$, and $\z_i$ are the $i$-th columns of $\C$, $\E$, and $\Z$, repsectively.

\smallskip\noindent
(a) Consider the optimization program \eqref{eq:L1noiseoutlier1-indvidualap} without the term $\z_i$ and denote the objective function value by
\begin{equation}
\operatorname{cost}(\c_i,\e_i) \triangleq \| \c_i \|_1 + \lambda_{e} \| \e_i \|_1.
\end{equation}
Note that a feasible solution of \eqref{eq:L1noiseoutlier1-indvidualap} is given by $(\0,\e_i)$ for which the value of the objective function is equal to 
\begin{equation}
\operatorname{cost}(\0,\e_i)  = \lambda_e \| \y_i \|_1.
\end{equation}
On the other hand, using matrix norm properties, for any feasible solution $(\c_i,\e_i)$ of \eqref{eq:L1noiseoutlier1-indvidualap} we have 
\begin{equation}
\| \y_i \|_1 = \| \Y \c_i + \e_i \|_1 \leq (\max_{j \neq i} \| \y_j \|_1) \, \| \c_i \|_1 + \| \e_i \|_1,
\end{equation}
where we used the fact that $c_{ii} = 0$. Multiplying both sides of the above inequality by $\lambda_e$ we obtain
\begin{equation}
\operatorname{cost}(\0,\y_i) = \lambda_e \| \Y \|_1 \leq (\lambda_e \max_{j \neq i} \| \y_j \|_1) \, \| \c_i \|_1 + \lambda_e \| \e_i \|_1,
\end{equation}
Note that if $\lambda_e < \frac{1}{\max_{j \neq i} \| \y_j \|_1}$, then from the above equation we have
\begin{equation}
\operatorname{cost}(\0,\y_i) \leq \operatorname{cost}(\c_i,\e_i).
\end{equation}
In other words, $(\c_i=\0,\e_i=\y_i)$ achieve the minimum cost among all feasible solutions of \eqref{eq:L1noiseoutlier1-indvidualap}. Hence, if $\lambda_e < \max_{i} \frac{1}{\max_{j \neq i} \| \y_j \|_1}$, then there exists $\ell \in \{1,\cdots, N\}$ such that in the solution of the optimization program \eqref{eq:L1noiseoutlier1ap} we have $(\c_{\ell},\e_\ell) = (\0,\y_\ell)$.

\smallskip\noindent
(b) Consider the optimization program \eqref{eq:L1noiseoutlier1-indvidualap} without the term $\e_i$, which, using $\z_i = \y_i - \Y \c_i$, can be rewritten as
\begin{eqnarray}
\label{eq:L1noiseoutlier1-indvidual-z2ap}
\begin{split}
&\min \;\; \| \c_i \|_1 + \frac{\lambda_z}{2} \| \y_i - \Y \c_i \|_2^2 ~~ \st ~~ c_{ii} = 0.
\end{split}
\end{eqnarray}
From Lemma \ref{lem:lassoap} we have that, for $\lambda_z < \frac{1}{\max_{j \neq i} | \y_j\top \y_i |}$, the solution of \eqref{eq:L1noiseoutlier1-indvidual-z2ap} is equal to $\c_i = 0$, or equivalently, the solution of \eqref{eq:L1noiseoutlier1-indvidualap} is given by $(\c_i,\z_i) = (\0,\y_i)$. As a result, if $\lambda_z < \max_{i} \frac{1}{\max_{j \neq i} | \y_j^{\top} \y_i |}$, then there exists $\ell \in \{1,\cdots, N\}$ such that in the solution of the optimization program \eqref{eq:L1noiseoutlier1ap} we have $(\c_{\ell},\z_\ell) = (\0,\y_\ell)$.

\end{proof}

\section*{Proof of Theorem \ref{thm:independent}}
In this section, we prove Theorem \ref{thm:independent} in the paper, where we showed that for data points in a union of independent subspaces, the solution of the $\ell_q$-minimization recovers subspace-sparse representations of data points.

\vspace{1.5mm}
%
\textbf{Theorem \ref{thm:independent}:}
\emph{Consider a collection of data points drawn from $n$ independent subspaces $\{\S_i\}_{i=1}^n$ of dimensions $\{ d_i \}_{i=1}^{n}$. Let $\Y_i$ denote $N_i$ data points in $\S_i$, where $\rank(\Y_i) = d_i$, and let $\Y_{-i}$ denote data points in all subspaces except $\S_i$. Then, for every $\S_i$ and every nonzero $\y$ in $\S_i$, the $\ell_q$-minimization program
\begin{equation}
\label{eq:Lq2ap}
\begin{bmatrix} \c^* \\ \c^*_{-} \end{bmatrix} = \argmin \left \| \begin{bmatrix} \c \\ \c_{-} \end{bmatrix} \right \|_q ~~ \operatorname{s.t.} ~~ \y = [ \Y_i ~~ \Y_{-i} ] \begin{bmatrix} \c \\ \c_{-} \end{bmatrix},
\end{equation}
for $q < \infty$, recovers a subspace-sparse representation, \ie, $\c^* \neq \0$ and $\c^*_{-} = \0$.
}
\vspace{1.5mm}

\begin{proof}
We prove the result using contradiction. Assume $\c^*_{-} \neq 0$. Then we can write
\begin{equation}
\label{eq:pf1ap}
\y = \Y_i \c^* + \Y_{-i} \c^*_{-}.
\end{equation}
Since $\y$ is a data point in subspace $\S_i$, there exists a $\c$ such that $\y = \Y_i \c$. Substituting this into \eqref{eq:pf1ap} we get
\begin{equation}
\label{eq:pf2ap}
\Y_i (\c - \c^*) = \Y_{-i} \c^*_{-}.
\end{equation}
Note that the left hand side of equation \eqref{eq:pf2ap} corresponds to a point in the subspace $\S_i$ while the right hand side of \eqref{eq:pf2ap} corresponds to a point in the subspace $\oplus_{j \neq i}{\S_j}$. By the independence assumption, the two subspaces $\S_i$ and $\oplus_{j \neq i}{\S_j}$ are also independent hence disjoint and intersect only at the origin. Thus, from \eqref{eq:pf2ap} we must have $\Y_{-i} \c^*_{-} = \0$ and from \eqref{eq:pf1ap} we obtain $\y = \Y_i \c^*$. In other words, $\begin{bmatrix} \c^{*\top} &  \0^{\top} \end{bmatrix}^{\top}$ is a feasible solution of the optimization problem \eqref{eq:Lq2ap}. Finally, from the assumption of $\c^*_{-} \neq 0$, we have
\begin{equation}
\left \| \begin{bmatrix} \c^* \\ \0 \end{bmatrix} \right \|_q < \left \| \begin{bmatrix} \c^* \\ \c^*_{-} \end{bmatrix} \right \|_q
\end{equation}
that contradicts the optimality of $ \begin{bmatrix} \c^{*\top} & \c^*_{-} \end{bmatrix}^{\top}$. Thus, we must have ${\c}^* \neq \0$ and $\c^*_{-} = \0$, obtaining the desired result.
\end{proof}
%

\section*{Proof of Theorem \ref{thm:sufficient-condition}}
In this section, we prove Theorem \ref{thm:sufficient-condition} in the paper, where we provide a necessary and sufficient condition for subspace-sparse recovery in a union of disjoint subspaces. To do so, we consider a vector $\x$ in the intersection of $\S_i$ with $\oplus_{j \neq i}{\S_j}$ and let the optimal solution of the $\ell_1$-minimization when we restrict the dictionary to the points from $\S_i$ be
\begin{equation}
\label{eq:L1samesubspaceap}
\a_i = \argmin \| \a \|_1 \quad \operatorname{s.t.} \quad \x = \Y_i \; \a .
\end{equation}
We also let the optimal solution of the $\ell_1$ minimization when we restrict the dictionary to the points from all subspaces except $\S_i$ be
\begin{equation}
\label{eq:L1allsubspaceap}
\a_{-i} = \argmin \| \a \|_1 \quad \operatorname{s.t.} \quad \x = \Y_{-i} \; \a.
\end{equation}
We show that if for every nonzero $\x$ in the intersection of $\S_i$ with $\oplus_{j \neq i}{\S_j}$, the $\ell_1$-norm of the solution of \eqref{eq:L1samesubspaceap} is strictly smaller than the $\ell_1$-norm of the solution of \eqref{eq:L1allsubspaceap}, \ie, 
\begin{equation}
\label{eq:suffcondgeneralap}
\forall \, \x \in \S_i \cap (\oplus_{j \neq i}{\S_j}), \x \neq \0 \implies \| \a_i \|_1 < \| \a_{-i} \|_1,
\end{equation}
then the SSC algorithm succeeds in recovering subspace-sparse representations of all the data points in $\S_i$.

\vspace{1.5mm}
\textbf{Theorem \ref{thm:sufficient-condition}:}
\emph{Consider a collection of data points drawn from $n$ disjoint subspaces $\{ \S_i \}_{i=1}^n$ of dimensions $\{ d_i \}_{i=1}^{n}$. Let $\Y_i$ denote $N_i$ data points in $\S_i$, where $\rank(\Y_i) = d_i$, and let $\Y_{-i}$ denote data points in all subspaces except $\S_i$. The $\ell_1$-minimization 
\begin{equation}
\label{eq:L1ap}
\begin{bmatrix} \c^* \\ \c^*_{-} \end{bmatrix} = \argmin \left \| \begin{bmatrix} \c \\ \c_{-} \end{bmatrix} \right \|_1 ~~ \operatorname{s.t.} ~~ \y = [ \Y_i ~~ \Y_{-i} ] \begin{bmatrix} \c \\ \c_{-} \end{bmatrix},
\end{equation}
recovers a subspace-sparse representation of every nonzero $\y$ in $\S_i$, \ie, $\c^* \neq \0$ and $\c^*_{-} = \0$, if and only if \eqref{eq:suffcondgeneralap} holds.
}
%
\vspace{1.5mm}
%

%
\begin{proof}
$(\Longleftarrow)$ We prove the result using contradiction. Assume $\c^*_{-} \neq \0$ and define
\begin{equation}
\label{eq:defxap}
\x \triangleq \y - \Y_i \c^* = \Y_{-i} \c^*_{-}.
\end{equation}
Since $\y$ lies in $\S_i$ and $\Y_i \c^*$ is a linear combination of points in $\S_i$, from the first equality in \eqref{eq:defxap} we have that $\x$ is a vector in $\S_i$. Let $\a_i$ be the solution of \eqref{eq:L1samesubspaceap} for $\x$. We have
\begin{equation}
\label{eq:x1ap}
\x = \y - \Y_i \c^* = \Y_i \a_i ~ \Rightarrow ~ \y = \Y_i (\c^* + \a_i).
\end{equation}
On the other hand, since $\Y_{-i} \c^*_{-}$ is a linear combination of points in all subspaces except $\S_i$, from the second equality in \eqref{eq:defxap} we have that $\x$ is a vector in $\oplus_{j \neq i}{\S_j}$. Let $\a_{-i}$ be the solution of \eqref{eq:L1allsubspaceap} for $\x$. We have
\begin{equation}
\label{eq:x2ap}
\x = \Y_{-i} \c^*_{-} = \Y_{-i} \a_{-i} ~ \Rightarrow ~ \y = \Y_i \c^* + \Y_{-i} \a_{-i}.
\end{equation}
Note that the left hand side of \eqref{eq:x2ap} together with the fact that $\a_{-i}$ is the optimal solution of \eqref{eq:L1allsubspaceap} imply that
\begin{equation}
\label{eq:medineqap}
\| \a_{-i} \|_1 \leq  \| \c^*_{-} \|_1.
\end{equation}
From \eqref{eq:x1ap} and \eqref{eq:x2ap} we have that $\begin{bmatrix} \c^*+\a_i \\ \0 \end{bmatrix}$ and $\begin{bmatrix} \c^* \\ \a_{-i} \end{bmatrix}$ are feasible solutions of the original optimization program in \eqref{eq:L1ap}. Thus, we have
\begin{equation}
\left \| \begin{bmatrix} \c^*+\a_i \\ \0 \end{bmatrix} \right \|_1 \! \leq \! \| \c^* \|_1 + \| \a_i \|_1 < \| \c^* \|_1 + \| \a_{-i} \|_1 \leq \left \| \begin{bmatrix} \c^* \\ \c^*_{-}\end{bmatrix} \right \|_1\!\!,
\end{equation}
where the first inequality follows from triangle inequality, the second strict inequality follows from the sufficient condition in \eqref{eq:suffcondgeneralap}, and the last inequality follows from \eqref{eq:medineqap}. This contradicts the optimality of $\begin{bmatrix} \c^{*\top} & \c_{-}^{*\top} \end{bmatrix}^{\top}$ for the original optimization program in \eqref{eq:L1ap}, hence proving the desired result.

\smallskip\noindent $(\Longleftarrow)$ We prove the result using contradiction. Assume the condition in \eqref{eq:suffcondgeneralap} does not hold, \ie, there exists a nonzero $\x$ in the intersection of $\S_i$ and $\oplus_{j \neq i}{\S_j}$ for which we have $\| \a_{-i} \|_1 \leq \| \a_i \|_1$. As a result, for $\y = \x$, a solution of the $\ell_1$-minimization program \eqref{eq:L1ap} corresponds to selecting points from all subspaces except $\S_i$, which contradicts the subspace-sparse recovery assumption.
\end{proof}
%
%

\section*{Proof of Theorem \ref{thm:sufficient-condition2}}

\vspace{1.5mm}
\textbf{Theorem \ref{thm:sufficient-condition2}:}
\emph{Consider a collection of data points drawn from $n$ disjoint subspaces $\{ \S_i \}_{i=1}^n$ of dimensions $\{d_i\}_{i=1}^n$. Let $\mathbb{W}_i$ be the set of all full-rank submatrices $\tilde{\Y}_i \in \Re^{D \times d_i}$ of $\Y_i$, where $\rank(\Y_i) = d_i$. 
If the condition
\begin{equation}
\label{eq:suffcond2ap}
\max_{\tilde{\Y}_i \in \mathbb{W}_i} \sigma_{d_i}(\tilde{\Y}_i ) > \sqrt{d_i} \, \| \Y_{-i} \|_{1,2} \, \max_{j \neq i} {\cos( \theta_{ij} )}   
\end{equation}
holds, then for every nonzero $\y$ in $\S_i$, the $\ell_1$-minimization in \eqref{eq:L1ap} recovers a subspace-sparse solution, \ie, $\c^* \neq \0$ and $\c_{-}^* = \0$.\footnote{$\| \Y_{-i} \|_{1,2}$ denotes the maximum $\ell_2$-norm of the columns of $\Y_{-i}$.}
}
\vspace{1.5mm}

\begin{proof}
We prove the result in two steps. In step 1, we show that $\| \a_i \|_1 \leq \beta_i$. In step 2, we show that $\beta_{-i} \leq \| \a_{-i} \|_1$. Then, the sufficient condition $\beta_i < \beta_{-i}$ establishes the result of the theorem, since it implies
\begin{equation}
\| \a_i \|_1 \leq \beta_i < \beta_{-i} \leq \| \a_{-i} \|_1,
\end{equation}
\ie, the condition of Theorem \ref{thm:sufficient-condition} holds.

\smallskip\noindent\textbf{Step 1:} Upper bound on the $\ell_1$-norm of \eqref{eq:L1samesubspaceap}
Let $\mathbb{W}_i$ be the set of all submatrices $\tilde{\Y}_i \in \Re^{D \times d_i}$ of $\Y_i$ that are full column rank. We can write the vector $\x \in \S_i \cap (\oplus_{j \neq i}{\S_j})$
\begin{align}
\x = \tilde{\Y}_i \tilde{\a} \implies \tilde{\a} = (\tilde{\Y}_i^{\top} \tilde{\Y}_i)^{-1} \tilde{\Y}_i^{\top} \x.
\end{align}
Using vector and matrix norm properties, we have
\begin{multline}
\| \tilde{\a} \|_1 \leq \sqrt{d_i} \| \tilde{\a} \|_2 = \sqrt{d_i} \, \| (\tilde{\Y}_i^{\top} \tilde{\Y}_i)^{-1} \tilde{\Y}_i^{\top} \x \|_2 \\
\leq \sqrt{d_i} \, \| (\tilde{\Y}_i^{\top} \tilde{\Y}_i)^{-1} \tilde{\Y}_i^{\top} \|_{2,2} \| \x \|_2 = \frac{ \sqrt{d_i} } { \sigma_{d_i}(\tilde{\Y}_i) } \; \| \x \|_2 ,
\end{multline}
where $\sigma_{d_i}(\tilde{\Y}_i)$ denotes the $d_i$-th largest singular value of $\tilde{\Y}_i$. Thus, for the solution of the optimization problem in \eqref{eq:L1samesubspaceap}, we have
\begin{equation}
\label{eq:upperbound2ap}
\| \a_i \|_1 \leq \min_{\tilde{\Y}_i \in \mathbb{W}_i } \| \tilde{\a} \|_1 \leq \min_{\tilde{\Y}_i \in \mathbb{W}_i} \, \frac{ \sqrt{d_i} } { \sigma_{d_i}(\tilde{\Y}_i) }  \| \x \|_2 \triangleq \beta_i,
\end{equation}
which established the upper bound on the $\ell_1$-norm of the solution of the optimization program in \eqref{eq:L1samesubspaceap}.

\smallskip\noindent\textbf{Step 2:}{ Lower bound on the $\ell_1$-norm of \eqref{eq:L1allsubspaceap}}
For the solution of \eqref{eq:L1allsubspaceap} we have $\x = \Y_{-i} \a_{-i}$. If we multiply both sides of this equation from left by $\x^{\top}$, we get
\begin{equation}
\| \x \|_2^2 = \x^{\top} \x = \x^{\top} \Y_{-i} \a_{-i}.
\end{equation}
Applying the Holder's inequality $( |\u^{\top} \v| \leq \| \u \|_{\infty} \| \v \|_1 )$ to the above equation, we obtain
\begin{equation}
\| \x \|_2^2 \leq \| \Y_{-i}^{\top} \x \|_{\infty} \| \a_{-i} \|_1.
\end{equation}
By recalling the definition of the smallest principal angle between two subspaces, we can write
\begin{equation}
\label{eq:eqholderap}
\| \x \|_2^2 \leq \max_{j \neq i} \cos(\theta_{ij}) \, \| \Y_{-i} \|_{1,2} \, \| \x \|_2 \, \| \a_{-i} \|_1,
\end{equation}
where $\theta_{ij}$ is the first principal angle between $\S_i$ and $\S_j$ and $\| \Y_{-i} \|_{1,2}$ is the maximum $\ell_2$-norm of the columns of $\Y_{-i}$, \ie, data points in all subspaces except $\S_i$. We can rewrite \eqref{eq:eqholderap} as
\begin{equation}
\label{eq:lowerboundap}
\beta_{-i} \triangleq \frac{\| \x \|_2 }{ \max_{j \neq i} {\cos( \theta_{ij} )} \;  \| \Y_{-i} \|_{1,2} } \leq \| \a_{-i} \|_1
\end{equation}
which establishes the lower bound on the $\ell_1$ norm of the solution of the optimization program in \eqref{eq:L1allsubspaceap}.
\end{proof}
\vspace{1mm}
%

\section*{Solving the Sparse Optimization Program}

Note that the proposed convex programs can be solved using generic convex solvers such as CVX\footnote{CVX is a Matlab-based software for convex programming and can be downloaded from http://cvxr.com.}. However, generic solvers typically have high computational costs and do not scale well with the dimension and the number of data points. 

In this section, we study efficient implementations of the proposed sparse optimizations using an Alternating Direction Method of Multipliers (ADMM) method \cite{Boyd:FTML10, Gabay:CMA76}. We fist consider the most general optimization program
\begin{eqnarray}
\label{eq:ssc-implementation1ap}
\begin{split}
&\min_{(\C,\E,\Z)} \; \| \C \|_1 + \lambda_{e} \| \E \|_1 + \frac{\lambda_z}{2} \| \Z \|_F^2  \\ &\st \;\;\; \Y = \Y \C + \E + \Z, \;\; \C^\top \1 = \1, ~~\diag(\C) = \0,
\end{split}
\end{eqnarray}
and present an ADMM algorithm to solve it. 

First, note that using the equality constraint in \eqref{eq:ssc-implementation1ap}, we can eliminate $\Z$ from the optimization program and equivalently solve
\begin{eqnarray}
\label{eq:ssc-implementation2ap}
\begin{split}
&\min_{(\C,\E)} \; \| \C \|_1 + \lambda_{e} \| \E \|_1 + \frac{\lambda_z}{2} \|  \Y - \Y \C - \E  \|_F^2  \\ &\st \;\;\; \C^\top \1 = \1, ~~\diag(\C) = \0.
\end{split}
\end{eqnarray}
The overall procedure of the ADMM algorithm is to introduce appropriate auxiliary variables into the optimization program, augment the constraints into the objective function, and iteratively minimize the Lagrangian with respect to the primal variables and maximize it with respect to the Lagrange multipliers. With an abuse of notation, throughout this section, we denote by $\diag(\C)$ both a vector whose elements are the diagonal entries of $\C$ and a diagonal matrix whose diagonal elements are the diagonal entries of $\C$. 

\begin{algorithm*}[t!]
\caption{\bf: Solving \eqref{eq:ssc-implementation1ap} via an ADMM Algorithm}
\label{alg:ssc-admm1ap}
\textbf{Initialization:} Set maxIter $ = 10^4$, $k = 0$, and Terminate $\leftarrow$ False. Initialize $\C^{(0)}, \A^{(0)}, \E^{(0)}, \deltab^{(0)},$ and $\Deltab^{(0)}$ to zero.
\begin{algorithmic}[1]
\While{(Terminate == False)}
\State update $\A^{(k+1)}$ by solving the following system of linear equations
\begin{equation*}
(\lambda_z \Y^\top \Y + \rho \I + \rho \1 \1^{\top}) \A^{(k+1)} = \lambda_z \Y^\top (\Y - \E^{(k)}) + \rho ( \1 \1^\top + \C^{(k)}) - \1 \deltab^{(k) \top} - \Deltab^{(k)},
\end{equation*} 
%
\State update $\C^{(k+1)}$ as $\C^{(k+1)} = \J - \diag(\J)$, where $\J \triangleq \mathcal{T}_{\frac{1}{\rho}}(\A^{(k+1)} + \Deltab^{(k)}/\rho)$,
\State update $\E^{(k+1)}$ as $\E^{(k+1)} = \mathcal{T}_{\frac{\lambda_e}{\lambda_z}}(\Y - \Y \A^{(k+1)})$,
\State update $\deltab^{(k+1)}$ as $\deltab^{(k+1)} = \deltab^{(k)} + \rho \, (\A^{(k+1)\top} \1 - \1)$,
\State update $\Deltab^{(k+1)}$ as $\Deltab^{(k+1)} = \Deltab^{(k)} + \rho \, (\A^{(k+1)} - \C^{(k+1)})$,
\State $k \leftarrow k + 1$,
\If {$(\| \A^{(k)\top} \1 - \1 \|_\infty \leq \epsilon$ and $\| \A^{(k)} - \C^{(k)} \|_\infty \leq \epsilon$ and $\| \A^{(k)} - \A^{(k-1)} \|_\infty \leq \epsilon$ and $\| \E^{(k)} - \E^{(k-1)} \|_\infty \leq \epsilon$ or $(k \geq $ maxIter$)$}
\State Terminate $\leftarrow$ True
\EndIf
\EndWhile
\end{algorithmic}
\textbf{Output:} Optimal sparse coefficient matrix $\C^* = \C^{(k)}$.
\end{algorithm*}
To start, we introduce an auxiliary matrix $\A \in \Re^{N \times N}$ and consider the optimization program
\begin{eqnarray}
\label{eq:ssc-implementation3ap}
\begin{split}
&\min_{(\C,\E,\A)} \; \| \C \|_1 + \lambda_{e} \| \E \|_1 + \frac{\lambda_z}{2} \|  \Y - \Y \A - \E  \|_F^2  \\ &\st \;\;\; \A^\top \1 = \1, ~~\A = \C -\diag(\C).
\end{split}
\end{eqnarray}
whose solution for $(\C,\E)$ coincides with the solution of \eqref{eq:ssc-implementation2ap}. As we will see shortly, introducing $\A$ helps to obtain efficient 
updates on the optimization variables. 
Next, using a parameter $\rho > 0$, we add to the objective function of \eqref{eq:ssc-implementation3ap} two penalty terms corresponding to the constraints $\A^\top \1 = \1$ and $\A = \C - \diag(\C)$ and consider the following optimization program
\begin{eqnarray}
\label{eq:ssc-implementation4ap}
\begin{split}
\min_{(\C,\E,\A)} \; & \| \C \|_1 + \lambda_{e} \| \E \|_1 + \frac{\lambda_z}{2} \|  \Y - \Y \A - \E  \|_F^2 \\ & + \frac{\rho}{2} \| \A^\top \1 - \1 \|_2^2 \quad  + \frac{\rho}{2} \| \A - (\C - \diag(\C)) \|_F^2 \\ 
&\; \st \;\;\; \A^\top \1 = \1, ~~\A = \C -\diag(\C).
\end{split}
\end{eqnarray}
Note that adding the penalty terms to \eqref{eq:ssc-implementation3ap} do not change its optimal solution, \ie, both \eqref{eq:ssc-implementation3ap} and \eqref{eq:ssc-implementation4ap} have the same solutions, since for any feasible solution of \eqref{eq:ssc-implementation4ap} that satisfies the constraints, the penalty terms vanish. However, adding the penalty terms makes the objective function strictly convex in terms of the optimization variables $(\C,\E,\A)$, which allows using the ADMM approach.

Introducing a vector $\deltab \in \Re^N$ and and a matrix $\Deltab \in \Re^{N \times N}$ of Lagrange multipliers for the two equality constraints in \eqref{eq:ssc-implementation4ap}, we can write the Lagrangian function of \eqref{eq:ssc-implementation4ap} as
\begin{multline}
\L(\C,\A,\E,\deltab,\Deltab) = \; \| \C \|_1 + \lambda_{e} \| \E \|_1 + \frac{\lambda_z}{2} \|  \Y - \Y \A - \E  \|_F^2 \\ + \frac{\rho}{2} \| \A^\top \1 - \1 \|_2^2 + \frac{\rho}{2} \| \A - (\C - \diag(\C)) \|_F^2 \\ +  \deltab^{\top} (\A^\top \1 - \1) + \tr( \Deltab^{\top}(\A - \C + \diag(\C)) ),
\end{multline}
where $\tr(\cdot)$ denotes the trace operator of a given matrix. The ADMM approach then consists of an iterative procedure as follows: Denote by $(\C^{(k)}, \E^{(k)}, \A^{(k)})$ the optimization variables at iteration $k$, and by $(\deltab^{(k)},\Deltab^{(k)})$ the Lagrange multipliers at iteration $k$ and
\begin{itemize}
\item Obtain $\A^{(k+1)}$ by minimizing $\L$ with respect to $\A$, while $(\C^{(k)}, \E^{(k)},\deltab^{(k)},\Deltab^{(k)})$ are fixed. Note that computing the derivative of $\L$ with respect to $\A$ and setting it to zero, we obtain
\begin{multline}
\label{eq:Aupdateap}
(\lambda_z \Y^\top \Y + \rho \I + \rho \1 \1^{\top}) \A^{(k+1)} = \lambda_z \Y^\top (\Y - \E^{(k)}) \\+ \rho ( \1 \1^\top + \C^{(k)}) - \1 \deltab^{(k) \top} - \Deltab^{(k)}.
\end{multline}
In other words, $\A^{(k+1)}$ is obtained by solving an $N \times N$ system of linear equations. When $N$ is not very large, one can simply matrix inversion to obtain $\A^{(k+1)}$ from \eqref{eq:Aupdateap}. For large values of $N$, conjugate gradient methods should be employed to solve for $\A^{(k+1)}$.
\item Obtain $\C^{(k+1)}$ by minimizing $\L$ with respect to $\C$, while $(\A^{(k)}, \E^{(k)},\deltab^{(k)},\Deltab^{(k)})$ are fixed. Note that the update on $\C$ also has a closed-form solution given by
\begin{eqnarray}
\C^{(k+1)} = \J - \diag(\J),\\
\J \triangleq \mathcal{T}_{\frac{1}{\rho}}(\A^{(k+1)} + \Deltab^{(k)}/\rho),
\end{eqnarray}
where $\mathcal{T}_{\eta}(\cdot)$ is the shrinkage-thresholding operator acting on each element of the given matrix, and is defined as
\begin{equation}
\mathcal{T}_{\eta}(v) = (|v| - \eta)_{+} \operatorname{sgn}(v).
\end{equation}
The operator $(\cdot)_{+}$ returns its argument if it is non-negative and returns zero otherwise.
\item Obtain $\E^{(k+1)}$ by minimizing $\L$ with respect to $\E$, while $(\C^{(k+1)}, \A^{(k+1)},\deltab^{(k)},\Deltab^{(k)})$ are fixed. The update on $\E$ can also be computed in closed-form as
\begin{equation}
\E^{(k+1)} = \mathcal{T}_{\frac{\lambda_e}{\lambda_z}}(\Y \A^{(k+1)} - \Y),
\end{equation}
\item Having $(\C^{(k+1)}, \A^{(k+1)},\E^{(k+1)}$ fixed, perform a gradient ascent update with the step size of $\rho$ on the Lagrange multipliers as
\begin{eqnarray}
\deltab^{(k+1)} &=& \deltab^{(k)} \;\,+ \rho \, (\A^{(k+1) \top} \1 - \1),\\
\Deltab^{(k+1)} &=& \Deltab^{(k)} + \rho \, (\A^{(k+1)} - \C^{(k+1)}).
\end{eqnarray}
\end{itemize}
These three steps are repeated until convergence is achieved or the number of iterations exceeds a maximum iteration number. Convergence is achieved when we have $\| \A^{(k)\top} \1- \1 \|_\infty \leq \epsilon$, $\| \A^{(k)}- \C^{(k)} \|_\infty \leq \epsilon$, $\| \A^{(k)}- \A^{(k-1)} \|_\infty \leq \epsilon$ and $\| \E^{(k)} - \E^{(k-1)} \|_\infty \leq \epsilon$, where $\epsilon$ denotes the error tolerance for the primal and dual residuals. In practice, the choice of $\epsilon = 10^{-4}$ works well in real experiments. In summary, Algorithm \ref{alg:ssc-admm1ap} shows the updates for the ADMM implementation of the optimization program \eqref{eq:ssc-implementation1ap}. 

\section*{Computational Time Comparison}

\begin{table}[h!]
\caption{\footnotesize Average computational time (sec.) of the algorithms on the Extended~Yale~B~dataset as a function of the number of subjects.} \centering
\vspace{-1mm}
\begin{small}
\begin{tabular}{|@{\;\,}c@{\;\,}|@{\;\,}c@{\;\,}|@{\;\,}c@{\;\,}|@{\;\,}c@{\;\,}|@{\;\,}c@{\;\,}|@{\;\,}c@{\;\,}|}
\hline
                    & LSA & SCC & LRR & LRSC & SSC\\
\hline
2 Subjects &  $5.2$ & $262.8$ & $1.6$ & $1.1$ & $1.8$\\
\hline
3 Subjects &  $13.4$ & $451.5$ & $2.2$ & $1.9$ & $3.29$\\
\hline
5 Subjects &  $62.7$ & $630.3$ & $7.6$ & $5.7$ & $11.4$\\
\hline
8 Subjects &  $180.2$ & $1020.5$ & $22.1$ & $16.3$ & $42.6$\\
\hline
10 Subjects &  $405.3$ & $1439.8$ & $255.0$ & $96.9$ & $160.3$\\
\hline
\end{tabular}
\end{small}
\label{tab:compTime}
\end{table}
%


Table \ref{tab:compTime} shows the computational time of different algorithms on the Extended~Yale~B~dataset as a function of the number of subjects. Note that these computational times are based on the codes of the algorithms used by their authors. It is important to mention that LRR and SSC can be implemented using faster optimization solvers. More specifically, LRR can be  made faster using LADMAP method proposed in: ``Z. Lin and R. Liu and Z. Su, Linearized Alternating Direction Method with Adaptive Penalty for Low-Rank Representation, NIPS 2011." Also, SSC can be made faster using LADM method proposed in ``J. Yang and Y. Zhang. Alternating direction algorithms for $\ell_1$ problems in compressive sensing. SIAM J. Scientific Computing, 2010."

\end{document}